\pdfoutput=1

\documentclass[11pt]{article}

\usepackage[]{acl}

\usepackage{times}
\usepackage{latexsym}

\usepackage[T1]{fontenc}

\usepackage[utf8]{inputenc}

\usepackage{microtype}

\usepackage{inconsolata}

\usepackage{graphicx}

\usepackage{amsmath,amsfonts,bm}
\usepackage{wrapfig}
\usepackage{booktabs}
\usepackage{multirow}
\usepackage{array,multirow}
\usepackage{float}
\usepackage{colortbl}
\usepackage{afterpage}
\usepackage{placeins}
\usepackage{caption}
\usepackage{subcaption}

\definecolor{patch_color}{RGB}{123,175,171}
\definecolor{context_color}{RGB}{171,178,116}
\definecolor{perturbed_color}{RGB}{195,156,106}

\newcommand{\mpararel}{\textsc{mParaRel}}
\newcommand{\pararel}{\textsc{ParaRel}}
\newcommand{\xglm}{\textsc{XGLM}}
\newcommand{\mtfive}{mT5}
\newcommand{\eurollm}{\textsc{EuroLLM}}

\title{How Do Multilingual Language Models Remember Facts?}

\author{Constanza Fierro$^{\dagger}$ \ \ Negar Foroutan$^{\ddagger}$ \ \ 
        Desmond Elliott$^{\dagger}$ \ \ \textbf{Anders S{\o}gaard}$^{\dagger}$ \vspace{0.2cm} \\
        $^{\dagger}$ Department of Computer Science, University of Copenhagen \\ 
        $^{\ddagger}$ EPFL
        }

\begin{document}
\maketitle

\def\thefootnote{*}\footnotetext{Correspondence: Constanza Fierro <\href{mailto:c.fierro@di.ku.dk}{c.fierro@di.ku.dk}>.}\def\thefootnote{\arabic{footnote}}

\begin{abstract}
Large Language Models (LLMs) store and retrieve vast amounts of factual knowledge acquired during pre-training. Prior research has localized and identified mechanisms behind knowledge recall; however, it has only focused on English monolingual models. The question of how these mechanisms generalize to non-English languages and multilingual LLMs remains unexplored. In this paper, we address this gap by conducting a comprehensive analysis of three multilingual LLMs. First, we show that previously identified recall mechanisms in English largely apply to multilingual contexts, with nuances based on language and architecture. Next, through patching intermediate representations, we localize the role of language during recall, finding that subject enrichment is language-independent, while object extraction is language-dependent. Additionally, we discover that the last token representation acts as a Function Vector (FV), encoding both the language of the query and the content to be extracted from the subject. Furthermore, in decoder-only LLMs, FVs compose these two pieces of information in two separate stages. These insights reveal unique mechanisms in multilingual LLMs for recalling information, highlighting the need for new methodologies—such as knowledge evaluation, fact editing, and knowledge acquisition—that are specifically tailored for multilingual LLMs.\footnote{\url{https://github.com/constanzafierro/multilingual_factual_memorization}}
\end{abstract}

\section{Introduction}

Large Language Models (LLMs) learn extensive factual knowledge during pre-training, including propositional facts like ``The capital of France is \_\_'' \cite{petroni-etal-2019-language}. While multilingual models also acquire such knowledge, their performance varies significantly across languages \citep{kassner-etal-2021-multilingual,jiang-etal-2020-x,yin-etal-2022-geomlama}, raising questions about whether this variation stems from language-specific storage or phrasing sensitivity \citep{elazar-etal-2021-measuring}. Such assessments will be critical for determining the trustworthiness of multilingual LLMs, if trust relies on knowledge \citep{Grasswick2010-GRASAL-2,hawleytrust2012,Nguyen2022-NGUTAA}.

\begin{figure}[t]
\centering
\begin{minipage}[t]{0.9\linewidth}
    \centering
    \includegraphics[width=\linewidth]{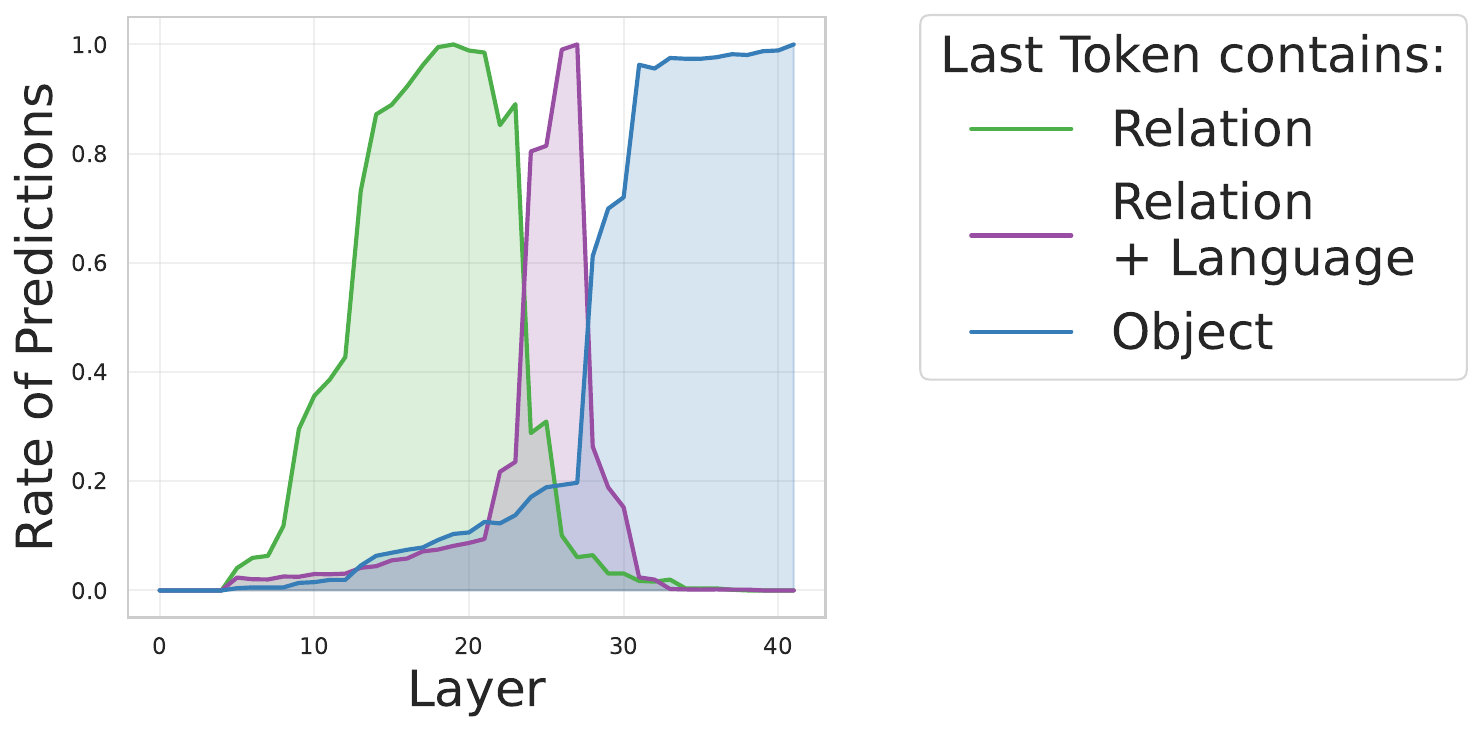}
\end{minipage}
\vspace{-3mm}
  \caption{Aggregated patching results (\S\ref{sec:patching}) for \eurollm{}. In propositional factual statements, e.g. ``Paris is the capital of'', the last token representation contains the function that solves the task. This function is formed in two stages: first, the relation to extract is encoded (green), and then the language is composed into the function (purple). Finally, the function is applied to the subject, and the predicted object is resolved (blue).}\label{fig:summary_patching_eurollm}
  \vspace{-4mm}
\end{figure}

Mechanistic interpretability research has begun uncovering how models store and retrieve knowledge internally, with recent studies identifying specific components for knowledge storage \citep{meng2022locating, sharma2024locating} and retrieval mechanisms \citep{geva-etal-2023-dissecting, chughtai2024summing}. However, they have focused exclusively on English LLMs, mainly autoregressive ones.\footnote{Except \citet{sharma2024locating}, who extend these analyses to Mamba \citep{gu2023mamba}, a state-space language model.} Encoder-decoder architectures, which could enable better cross-lingual representations \citep{li-etal-2024-comparison}, remain underexplored. Additionally, recent studies have shown that LLMs share circuits across languages for specific tasks \citep{ferrando-costa-jussa-2024-similarity,zhang2025the}, but these are limited to syntactic tasks and do not address how concepts are represented or recalled cross-lingually. While \citet{dumas2024how} examined how disentangled language is from concepts using a translation task, here we analyze fact recall, which allows us to offer broader insights into how language is encoded.

In this paper, we study the mechanisms of factual recall in multilingual LLMs, focusing on two architectures: decoder-only (\xglm{} and \eurollm{}) and encoder-decoder (\mtfive{}). We analyze a simple form of information extraction, where the input contains a subject and a relation, and the model predicts the corresponding object (e.g. ``Paris'' in the earlier example). Our analysis centers on three key questions: (1) Does the localization of factual knowledge in English LLMs extend to multilingual LLMs? (2) Are the factual recall mechanisms found for English LLMs also present in multilingual LLMs? (3) When does language play a role in the recall mechanism?

To address the first question, we use causal tracing analysis to assess if early MLP modules processing the final subject token are as decisive in multilingual models as in English-centric ones \citep{meng2022locating}. Our results (\S\ref{sec:causal_tracing}) show that \eurollm{} and \mtfive{} exhibit strong causal effects for the last subject token—\eurollm{} in earlier layers, and \mtfive{} across all encoder layers. Additionally, in all models, both MLPs and MHSAs in later layers play a decisive role in recovering factual information---unlike in  \citet{meng2022locating}, where only MHSAs were active at the late site.

Next, we investigate the second question by analyzing the three-step process described by \citet{geva-etal-2023-dissecting}: the relation information flows to the final token, followed by subject information, and finally, the attention layers extract the object (\S\ref{sec:geva_analyses}). We find that in multilingual LLMs, subject information flows similarly to monolingual ones, but non-subject token flow differs, and this analysis alone cannot determine the path of relation information. Regarding the final extraction, both feed-forward and attention sublayers contribute in decoder-only LLMs---unlike in English autoregressive LLMs, where attention modules dominate. In \mtfive{}, this mechanism is primarily handled by cross-attention.
Overall, our findings indicate that some of the localization (\S\ref{sec:causal_tracing}) and mechanisms (\S\ref{sec:geva_analyses}) of fact retrieval in English LLMs generalize to multilingual LLMs, but with key variations.

Finally, to address the third question, we investigate where language plays a role within the three-step process, to characterize how and where factual knowledge cross-lingual transfer may occur. Using activation patching \citep{zhang2024towards} we insert the intermediate representation of the last token from an English forward pass into the forward pass of another language (\S\ref{sec:patching}). Our results reveal that the last token acts as a Function Vector (FV) \cite{todd2024function}, encoding both the relation \emph{and} the output language, which is then applied to the subject in the context. Crucially, the fact that the FV formed in one language can be used with an input in another language demonstrates that the subject and relation representations are largely language-independent, while the extraction event is language-specific. Furthermore, in decoder-only LLMs, the FV is constructed in two distinct phases: first, it encodes only the relation, and later, the language is incorporated (Figure \ref{fig:summary_patching_eurollm}).

These findings advance our understanding of factual recall, with implications for cross-lingual knowledge transfer, knowledge editing, and trustworthiness in multilingual LLMs. Our results on language flow open new avenues for studying whether models `think' in English \citep{wendler-etal-2024-llamas}, by examining how the FV is altered across languages. Additionally, the late-site causal effect of MLPs and their dominance in the extraction phase suggest that fact editing techniques and knowledge evaluations must extend beyond early MLPs and attention layers \citep{mela-etal-2024-mass}.

\section{Related Work}

\paragraph{Factual Knowledge Recall} \citet{petroni-etal-2019-language} analyzed factual knowledge in pre-trained language models using the LAMA dataset, which pairs cloze-test templates with WikiData triplets.\footnote{One might object to treating cloze-test performance as indicative of knowledge, given LLM inconsistencies. However, following \citet{Fierro2024-FIEDKB}, we adopt this view, as LLMs are often consistent and can sometimes justify responses through world models or training data attribution.} This was extended to multilingual settings by translating LAMA \citep{kassner-etal-2021-multilingual,jiang-etal-2020-x}. Later, \citet{elazar-etal-2021-measuring} evaluated the \textit{consistency} of such knowledge by manually curating paraphrases of LAMA to create \pararel{}. Further work examined factual consistency cross-lingually \citep{fierro-sogaard-2022-factual,qi-etal-2023-cross}, using machine-translated \pararel{} templates and WikiData-based subject/object translations to produce \mpararel{}.

\paragraph{Interpretability on Factual Knowledge Recall}
Early studies by \citet{meng2022locating} and \citet{geva-etal-2023-dissecting} mechanistically analyzed knowledge localization and information flow in GPT models \citep{brown2020language}, using English data. Later, \citet{chughtai2024summing} showed that answers emerge from summing independent components in GPT and Pythia models, while \citet{sharma2024locating} extended these findings to Mamba \citep{gu2023mamba}, also focusing on English.

\paragraph{Activation Patching} \citet{ghandeharioun2024patchscopes} introduced Patchscope, a framework for decoding intermediate representations by patching them into forward passes. \citet{dumas2024how} applied this method to last-token representations in a translation task, showing that models encode language-agnostic concepts---consistent with our findings and those of \citet{foroutan-etal-2022-discovering}. \citet{wang-etal-2024-locating} similarly found that the last token encodes the relation in factual English queries at a specific stage of computation. We extend these findings to a cross-lingual factual recall setting. While \citet{dumas2024how} concluded that models first resolve the output language, we find that models first resolve the query’s relation, then the language. We thus interpret the last token as a function vector \citep{todd2024function}, which encodes the task-specific function---e.g., translating in the case of \citet{dumas2024how}. Additionally, we show that in decoder-only models, the FV composes the output language onto the relation function vector.
\section{Experimental Setup}
We focus on a simple form of factual knowledge recall, where LMs are tasked with predicting the correct object \(o\) for a given subject \(s\) and a relation \(r\). 
These \((s,r,o)\) triplets are obtained from WikiData, and natural language templates are used to describe the relation, with placeholders for the subject and object.\footnote{
For example, the relation born-in could use the template ``[X] was born in [Y]'', where [X] is the subject and [Y] is the object to be predicted.} For our analysis, we select 10 typologically diverse languages representing various scripts, families, and word orders: English (\textit{en}), Spanish (\textit{es}), Vietnamese (\textit{vi}), Turkish (\textit{tr}), Russian (\textit{ru}), Ukrainian (\textit{uk}), Japanese (\textit{ja}), Korean (\textit{ko}), Hebrew (\textit{he}), Persian (\textit{fa}), and Arabic (\textit{ar}). See Table \ref{tab:languages_chars} for language characteristics.

\paragraph{Models} We analyze decoder-only and encoder-decoder architectures.\footnote{For decoder-only models, we only use the templates in \mpararel{} that have the object placeholder at the end of the sentence, while for encoder-decoder, we use all the templates.} The decoder-only LLMs are \xglm{} \citep{lin2021few}, with 7.5B parameters and 32 layers, and \eurollm{} \citep{martins2024eurollm} with 9B parameters and 42 layers; the encoder-decoder \mtfive{}-xl \citep{xue-etal-2021-mt5} has with 3.7B parameters and 24 encoder-decoder layers. The pre-training data varies across models: \mtfive{} is pretrained on 101 languages, covering all the languages in our study; \xglm{} covers 30 languages, excluding \textit{uk}, \textit{he}, and \textit{fa}; while \eurollm{} covers 35 languages, excluding \textit{vi}, \textit{he}, and \textit{fa}.

\paragraph{Data} We use the \mpararel{} dataset \citep{fierro-sogaard-2022-factual}, which includes triplets and templates for 45 languages.\footnote{We augment the objects in \mpararel{}, filter out trivial examples, and when there are enough examples, we use a crosslingual subset (see Appendix \ref{appendix:experimental_setup}).}

\begin{figure*}[t]
    \centering
    \vspace{-7mm}
    \includegraphics[width=0.7\textwidth]{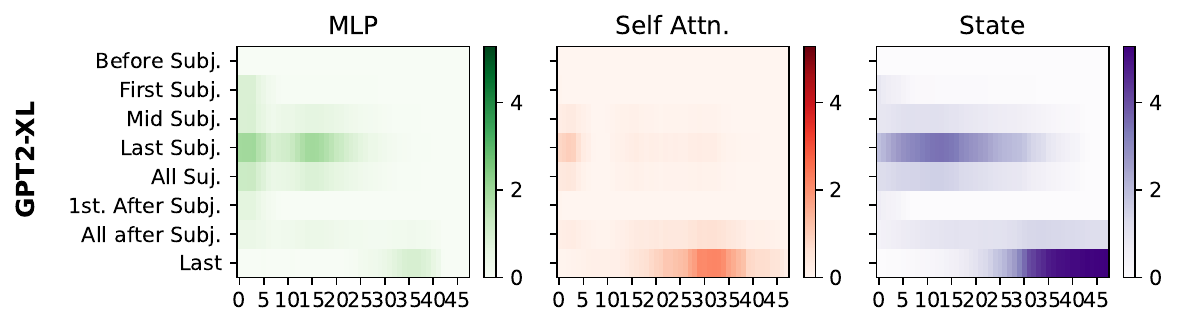}
    \includegraphics[width=0.7\textwidth]{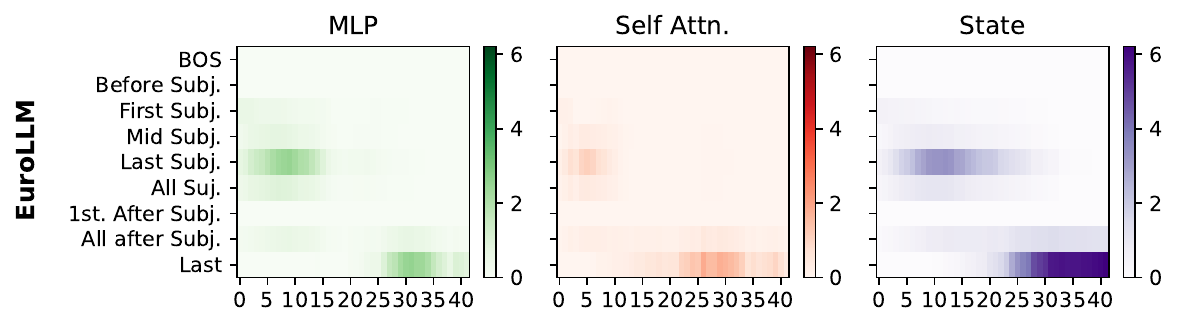}
    \vspace{-3mm}
    \caption{Average indirect effect (IE) on logit scores, when the subject input is corrupted and some activations are restored to their uncorrupted values. The \(y\) axis indicates the layer where the restoration was performed, and the \(x\) axis specifies the token(s) over which we average the IE. The MHSA and MLP are restored in windows of 12\% consecutive layers. Top GPT2-XL (only English data), bottom \eurollm{} (\xglm{} and \mtfive{} in Figure \ref{fig_appendix:avg_causal_analysis}).}
    \label{fig:avg_causal_analysis}
    \vspace{-3mm}
\end{figure*}

To investigate the process of knowledge recall we only consider examples where the model predicts the \textit{correct} object completion \citep{meng2022locating, geva-etal-2023-dissecting}. Since \mpararel{} provides multiple paraphrased templates for each relation, we greedy-decode for every available template corresponding to a given triplet and check for an exact match. \begin{wraptable}{r}{4.2cm}
\centering\vspace{-0.35\intextsep}\footnotesize\setlength{\tabcolsep}{3pt}
\begin{tabular}{lccc}
\toprule
 & \xglm{} & \eurollm{} & \mtfive{} \\
\midrule
en & 1812 & 2332 & 1543 \\
es & 1380 & 1913 & 1192 \\
vi & 1646 & 779 & 993 \\
tr & 418 & 799 & 1058 \\
ru & 830 & 1680 & 683 \\
uk & 213 & 1244 & 456 \\
ko & 116 & 308 & 630 \\
ja & 6 & 42 & 358 \\
he & 13 & 107 & 565 \\
fa & 7 & 31 & 406 \\
ar & 811 & 1790 & 488 \\ \bottomrule
\end{tabular}
\caption{Number of facts \((s,r,o)\) correctly predicted.}\label{table_mem_examples}
\vspace{-2mm}
\end{wraptable} If multiple templates yield a match, we randomly select one for the analysis. In cases where an article or other filler tokens precede the object, we include these tokens in the input text to ensure that when the example is fed into the model for our analysis,  the next predicted token is the first token of the object (Implementation details in Appendix \ref{appendix:prompt_details}).
Table~\ref{table_mem_examples} presents the number of  examples for which the correct object is predicted. We exclude languages with too few examples from our analysis, namely \textit{ko}, \textit{ja}, \textit{he}, and \textit{fa} for \xglm{}, and \textit{ja}, \textit{he}, and \textit{fa} for \eurollm{}.

\paragraph{Notation} Given a transformer model with \(L\) layers, let \(h_t^l\) be the representation of the token \(t\) at layer \(l\). When the model is an encoder-decoder, let \(e_i^l\) be the representation of the encoder layer \(l\) for the \(i\)-th token in the encoder input. Then, the encoder layer computes \(h_t^{l+1} = h_t^l + s^l + f^l\) and the decoder \(h_t^{l+1} = h_t^l + s^l + c^l + f^l\), where \(s^l = \text{Self Attn.}(h_0^l \ldots h_t^l)\), \(c^l = \text{Cross Attn.}(h_t^l, e_0^l \ldots e_n^l)\) and \(f^l = \text{MLP}(h_t^l + s^l + c^l)\). If decoder-only, then \(c^l\) does not apply.

\section{Causal Tracing}
\label{sec:causal_tracing}

We first analyze which hidden states in the model's computation are more important than others when recalling a fact. Following \citet{meng2022locating}, we trace the causal effects of hidden states using causal mediation analysis \citep{pearl2022direct}.  
Let \(\mathbb{P}(o)\) be the probability of the predicted object token, and \(LS(o)\) its logit score. We corrupt the input by adding Gaussian noise to the subject tokens,\footnote{We follow \citet{meng2022locating} and add \(\epsilon \sim \mathcal{N}(0, (3\sigma)^2)\), with \(\sigma\) being the standard deviation of the subjects tokens embeddings from the data used. We repeat the experiment ten times with different noise samples, and report the average.} and observe the corrupted probability \(\tilde{\mathbb{P}}(o)\) of the originally predicted token. Then, we run inference again on the corrupted input, but this time, we restore a specific hidden state in the model and track the probability \(\tilde{\mathbb{P}}_{\text{restored}}(o)\). We study the indirect effect of such component as \( \text{IE}_{\mathbb{P}}=\tilde{\mathbb{P}}_{\text{restored}}(o) - \tilde{\mathbb{P}}(o)\), or if using logits \(\text{IE}_{LS}=\tilde{LS}(o) - \tilde{LS}_{\text{restored}}(o)\). Specifically, we restore a \textit{state} by setting \(\tilde{h}_t^l \leftarrow h_t^l\), where \(h_t^l\) is the hidden state from the clean run and \(\tilde{h}_t^l\) that of the corrupted run; and similarly, we restore the self-attention layers contribution by setting \(\tilde{s}^l \leftarrow s^l\) for all the attention modules in a window of size \(w\) (analogous for \(c\) and \(f\)).\footnote{We use windows because, generally, the contributions of the sub-layers are gradual \citep{geva-etal-2021-transformer}. In other words, multiple layers contribute the same behavior to the residual, and their sum produces the observed effect.} We restore windows of \(12\%\) consecutive layers, so \(w\)=\(4\) for \xglm{}, \(w\)=\(5\) for \eurollm{}, and \(w\)=\(3\) for \mtfive{}.

We compare our results in multilingual LLMs to those of \citet{meng2022locating}, who analyzed the factual recall of GPT-2 XL in English and reached two key conclusions: (1) they identified an ``early site'', where the MLPs processing the last subject token in the early and middle layers play a crucial role in recovering from input corruption; and (2), they found a ``late site'', where the attention modules processing the last token in the later layers also significantly contribute to prediction recovery.\footnote{While they consider the late site unsurprising, as it directly precedes the final prediction, this observation primarily applies to hidden state restoration, not necessarily attention restoration. We, however, interpret the late-site attention as aligning with what \citet{geva-etal-2023-dissecting} referred to as the extraction event (\S\ref{sec:geva_analyses}).}

\begin{figure*}[t]
    \centering
    \vspace{-4mm}
    \includegraphics[width=0.83\textwidth]{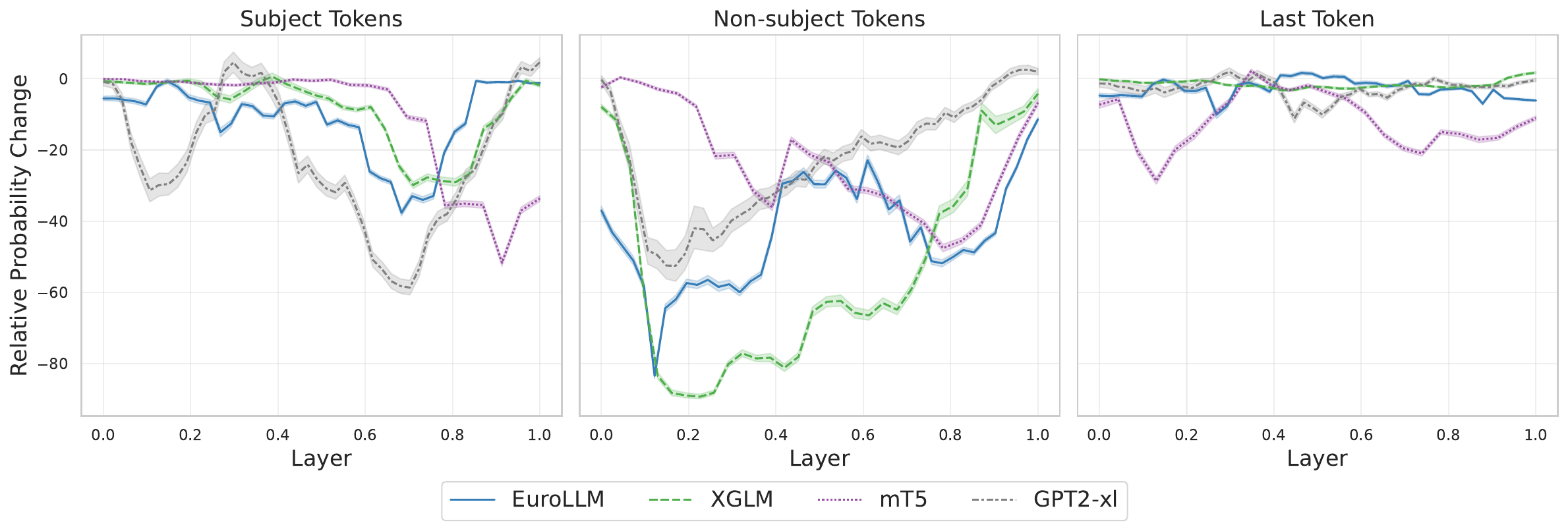}
    \vspace{0.5cm}
    \includegraphics[width=0.83\textwidth]{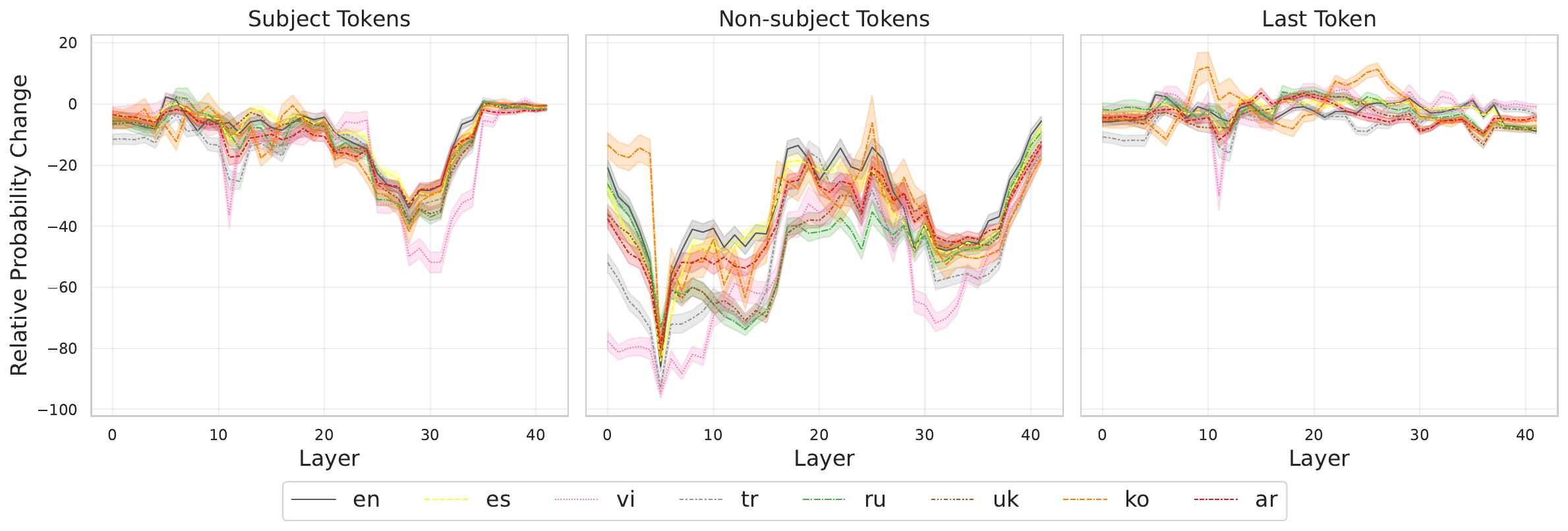}
    \vspace{-6mm}
    \caption{Attention knockout between the last token and a given set of tokens. Each layer represents the effect of the knockout on a window of \(w\) layers. Top is the average, bottom \eurollm{} (\(w=7\)). \xglm{} and \mtfive{} in Figure \ref{appendix:attention_knockout}.}
    \label{fig:attention_knockout}
    \vspace{-4mm}
\end{figure*}

We present the average causal analysis results in Figure \ref{fig:avg_causal_analysis} (results per language in Appendix~\ref{appendix:causal_tracing}).\footnote{In line with \citet{zhang2024towards} we find that analyzing the \(\text{IE}_{\mathbb{P}}\) overestimates the causal effect of some tokens over others. For example, in \eurollm{} the last subject token MHSA seem more relevant than the last token MLPs (see Figure \ref{fig:eurollm_causal_analysis_probs_1} vs Figure \ref{fig:eurollm_causal_analysis_logits_1}). So we base our main observations using \(\text{IE}_{LS}\).} Our results indicate that the early site in MLPs---centered on the last subject token---is also present in multilingual LLMs. We observe this causal effect in both \eurollm{} and \mtfive{}. In \mtfive{}, the first subject token also shows causal relevance, though marginally less so, and the early site spans all encoder layers. Across languages, the patterns are highly consistent. Interestingly, we do not observe an MLP early site in \xglm{}, either for English or in the MHSAs. As for the late site, we find a strong causal effect in both MLP and MHSA layers just before the final layers in all models. The position of the late site is consistent across languages, with only slight variation in its endpoint for \eurollm{}.

These findings suggest that some conclusions from English LLMs generalize to multilingual ones, but not all. The early MLP site is consistent across models, while at the late site, both MLPs and MHSA are important in multilingual LLMs—unlike in English LLMs, where MHSA dominates. This has implications for knowledge localization and fact editing, which should consider both early and late MLPs in multilingual contexts.
\section{Factual Recall Components}\label{sec:geva_analyses}

\citet{geva-etal-2023-dissecting} described the process of factual knowledge recall in English autoregressive models as a three-step mechanism: (a) 
the subject representation is enriched (i.e., related attributes are encoded); (b) the relation and subject information are propagated to the last token; and (c) the final predicted attribute is extracted by attention layers.
We analyze the information flow to the last token, and then the extraction of the predicted attribute.

\paragraph{Information Flow} We use attention knockout \citep{geva-etal-2023-dissecting} to analyze how information propagates to the final token. This method involves intervening in the attention mechanism: for \xglm{} and \eurollm{}, we modify the last token’s self-attention, while for \mtfive{}, we intervene in the decoder’s cross-attention for the final token. We knock out attention connections by setting the attention scores to zero between the final token and a token set \(\{t\}\), which can be: subject tokens, non-subject tokens, or the last token itself. Following \citet{geva-etal-2023-dissecting} we apply this intervention across consecutive layers within a window of size \(w\) centered on layer \(l\), covering \(18\%\) of the total layers (\(w\)=\(6\) for \xglm{}, \(w\)=\(7\) for \eurollm{}, and \(w\)=\(4\) for \mtfive{}). The relative probability change is given by \((\tilde{\mathbb{P}}(o)-\mathbb{P}(o))/\mathbb{P}(o)\), where \(\tilde{\mathbb{P}}(o)\) is the probability of the originally predicted token \(o\) after attention knockout .A significant drop indicates that the knocked-out tokens contribute critically to the final token’s prediction at that layer.

In Figure~\ref{fig:attention_knockout} we present average results, plots per language can be found in Appendix~\ref{appendix:attn_knockout}. The results show that in each model the information flows fairly similarly for all the languages, and somewhat similar patterns to those in English GPT2-xl. On the one hand, the subject information flows to the last token most critically at the later layers in all models. On the other hand, the non-subject tokens information flows throughout all the layers, with \eurollm{} and \mtfive{} having more similar curves. Nevertheless, we note two differences from the conclusions reached with English GPT \citep{geva-etal-2023-dissecting}: (1) the propagation of non-subject tokens to the last token does not strictly precedes the subject propagation, in \xglm{} the drop is the highest in the earlier layers but it continues to be important until the end, and in \mtfive{} and \eurollm{} this information has two peaks, in the early and later layers; and (2) the last token of \mtfive{} encodes critical information that flows from one layer to the next through the cross-attention (as opposed to the negligible flow of the last token in decoder-only models).\footnote{We hypothesize this may occur because the last token encodes which sentinel token is being generated, indicating where to fill in the input. Since we see an almost identical curve when the last token cannot attend to the sentinel token in the decoder (Figure \ref{fig:mt5_attn_knockout_each_lang}).}

\begin{figure}[t]
    \vspace{-5mm}
    \centering
    \begin{minipage}{0.5\linewidth}
        \centering
        \includegraphics[width=\linewidth]{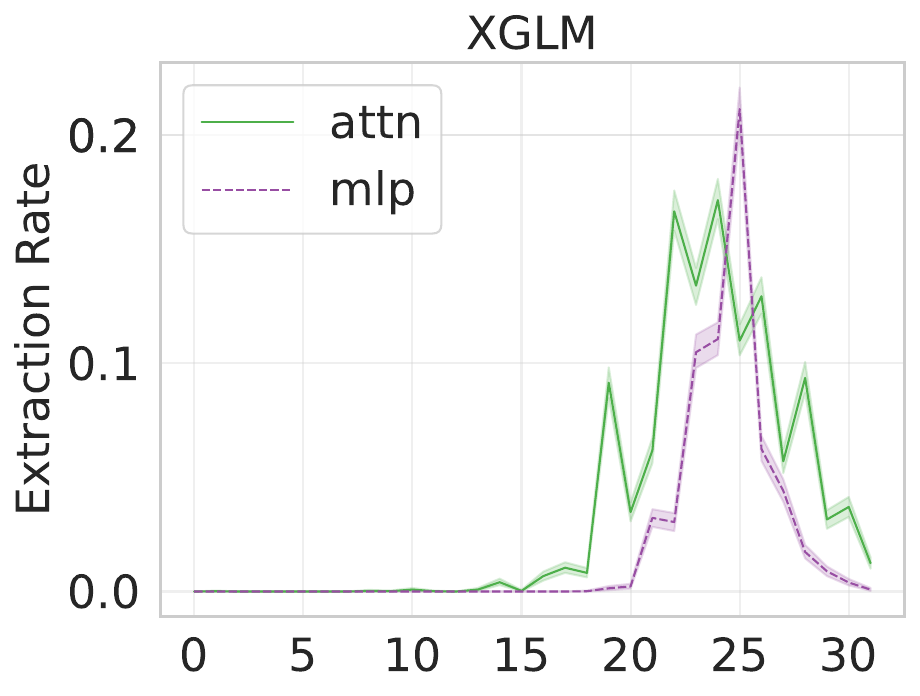}
    \end{minipage}%
    \begin{minipage}{0.5\linewidth}
        \centering
        \includegraphics[width=\linewidth]{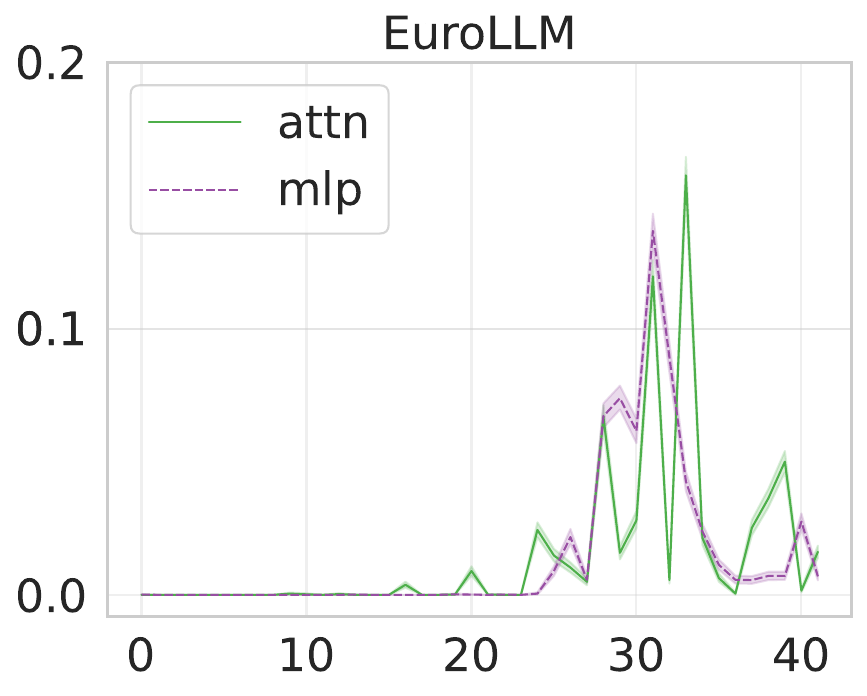}
    \end{minipage}
    \begin{minipage}{0.5\linewidth}
        \centering
        \includegraphics[width=\linewidth]{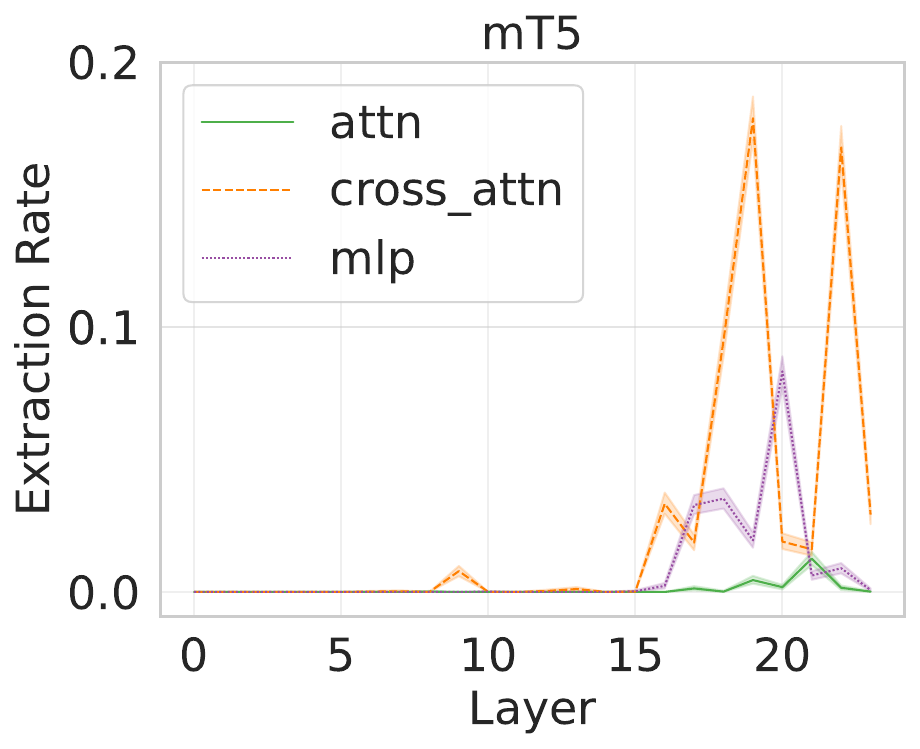}
    \end{minipage}%
    \begin{minipage}{0.5\linewidth}
        \centering
        \includegraphics[width=\linewidth]{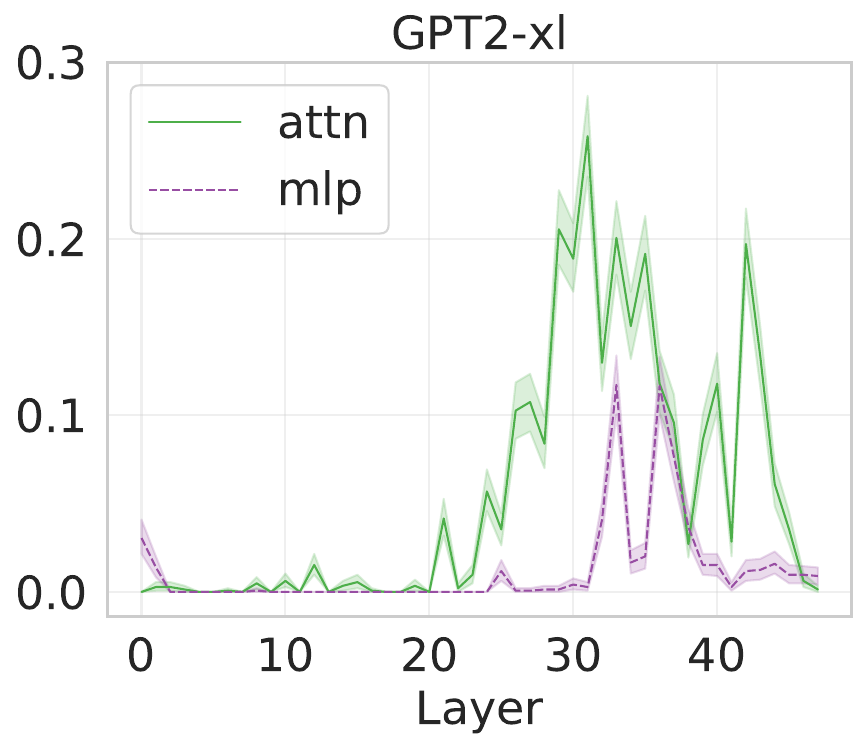}
    \end{minipage}%
    \vspace{-3mm}
    \caption{Extraction rates. Fraction of examples where the top-1 token under the vocabulary projection  matches the final output token.}\label{fig:extraction_rates}
    \vspace{-4mm}
\end{figure}

\paragraph{Prediction Extraction}

Following \citet{geva-etal-2023-dissecting}, we measure extraction events at each layer. Let \(h^l\) be the representation of the \textit{last} token at layer \(l\), and let \(E\) be the embedding matrix; the predicted token is \(o=\arg\,\max(E h^L)\).
An extraction event occurs at layer \(l\) if \(\arg\,\max(E s^l) = o\) (similarly for \(c^l\) and \(f^l\)). The extraction rate is the proportion of examples for which an extraction event occurs at a given layer.

Our results demonstrate that extraction events can be detected in multilingual LLMs (Figure~\ref{fig:extraction_rates}), though rates vary across languages (Appendix~\ref{appendix:prediction_extraction}).\footnote{This variance may result from the strict criteria applied during the extraction event analyses, where only top-1 matches were considered.} A key finding is the prominence of MLP modules in object extraction for multilingual decoder-only models (\xglm{} and \eurollm{}). To rule out the possibility that MLPs simply forward extracted objects from preceding attention layers, we measure how often MLPs perform an extraction without prior attention extraction (Figure \ref{fig:extraction_rates_mlp_vs_attn}). We find that MLPs indeed perform the extraction for most languages, though in English, attention modules dominate, aligning with GPT2-xl results. In \mtfive{}, cross-attention layers drive the extraction, with MLPs playing a secondary role, resembling GPT’s behavior. Additionally, in \mtfive{} the extraction events occur in later layers, with peaks in 19 and 22.

The results show that in multilingual LLMs the extraction mechanism is more complex, involving both attention and MLP modules. This implies that editing techniques should focus not only on early MLPs enhancing subject representations but also on the later MLPs that extract the object.
\begin{figure}[t]
\vspace{-5mm}
  \centering
  \includegraphics[width=0.55\linewidth]{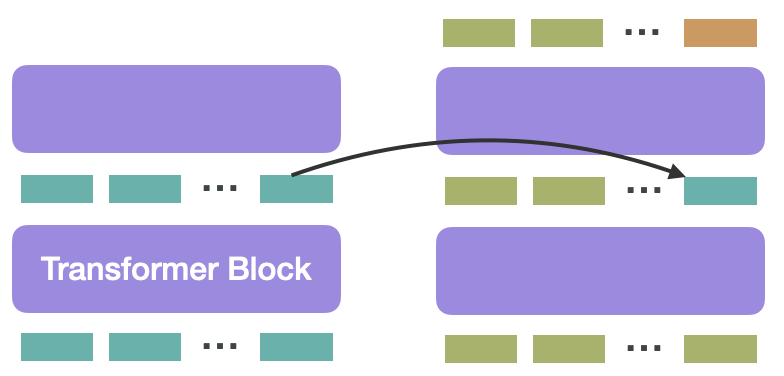}
  \vspace{-2mm}
  \caption{Patching strategy. The \colorbox{patch_color}{patch} example's last token is inserted in the \colorbox{context_color}{context} example's forward pass, which perturbs the \colorbox{perturbed_color}{output} of subsequent layers.}\label{fig:patch_explanation}
  \vspace{-4mm}
\end{figure}

\section{Language Information Flow}\label{sec:patching}
Previous analyses do not provide insights into \emph{how} the language is included and used during recall. In this section, we use activation patching, similarly to \citet{dumas2024how}, to: (1) disentangle when the relation information flows to the last token and when the language information flows, (2) identify which tokens contribute the language information, and (3) interpret the last token representation as a Function Vector~\cite{todd2024function}\footnote{Function Vectors (FV) were originally defined as the mean vector of outputs from specific attention heads. Here, we interpret the last token representation as an FV because it triggers a specific execution for different contexts.}.

\begin{figure*}[t]
    \vspace{-5mm}
    \centering
    \begin{subfigure}{\textwidth}
        \centering
        \hspace{9em}
        \begin{subfigure}{0.45\textwidth}
            \includegraphics[width=\linewidth]{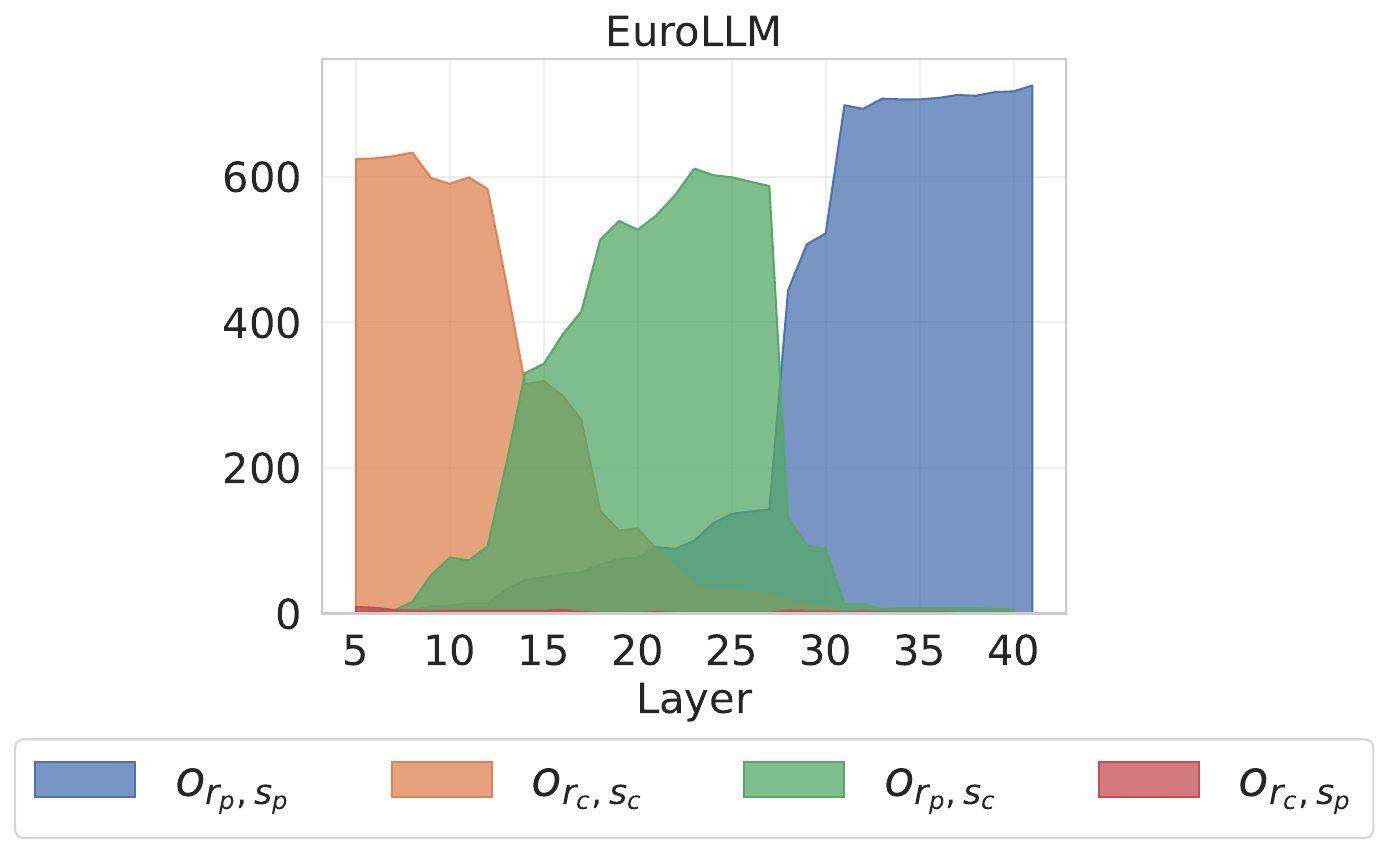}
        \end{subfigure}
        \hspace{-30em}
        \raisebox{7mm}{
        \begin{subfigure}{0.29\textwidth}
            \includegraphics[width=\linewidth]{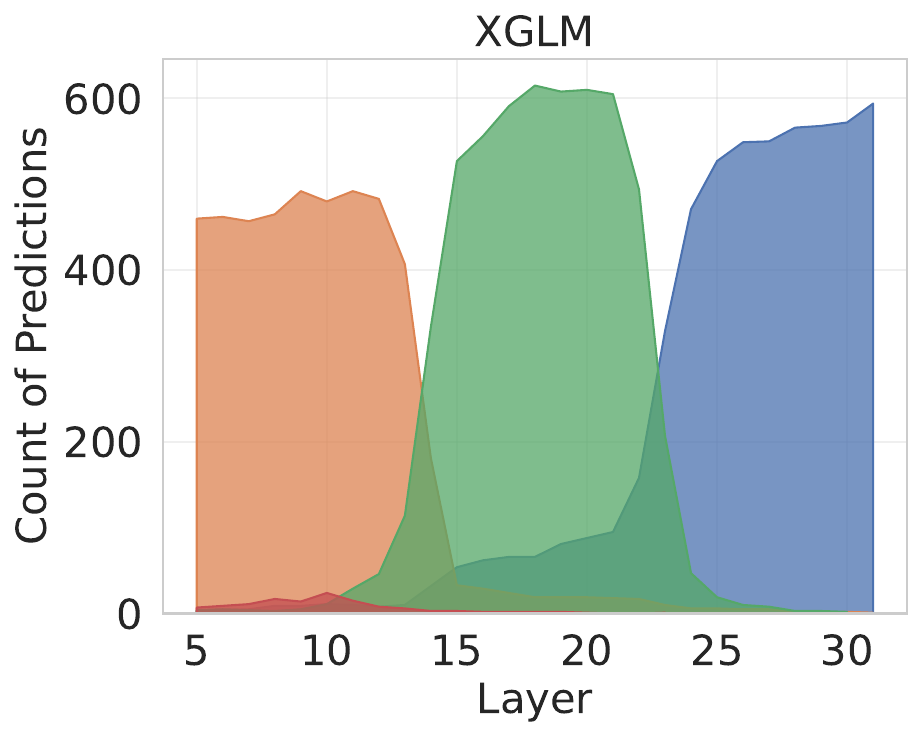}
        \end{subfigure}}
        \hspace{13.5em}
        \raisebox{6.2mm}{
        \begin{subfigure}{0.28\textwidth}
            \includegraphics[width=\linewidth]{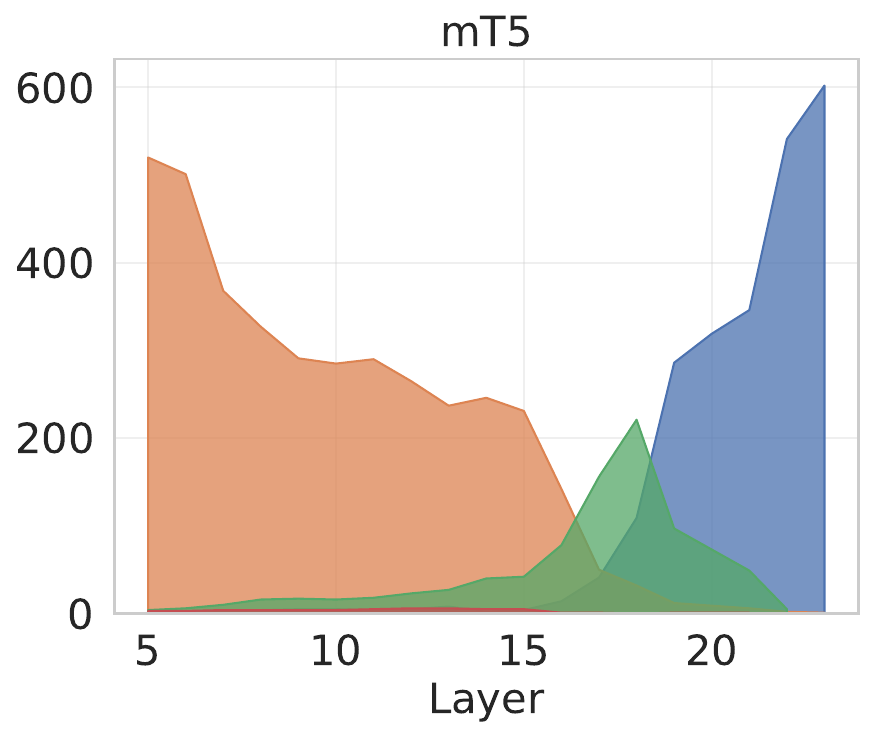}
        \end{subfigure}}
        \vspace{-2mm}
        \caption{Number of examples where the patch produces one of the four valid objects. The green curve (middle) represents layers where the last token's representation encodes the relation but not yet the object.}
        \label{fig:en_all_diff_results}
    \end{subfigure}
    \par\medskip
    \begin{subfigure}{\textwidth}
        \centering
        \begin{subfigure}{0.3\textwidth}
            \includegraphics[width=\linewidth]{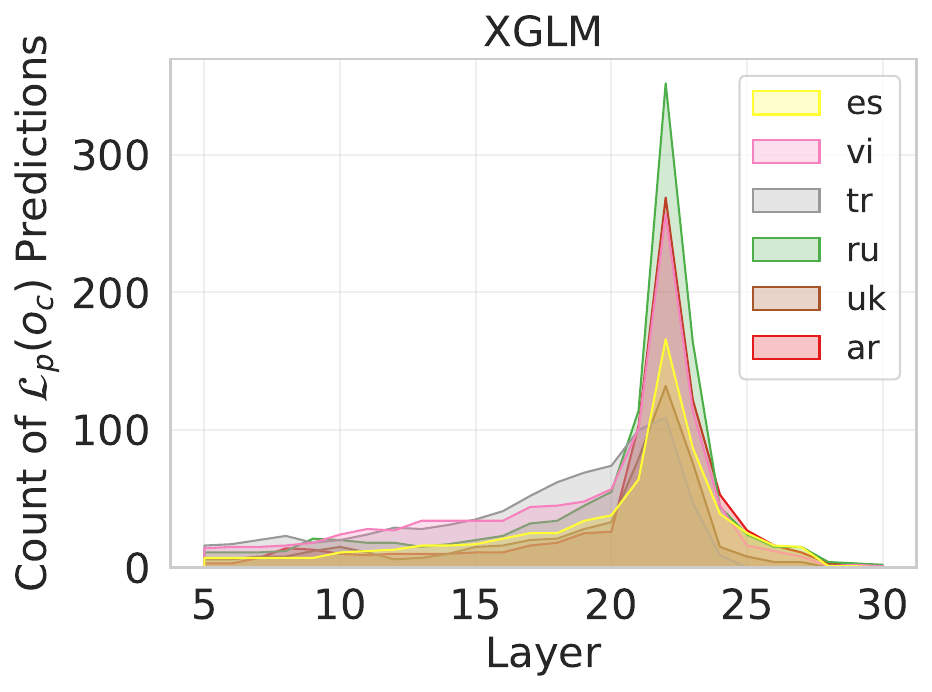}
        \end{subfigure}
        \begin{subfigure}{0.3\textwidth}
            \includegraphics[width=\linewidth]{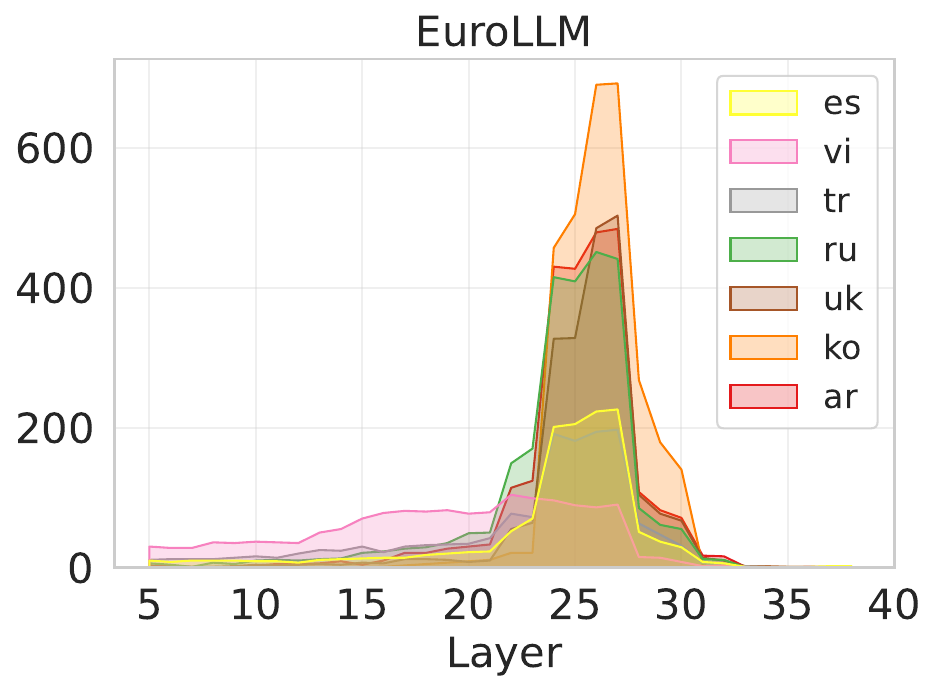}
        \end{subfigure}
        \begin{subfigure}{0.3\textwidth}
            \includegraphics[width=\linewidth]{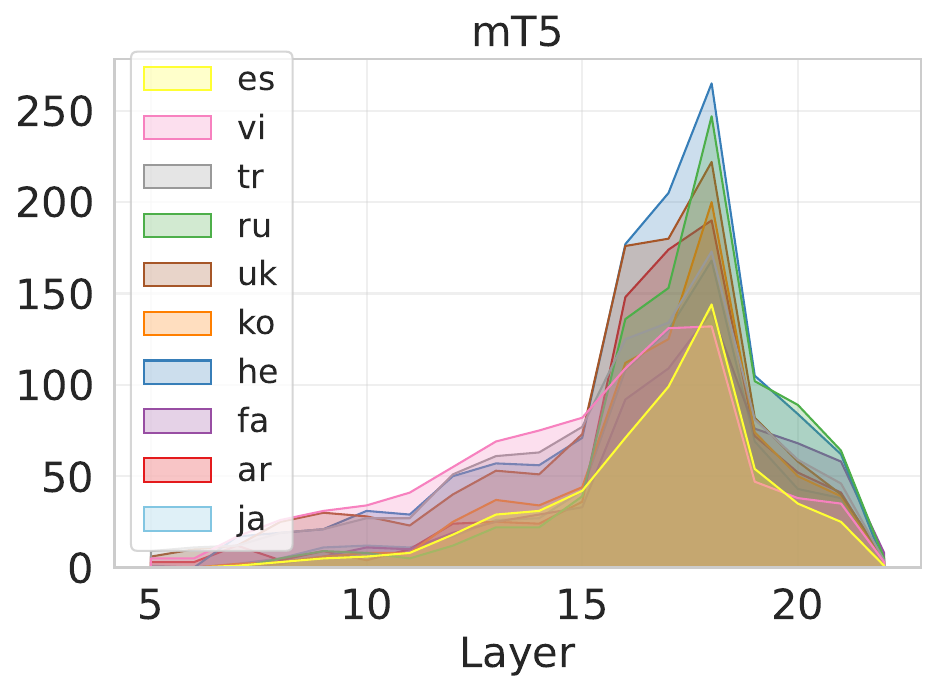}
        \end{subfigure}
        \vspace{-2mm}
        \caption{Number of examples where the patch causes the output to be \(\mathcal{L}_p(o_c)\) (the context object in the patch language). The last token's representation at these layers encodes the relation to extract \textit{and} in which language, but not yet the object.}
        \label{fig:same_r_diff_s_layers}
    \end{subfigure}
    
    \vspace{-1mm}
    \caption{Patches per layer triggering a specific object prediction. (a) Setup \(\{= \mathcal{L}, \neq r, \neq s\}\); (b) \(\{\neq \mathcal{L}, = r, \neq s\}\).}
    \label{fig:combined}
    \vspace{-2mm}
\end{figure*}

Let \(h^l_t\) be the representation at layer \(l\) of the \textit{last} token in the input. For a given input \(p\), we take \(h^l_{t,p}\) and patch it into the forward pass of another input \(c\) at the same layer; that is, we set \(h^l_{t,c} \leftarrow h^l_{t,p}\), and then continue the forward pass. We refer to the example \(p\) as the \textit{patch} example, and \(c\) as the \textit{context} example ( Figure~\ref{fig:patch_explanation}). By patching the last token representation in each layer and observing the model's predicted output, we can study when the representation contains the information about relation, language, and predicted object, and how cross-lingual these representations are. In all experiments, English is used as the patch language.

An input example expresses a relation \(r\) and a subject \(s\) in language \(\mathcal{L}\), which we note \((\mathcal{L}(r), \mathcal{L}(s))\). We conduct three experiments where the \(p\) and \(c\) examples share certain input characteristics: (1)~\(\{= \mathcal{L}, \neq r, \neq s\}\), the language is the same but the relation and subject differ; (2)~\(\{\neq \mathcal{L}, = r, \neq s\}\), the relation remains the same but the language and subject differ; and (3)~\(\{\neq \mathcal{L}, \neq r, = s\}\), the subject is the same but the language and relation differ. 
(1) allows us to study how the relation is encoded while controlling for the language, whereas (2)-(3) help us explore how the interaction between relation and language affects the extraction of the object from the subject. In Figure \ref{fig:summary_patching_eurollm} and \ref{fig:summary_patching_xglm} we show the aggregated results of these 3 settings, with setup (1) we localize when the object is resolved (blue), with setup (2) when the relation \textit{and} language are encoded (purple), and with setup (3) when the relation is encoded (green).

Let the patch input be \((\mathcal{L}_p(r_p), \mathcal{L}_p(s_p))\) and the context input be \((\mathcal{L}_c(r_c), \mathcal{L}_c(s_c))\). Without intervention, the model correctly predicts \(\mathcal{L}_p(o_{r_p,s_p})\) and \(\mathcal{L}_c(o_{r_c,s_c})\) respectively.\footnote{For simplicity, we may refer to the object as \(\mathcal{L}_p(o_p)\)  or \(\mathcal{L}_c(o_c)\) when \(r\) and \(s\) are from the same input.} To analyze patching effects, we examine both output probabilities and predicted tokens. For probabilities, we aggregate across examples by calculating the change relative to the original (unpatched) probability, as in \S\ref{sec:geva_analyses}. For predicted tokens, we check if they match the patch or context object, or a variant with swapped language or relation (e.g., if \(\mathcal{L}_p(o_c)\) is predicted in the setups (2)-(3)).\footnote{We only consider valid patch-context pairs where \(o_{r_p,s_c}\) and \(o_{r_c,s_p}\) exist, and for predicted token analysis, we enforce distinct spellings, e.g. to claim the predicted token is \(t=\mathcal{L}_p(o_c)\) we require \(t\neq\mathcal{L}_c(o_c)\), as languages may share spellings (e.g., ``Asia'' in English and Spanish). Consequently, the number of examples varies across analyses of output probabilities and token predictions.
See Table~\ref{appendix:table_counts_same_r} and \ref{appendix:table_counts_diff_r} for the number of examples for each model and experiment.}

\begin{figure*}[t]
\vspace{-5mm}
    \centering
    \begin{minipage}{0.3\textwidth}
        \centering
        \includegraphics[width=\linewidth]{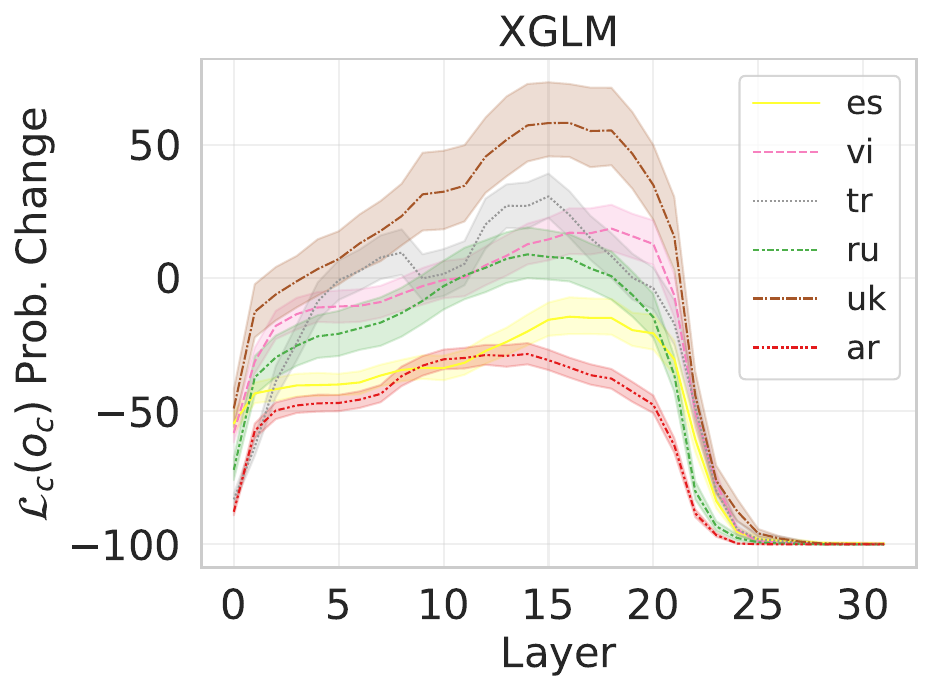}
    \end{minipage}\hfill
    \begin{minipage}{0.3\textwidth}
        \centering
        \includegraphics[width=\linewidth]{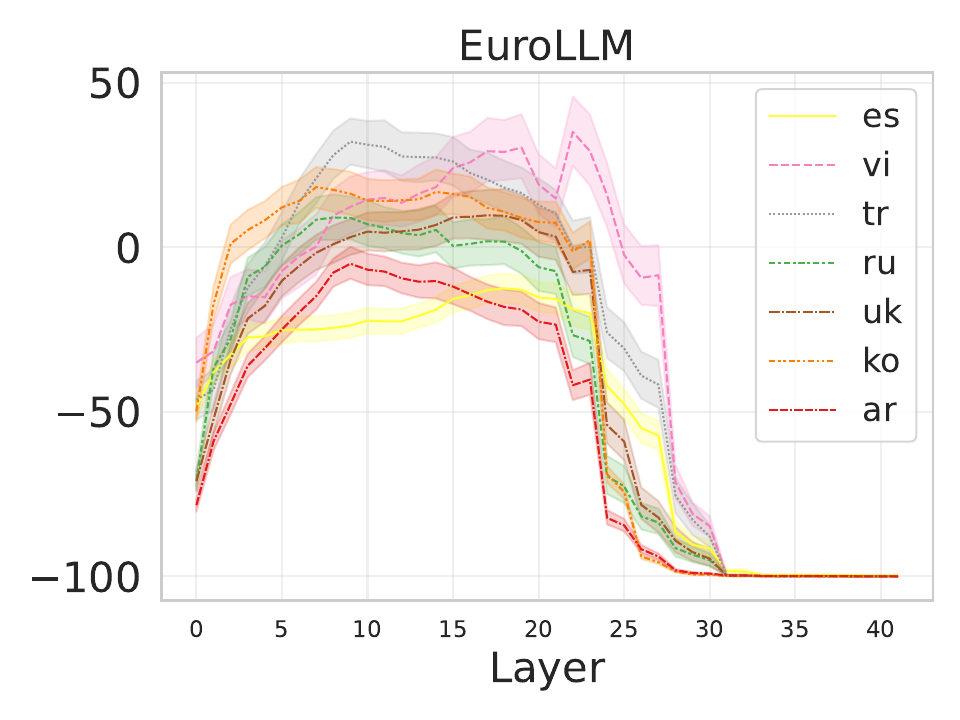}
    \end{minipage}\hfill
    \begin{minipage}{0.3\textwidth}
        \centering
        \includegraphics[width=\linewidth]{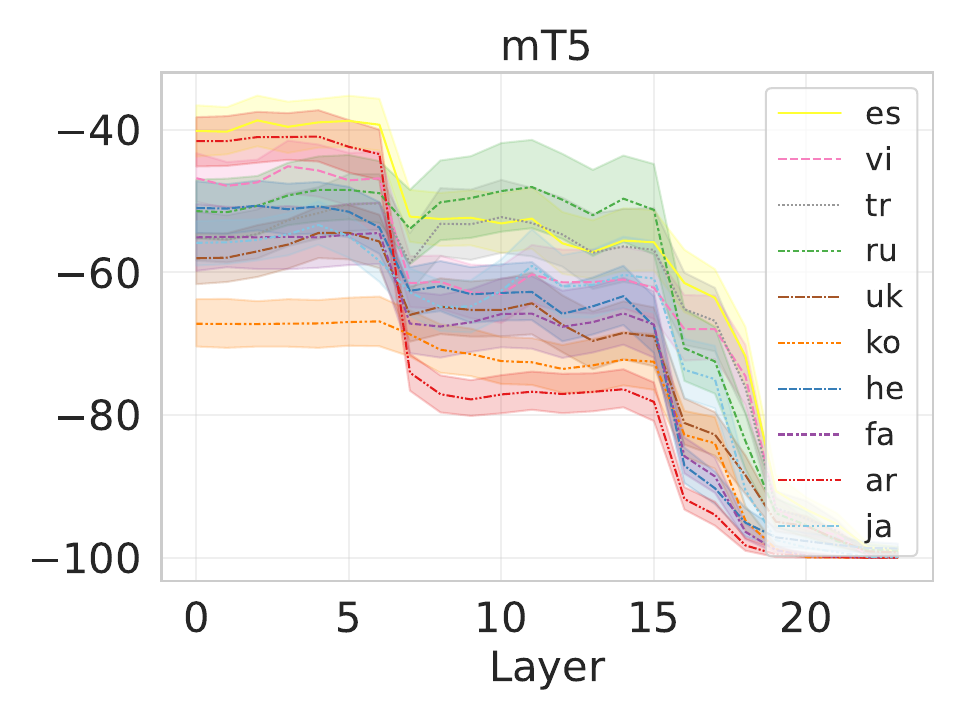}
    \end{minipage}\vspace{-3mm}\caption{Percentage change in the probability of the context object \(\mathbb{P}(\mathcal{L}_c(o_c))\) when patching \(\{\neq \mathcal{L}, =r, \neq s\}\).}\label{fig:same_r_diff_s_probabilities}
    
    \par\medskip
    
    \begin{minipage}{0.33\textwidth}
        \centering
        \includegraphics[width=\linewidth]{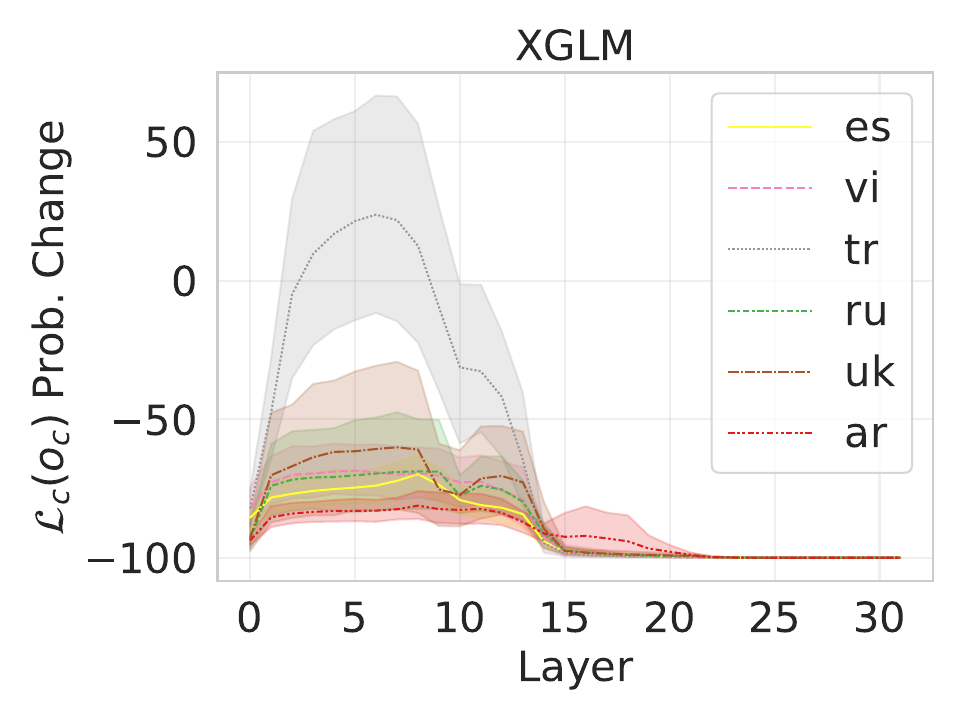}
    \end{minipage}\hfill
    \begin{minipage}{0.33\textwidth}
        \centering
        \includegraphics[width=\linewidth]{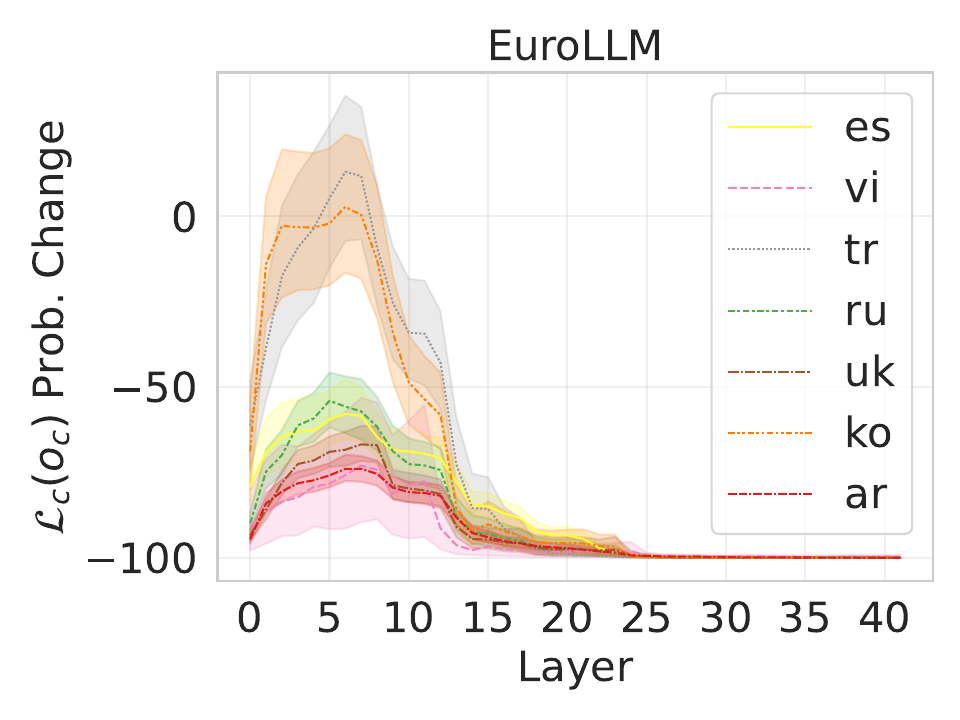}
    \end{minipage}\hfill
    \begin{minipage}{0.33\textwidth}
        \centering
        \includegraphics[width=\linewidth]{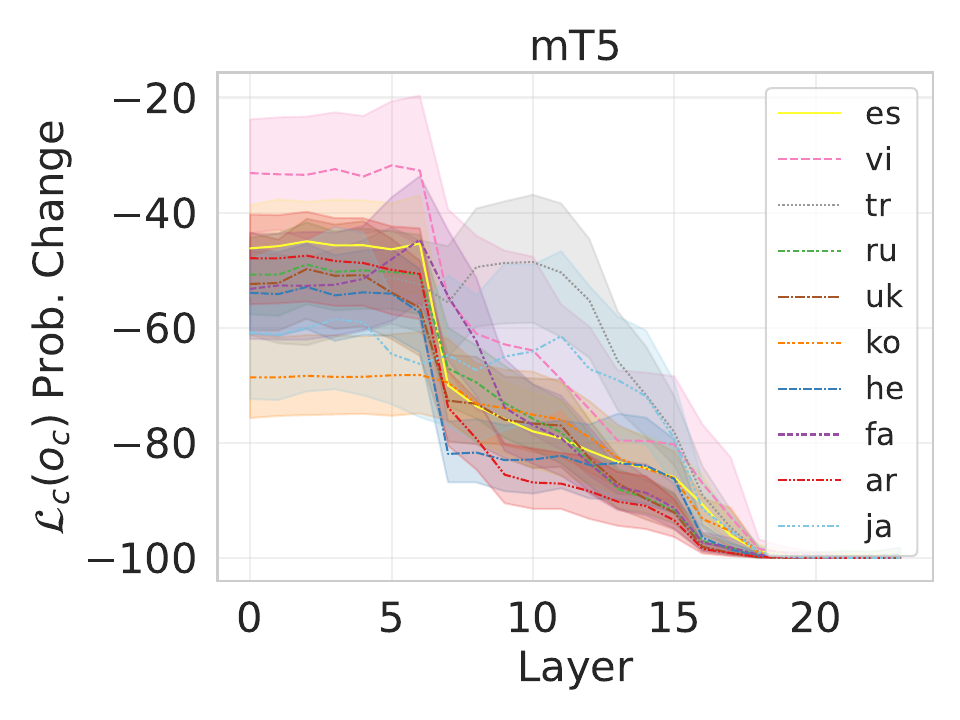}
    \end{minipage}\vspace{-3mm}\caption{Percentage change in the probability of the context object \(\mathbb{P}(\mathcal{L}_c(o_c))\) when patching \(\{\neq \mathcal{L}, \ne r, =s\}\).}\label{fig:diff_r_same_s_probabilities}
    \vspace{-2mm}
\end{figure*}

\paragraph{Different Relation, Different Subject}

First, we analyze the pairs of examples with \(\{= \mathcal{L}, \neq r, \neq s\}\). For example, \(r_p,s_p=\)``The capital of France is'' and \(r_c,s_c=\)``The language spoken in Germany is''. Then, \(o_{r_p,s_c}=\) Berlin and \(o_{r_c,s_p}=\) French. 
We sample 1000 examples for which \mpararel{} has the objects \(o_{r_c,s_p}\) and \(o_{r_p,s_c}\).

We present the prediction results in Figure~\ref{fig:en_all_diff_results} and the  probability plots in Figure~\ref{fig:en_all_diff_results_probs}.
An increase in the green curve indicates that relation information becomes available in the last token representation, as patching at that layer produces \(o_{r_p,s_c}\)---the object corresponding to the relation in the patch example. When the green curve declines and the blue curve rises, the object has been fully extracted and encoded in the last token, since predictions now yield \(o_{r_p,s_p}\) and are no longer influenced by \(c\). These transitions align with peaks in the extraction rate for English (Figure \ref{fig:xglm_extraction_rates_per_language}-\ref{fig:mt5_extraction_rates_per_language}). This confirms that extraction, as measured by the vocabulary projection, marks the point where the object is both available and encoded in the last token representation. If the object had been encoded earlier and only decoded at the extraction point, \(o_{r_p,s_p}\) would have been predicted from earlier-layer patches.

\paragraph{Same Relation, Different Subject}
We have localized the layers where the relation representation flows to the last token. Now, we analyze when the language of the input text propagates to the last token. Is the language information entangled to the subject or the relation representation? Here, we study pairs of patch-context with \(\{\neq \mathcal{L}, = r, \neq s\}\), e.g., ``France's capital is'' and ``La capital de Alemania es'' (Gloss: ``The capital of Germany is'').
If we obtain \(\mathcal{L}_p(o_c)\) (``Berlin''), it would suggest that the language is encoded in the last token representation before the extraction happens. On the other hand, if we obtain \(\mathcal{L}_c(o_p)\) (``Par\'{i}s''), we could infer that the language is encoded after the extraction.

We present the probability of the context answer token \(\mathcal{L}_c(o_c)\) in Figure~\ref{fig:same_r_diff_s_probabilities}.\footnote{The probability of \(\mathcal{L}_p(o_p)\) and per language plots are provided in Appendix~\ref{appendix:same_r_different_s}.}
We observe that for decoder-only models, patching in the early layers generally hurts the model's performance for most languages, while, patching in the middle layers either increases the probability or results in only a minimal decrease.
Given that the relation is the same in both examples, this suggests that by these middle layers, $r$ is encoded in \(h^l_{p}\), and in the subsequent layers the subject and language of the context are integrated to yield the final prediction, \(\mathcal{L}_p(r) + \mathcal{L}_c + s_c = \mathcal{L}_c(o_c)\).
 Moreover, for some languages the relation representation from the patch is better than the one constructed using the context input, as the relative probability is positive. In the case of \mtfive{}, however, the probability of \(\mathcal{L}_c(o_c)\) consistently decreases, plateauing in the middle layers before dropping to zero.
From the attention knockout analysis, recall that in \mtfive{} the subject is integrated into the last token representation only after layer 15, while the relation is consistently represented throughout the middle layers.  
Therefore, these patching results imply that the relation encoded in the last token from the patch input does not help in retrieving the correct context object in the context language in \mtfive{}, whereas it proves useful in \xglm{} and \eurollm{}.

In terms of predicted tokens, we find across-the-board that \(\mathcal{L}_p(o_c)\) is frequently predicted while \(\mathcal{L}_c(o_p)\) is not (Table~\ref{appendix:table_swapped_counts_same_r}). This suggests that the last token representation encodes the patch language \(\mathcal{L}_p\) but not yet the object, as the object is derived from the context subject. We plot the layers where \(\mathcal{L}_p(o_c)\) is predicted in 
Figure~\ref{fig:same_r_diff_s_layers} (\(\mathcal{L}_c(o_p)\) in Figure \ref{fig:same_r_diff_s_layers_appendix}). We can conclude that the language information flows to the last token right before the peak of \(\mathcal{L}_p(o_c)\) predictions, because if we patch earlier, the output is in the context language (see the probabilities of \(\mathcal{L}_c(o_c)\) in Figure \ref{fig:same_r_diff_s_probabilities}). Moreover, these peaks match the beginning of their corresponding extraction phases, thus the language information flows right before the extraction phase.

As a result, we interpret the last token representation as containing a Function Vector (FV), the relation that needs to be extracted and in which language, which is used in the extraction event. The FV can be transferred to contexts in another language, as the FV is constructed from the patch input and is used in the context subject representation to predict \(\mathcal{L}_p(o_c)\).

\paragraph{Different Relation, Same Subject}
We just saw that for all models the representation from the patch will encode at some point the output language but not yet the object. It could be that we observed the prediction \(\mathcal{L}_p(o_c)\) because the relation is language specific and encodes the output language. To analyze if this is the case, we now apply patching on examples with different languages and relations but the same subject \(\{\neq \mathcal{L}, \neq r, = s\}\), e.g., ``France's capital is'' and ``El idioma oficial de Francia es'' (Gloss: ``The official language of France is'').

We observe that the probability of \(\mathcal{L}_c(o_c)\) (Figures ~\ref{fig:diff_r_same_s_probabilities}) in decoder-only models, unlike the previous experiment, decreases early for all languages (except \textit{tr} and \textit{ko}), plateaus around the middle layers, and then drops to zero by the mid-layer range.
For \mtfive{}, the probability drops at the beginning but, instead of plateauing as before, it continues to decline until it reaches zero. 
When compared to the former experiment (Figure \ref{fig:same_r_diff_s_probabilities}), this suggests that the relation information is encoded in the middle layers, as \(\mathcal{L}_c(o_c)\) decreases earlier when the patch and context have different relations.

\begin{figure}[t]
\centering
\begin{minipage}[t]{0.6\linewidth}
    \centering
    \includegraphics[width=\linewidth]{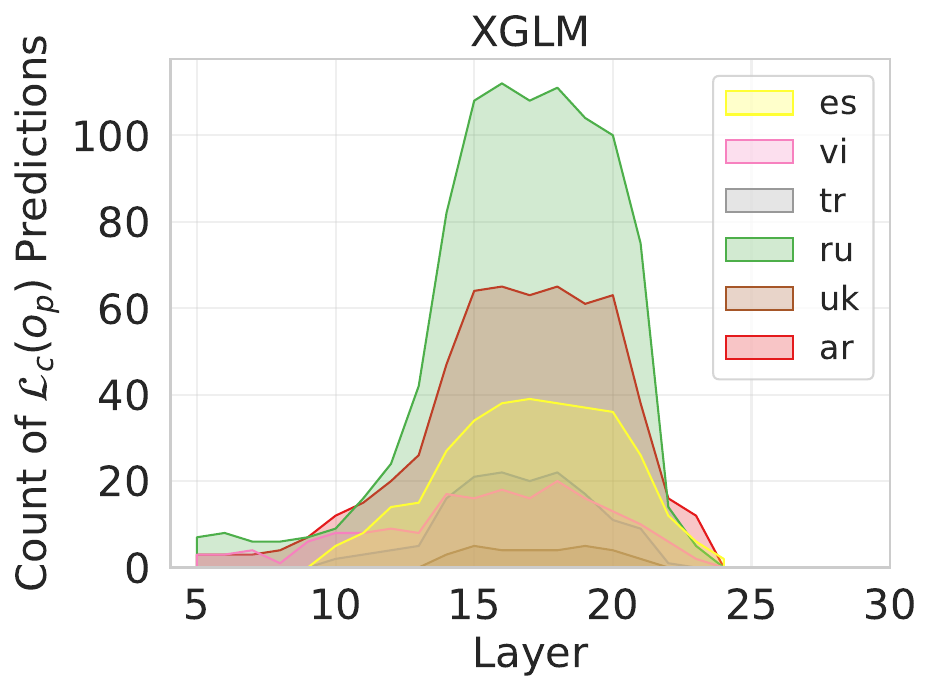}
\end{minipage}
\begin{minipage}[t]{0.6\linewidth}
    \centering
    \includegraphics[width=\linewidth]{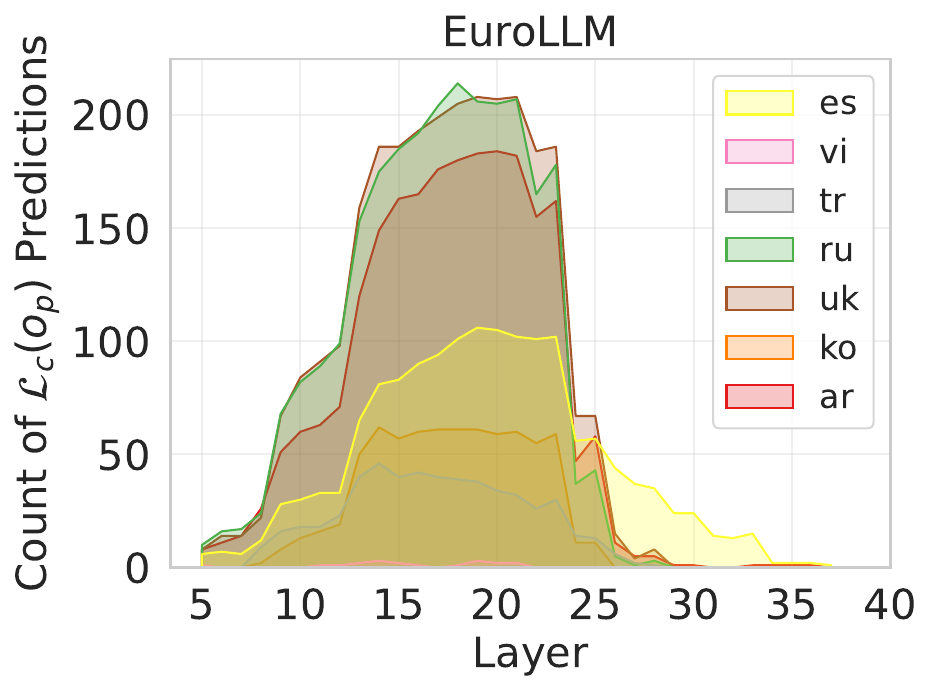}
\end{minipage}
\vspace{-3mm}
  \caption{Patches that cause the prediction to be \(\mathcal{L}_c(o_p)\), which shows the layers where the last token representation contains the relation but not yet the language.}\label{fig:diff_r_same_s_layers}
  \vspace{-4mm}
\end{figure}

In terms of predictions, we find that \(\mathcal{L}_c(o_p)\) is frequently predicted for \xglm{} and \eurollm{} (Figure \ref{fig:diff_r_same_s_layers}), while \(\mathcal{L}_p(o_c)\) appears but less often (Table~\ref{appendix:table_swapped_counts_diff_r}). In line with the previous observation, the plot shows that the last token representation in the middle layers where \(\mathcal{L}_c(o_p)\) is predicted, primarily captures the relation from the patch \(r_p\), without yet encoding the output language or subject information (as these are taken from the context).
Therefore, in decoder-only models, the relation and language representations are disentangled, as the relation flows to the last token before the output language does. Allowing the relation to be combined with different languages.

As for \mtfive{}, we observe very few examples where \(\mathcal{L}_c(o_p)\) or \(\mathcal{L}_p(o_c)\) are predicted, which aligns with the findings of the two former experiments, where we see that the relation is encoded in the last token in layers 15-21 (Figure \ref{fig:en_all_diff_results}), and the language flows to the last token around layer 15-19 (Figure \ref{fig:same_r_diff_s_layers}). We conclude that both the relation and language flow to the last token around the same time, and thus, in this experiment, we cannot see a disentangled behavior. This presents an interesting contrast with decoder-only models. The decoder in \mtfive{} has access to the \textit{same} encoder representations throughout all its layers, so it does not need (and thus does not learn) to attend to these earlier or in different stages.
By contrast, in a decoder-only model, the last token has access to representations that evolve across layers, so it learns to attend to relevant information when it becomes most salient.
Nonetheless, we cannot reach a definite conclusion on whether the language representation in \mtfive{} is entangled or not to the relation representation. More detailed analysis of what is being attended when the relation flows and when the language flows should be performed in future work.

\section{Conclusion}
In this paper, we analyzed factual knowledge recall mechanisms in 10 languages using multilingual transformer-based LMs, comparing them to recall in English autoregressive LLMs.
We discovered that some mechanisms, such as the flow of subject representations in the later layers and the extraction phase, are present in multilingual and monolingual models. 
However, we also identified notable differences, including the join role of late MLPs and attention modules during the extraction phase in multilingual models.
A key contribution of our work is the first-ever investigation of language encoding during recall, achieved through patching representations. In decoder-only models, the relation flows to the last token first, followed by the language. In contrast, in \mtfive{}, both relation and language flow to the last token at similar layers. This suggests that while relation and subject representations are multilingual and enable cross-lingual object extraction, the extraction phase itself is language-specific, as language encoding precedes extraction.
These findings provide new evidence to understand the factual knowledge recall in transformer LMs, and to how decoder-only LMs resolve tasks in stages. Contributing with new directions for the study of cross-lingual transfer and knowledge localization.

\section*{Limitations}
In this paper, we examined three model architectures, leaving out the effects of model sizes, instruction fine-tuning, or models like Llama that can behave multilingually but have less coverage and less multilingual pre-training data. Additionally, our analyses were conducted on 500-1000 examples per language, which we believe provides a sufficient sample size for generalization; however, the results are inherently limited by the relations present in the \mpararel{} dataset, which may not capture all factual nuances. Additionally, although we analyzed 10 diverse languages, many more languages exist, and further research is needed to confirm the generalizability of our findings across a broader linguistic spectrum. Lastly, we described the main mechanisms found in \xglm{}, \eurollm{} and \mtfive{}, however other weaker mechanisms could be at play, which could describe, for example, the low extraction rates found for some languages (Figure \ref{fig:mt5_extraction_rates_per_language}) or the few examples where the object seems to be encoded from early layers before the extraction takes place (Figure \ref{fig:same_r_diff_s_layers_appendix}).

\section*{Acknowledgments}
We thank our colleagues at the CoAStaL NLP group Laura Cabello, Rita Ramos, and Israfel Salazar for valuable comments on the final manuscript.
DE was supported by the European Union’s Horizon 2020 research and innovation program under grant agreement No.\ 101135671 (TrustLLM). 

\bibliography{anthology,custom}

\begin{thebibliography}{33}
\providecommand{\natexlab}[1]{#1}

\bibitem[{Brown et~al.(2020)Brown, Mann, Ryder, Subbiah, Kaplan, Dhariwal, Neelakantan, Shyam, Sastry, Askell et~al.}]{brown2020language}
Tom Brown, Benjamin Mann, Nick Ryder, Melanie Subbiah, Jared~D Kaplan, Prafulla Dhariwal, Arvind Neelakantan, Pranav Shyam, Girish Sastry, Amanda Askell, et~al. 2020.
\newblock Language models are few-shot learners.
\newblock \emph{Advances in neural information processing systems}, 33:1877--1901.

\bibitem[{Chughtai et~al.(2024)Chughtai, Cooney, and Nanda}]{chughtai2024summing}
Bilal Chughtai, Alan Cooney, and Neel Nanda. 2024.
\newblock Summing up the facts: Additive mechanisms behind factual recall in llms.
\newblock \emph{arXiv preprint arXiv:2402.07321}.

\bibitem[{Dumas et~al.(2024)Dumas, Veselovsky, Monea, West, and Wendler}]{dumas2024how}
Cl{\'e}ment Dumas, Veniamin Veselovsky, Giovanni Monea, Robert West, and Chris Wendler. 2024.
\newblock \href {https://openreview.net/forum?id=0ku2hIm4BS} {How do llamas process multilingual text? a latent exploration through activation patching}.
\newblock In \emph{ICML 2024 Workshop on Mechanistic Interpretability}.

\bibitem[{Elazar et~al.(2021)Elazar, Kassner, Ravfogel, Ravichander, Hovy, Sch{\"u}tze, and Goldberg}]{elazar-etal-2021-measuring}
Yanai Elazar, Nora Kassner, Shauli Ravfogel, Abhilasha Ravichander, Eduard Hovy, Hinrich Sch{\"u}tze, and Yoav Goldberg. 2021.
\newblock \href {https://doi.org/10.1162/tacl_a_00410} {Measuring and improving consistency in pretrained language models}.
\newblock \emph{Transactions of the Association for Computational Linguistics}, 9:1012--1031.

\bibitem[{Ferrando and Costa-juss{\`a}(2024)}]{ferrando-costa-jussa-2024-similarity}
Javier Ferrando and Marta~R. Costa-juss{\`a}. 2024.
\newblock \href {https://doi.org/10.18653/v1/2024.findings-emnlp.591} {On the similarity of circuits across languages: a case study on the subject-verb agreement task}.
\newblock In \emph{Findings of the Association for Computational Linguistics: EMNLP 2024}, pages 10115--10125, Miami, Florida, USA. Association for Computational Linguistics.

\bibitem[{Fierro et~al.(2024)Fierro, Dhar, Stamatiou, S\o{}gaard, and Garneau}]{Fierro2024-FIEDKB}
Constanza Fierro, Ruchira Dhar, Filippos Stamatiou, Anders S\o{}gaard, and Nicolas Garneau. 2024.
\newblock Defining knowledge: Bridging epistemology and large language models.
\newblock \emph{Proceedings of the 2024 Conference on Empirical Methods in Natural Language Processing}, 2024.

\bibitem[{Fierro and S{\o}gaard(2022)}]{fierro-sogaard-2022-factual}
Constanza Fierro and Anders S{\o}gaard. 2022.
\newblock \href {https://doi.org/10.18653/v1/2022.findings-acl.240} {Factual consistency of multilingual pretrained language models}.
\newblock In \emph{Findings of the Association for Computational Linguistics: ACL 2022}, pages 3046--3052, Dublin, Ireland. Association for Computational Linguistics.

\bibitem[{Foroutan et~al.(2022)Foroutan, Banaei, Lebret, Bosselut, and Aberer}]{foroutan-etal-2022-discovering}
Negar Foroutan, Mohammadreza Banaei, R{\'e}mi Lebret, Antoine Bosselut, and Karl Aberer. 2022.
\newblock \href {https://doi.org/10.18653/v1/2022.emnlp-main.513} {Discovering language-neutral sub-networks in multilingual language models}.
\newblock In \emph{Proceedings of the 2022 Conference on Empirical Methods in Natural Language Processing}, pages 7560--7575, Abu Dhabi, United Arab Emirates. Association for Computational Linguistics.

\bibitem[{Geva et~al.(2023)Geva, Bastings, Filippova, and Globerson}]{geva-etal-2023-dissecting}
Mor Geva, Jasmijn Bastings, Katja Filippova, and Amir Globerson. 2023.
\newblock \href {https://doi.org/10.18653/v1/2023.emnlp-main.751} {Dissecting recall of factual associations in auto-regressive language models}.
\newblock In \emph{Proceedings of the 2023 Conference on Empirical Methods in Natural Language Processing}, pages 12216--12235, Singapore. Association for Computational Linguistics.

\bibitem[{Geva et~al.(2021)Geva, Schuster, Berant, and Levy}]{geva-etal-2021-transformer}
Mor Geva, Roei Schuster, Jonathan Berant, and Omer Levy. 2021.
\newblock \href {https://doi.org/10.18653/v1/2021.emnlp-main.446} {Transformer feed-forward layers are key-value memories}.
\newblock In \emph{Proceedings of the 2021 Conference on Empirical Methods in Natural Language Processing}, pages 5484--5495, Online and Punta Cana, Dominican Republic. Association for Computational Linguistics.

\bibitem[{Ghandeharioun et~al.(2024)Ghandeharioun, Caciularu, Pearce, Dixon, and Geva}]{ghandeharioun2024patchscopes}
Asma Ghandeharioun, Avi Caciularu, Adam Pearce, Lucas Dixon, and Mor Geva. 2024.
\newblock \href {https://openreview.net/forum?id=5uwBzcn885} {Patchscopes: A unifying framework for inspecting hidden representations of language models}.
\newblock In \emph{Forty-first International Conference on Machine Learning}.

\bibitem[{Grasswick(2010)}]{Grasswick2010-GRASAL-2}
Heidi~E. Grasswick. 2010.
\newblock \href {https://doi.org/10.1007/s11229-010-9789-0} {Scientific and lay communities: Earning epistemic trust through knowledge sharing}.
\newblock \emph{Synthese}, 177(3):387--409.

\bibitem[{Gu and Dao(2023)}]{gu2023mamba}
Albert Gu and Tri Dao. 2023.
\newblock Mamba: Linear-time sequence modeling with selective state spaces.
\newblock \emph{arXiv preprint arXiv:2312.00752}.

\bibitem[{Hawley(2012)}]{hawleytrust2012}
Katherine Hawley. 2012.
\newblock \href {https://doi.org/10.1093/actrade/9780199697342.003.0007} {{64Knowledge and expertise}}.
\newblock In \emph{{Trust: A Very Short Introduction}}. Oxford University Press.

\bibitem[{Jiang et~al.(2020)Jiang, Anastasopoulos, Araki, Ding, and Neubig}]{jiang-etal-2020-x}
Zhengbao Jiang, Antonios Anastasopoulos, Jun Araki, Haibo Ding, and Graham Neubig. 2020.
\newblock \href {https://doi.org/10.18653/v1/2020.emnlp-main.479} {{X}-{FACTR}: Multilingual factual knowledge retrieval from pretrained language models}.
\newblock In \emph{Proceedings of the 2020 Conference on Empirical Methods in Natural Language Processing (EMNLP)}, pages 5943--5959, Online. Association for Computational Linguistics.

\bibitem[{Kassner et~al.(2021)Kassner, Dufter, and Sch{\"u}tze}]{kassner-etal-2021-multilingual}
Nora Kassner, Philipp Dufter, and Hinrich Sch{\"u}tze. 2021.
\newblock \href {https://doi.org/10.18653/v1/2021.eacl-main.284} {Multilingual {LAMA}: Investigating knowledge in multilingual pretrained language models}.
\newblock In \emph{Proceedings of the 16th Conference of the European Chapter of the Association for Computational Linguistics: Main Volume}, pages 3250--3258, Online. Association for Computational Linguistics.

\bibitem[{Li et~al.(2024)Li, Ji, Mickus, Segonne, and Tiedemann}]{li-etal-2024-comparison}
Zihao Li, Shaoxiong Ji, Timothee Mickus, Vincent Segonne, and J{\"o}rg Tiedemann. 2024.
\newblock \href {https://doi.org/10.18653/v1/2024.emnlp-main.888} {A comparison of language modeling and translation as multilingual pretraining objectives}.
\newblock In \emph{Proceedings of the 2024 Conference on Empirical Methods in Natural Language Processing}, pages 15882--15894, Miami, Florida, USA. Association for Computational Linguistics.

\bibitem[{Lin et~al.(2021)Lin, Mihaylov, Artetxe, Wang, Chen, Simig, Ott, Goyal, Bhosale, Du et~al.}]{lin2021few}
Xi~Victoria Lin, Todor Mihaylov, Mikel Artetxe, Tianlu Wang, Shuohui Chen, Daniel Simig, Myle Ott, Naman Goyal, Shruti Bhosale, Jingfei Du, et~al. 2021.
\newblock Few-shot learning with multilingual language models.
\newblock \emph{arXiv preprint arXiv:2112.10668}.

\bibitem[{Martins et~al.(2024)Martins, Fernandes, Alves, Guerreiro, Rei, Alves, Pombal, Farajian, Faysse, Klimaszewski et~al.}]{martins2024eurollm}
Pedro~Henrique Martins, Patrick Fernandes, Jo{\~a}o Alves, Nuno~M Guerreiro, Ricardo Rei, Duarte~M Alves, Jos{\'e} Pombal, Amin Farajian, Manuel Faysse, Mateusz Klimaszewski, et~al. 2024.
\newblock Eurollm: Multilingual language models for europe.
\newblock \emph{arXiv preprint arXiv:2409.16235}.

\bibitem[{Meng et~al.(2022)Meng, Bau, Andonian, and Belinkov}]{meng2022locating}
Kevin Meng, David Bau, Alex~J Andonian, and Yonatan Belinkov. 2022.
\newblock \href {https://openreview.net/forum?id=-h6WAS6eE4} {Locating and editing factual associations in {GPT}}.
\newblock In \emph{Advances in Neural Information Processing Systems}.

\bibitem[{Nguyen(2022)}]{Nguyen2022-NGUTAA}
C.~Thi Nguyen. 2022.
\newblock Trust as an unquestioning attitude.
\newblock \emph{Oxford Studies in Epistemology}, 7:214--244.

\bibitem[{Pearl(2022)}]{pearl2022direct}
Judea Pearl. 2022.
\newblock Direct and indirect effects.
\newblock In \emph{Probabilistic and causal inference: the works of Judea Pearl}, pages 373--392.

\bibitem[{Petroni et~al.(2019)Petroni, Rockt{\"a}schel, Riedel, Lewis, Bakhtin, Wu, and Miller}]{petroni-etal-2019-language}
Fabio Petroni, Tim Rockt{\"a}schel, Sebastian Riedel, Patrick Lewis, Anton Bakhtin, Yuxiang Wu, and Alexander Miller. 2019.
\newblock \href {https://doi.org/10.18653/v1/D19-1250} {Language models as knowledge bases?}
\newblock In \emph{Proceedings of the 2019 Conference on Empirical Methods in Natural Language Processing and the 9th International Joint Conference on Natural Language Processing (EMNLP-IJCNLP)}, pages 2463--2473, Hong Kong, China. Association for Computational Linguistics.

\bibitem[{Qi et~al.(2023)Qi, Fern{\'a}ndez, and Bisazza}]{qi-etal-2023-cross}
Jirui Qi, Raquel Fern{\'a}ndez, and Arianna Bisazza. 2023.
\newblock \href {https://doi.org/10.18653/v1/2023.emnlp-main.658} {Cross-lingual consistency of factual knowledge in multilingual language models}.
\newblock In \emph{Proceedings of the 2023 Conference on Empirical Methods in Natural Language Processing}, pages 10650--10666, Singapore. Association for Computational Linguistics.

\bibitem[{Sharma et~al.(2024)Sharma, Atkinson, and Bau}]{sharma2024locating}
Arnab~Sen Sharma, David Atkinson, and David Bau. 2024.
\newblock \href {https://openreview.net/forum?id=yoVRyrEgix} {Locating and editing factual associations in mamba}.
\newblock In \emph{First Conference on Language Modeling}.

\bibitem[{Tamayo et~al.(2024)Tamayo, Gonzalez-Agirre, Hernando, and Villegas}]{mela-etal-2024-mass}
Daniel Tamayo, Aitor Gonzalez-Agirre, Javier Hernando, and Marta Villegas. 2024.
\newblock \href {https://doi.org/10.18653/v1/2024.findings-acl.347} {Mass-editing memory with attention in transformers: A cross-lingual exploration of knowledge}.
\newblock In \emph{Findings of the Association for Computational Linguistics: ACL 2024}, pages 5831--5847, Bangkok, Thailand. Association for Computational Linguistics.

\bibitem[{Todd et~al.(2024)Todd, Li, Sharma, Mueller, Wallace, and Bau}]{todd2024function}
Eric Todd, Millicent Li, Arnab~Sen Sharma, Aaron Mueller, Byron~C Wallace, and David Bau. 2024.
\newblock \href {https://openreview.net/forum?id=AwyxtyMwaG} {Function vectors in large language models}.
\newblock In \emph{The Twelfth International Conference on Learning Representations}.

\bibitem[{Wang et~al.(2024)Wang, Whyte, and Xu}]{wang-etal-2024-locating}
Zijian Wang, Britney Whyte, and Chang Xu. 2024.
\newblock \href {https://doi.org/10.18653/v1/2024.findings-acl.287} {Locating and extracting relational concepts in large language models}.
\newblock In \emph{Findings of the Association for Computational Linguistics: ACL 2024}, pages 4818--4832, Bangkok, Thailand. Association for Computational Linguistics.

\bibitem[{Wendler et~al.(2024)Wendler, Veselovsky, Monea, and West}]{wendler-etal-2024-llamas}
Chris Wendler, Veniamin Veselovsky, Giovanni Monea, and Robert West. 2024.
\newblock \href {https://doi.org/10.18653/v1/2024.acl-long.820} {Do llamas work in {E}nglish? on the latent language of multilingual transformers}.
\newblock In \emph{Proceedings of the 62nd Annual Meeting of the Association for Computational Linguistics (Volume 1: Long Papers)}, pages 15366--15394, Bangkok, Thailand. Association for Computational Linguistics.

\bibitem[{Xue et~al.(2021)Xue, Constant, Roberts, Kale, Al-Rfou, Siddhant, Barua, and Raffel}]{xue-etal-2021-mt5}
Linting Xue, Noah Constant, Adam Roberts, Mihir Kale, Rami Al-Rfou, Aditya Siddhant, Aditya Barua, and Colin Raffel. 2021.
\newblock \href {https://doi.org/10.18653/v1/2021.naacl-main.41} {m{T}5: A massively multilingual pre-trained text-to-text transformer}.
\newblock In \emph{Proceedings of the 2021 Conference of the North American Chapter of the Association for Computational Linguistics: Human Language Technologies}, pages 483--498, Online. Association for Computational Linguistics.

\bibitem[{Yin et~al.(2022)Yin, Bansal, Monajatipoor, Li, and Chang}]{yin-etal-2022-geomlama}
Da~Yin, Hritik Bansal, Masoud Monajatipoor, Liunian~Harold Li, and Kai-Wei Chang. 2022.
\newblock \href {https://doi.org/10.18653/v1/2022.emnlp-main.132} {{G}eo{MLAMA}: Geo-diverse commonsense probing on multilingual pre-trained language models}.
\newblock In \emph{Proceedings of the 2022 Conference on Empirical Methods in Natural Language Processing}, pages 2039--2055, Abu Dhabi, United Arab Emirates. Association for Computational Linguistics.

\bibitem[{Zhang and Nanda(2024)}]{zhang2024towards}
Fred Zhang and Neel Nanda. 2024.
\newblock \href {https://openreview.net/forum?id=Hf17y6u9BC} {Towards best practices of activation patching in language models: Metrics and methods}.
\newblock In \emph{The Twelfth International Conference on Learning Representations}.

\bibitem[{Zhang et~al.(2025)Zhang, Yu, Zang, Eickhoff, and Pavlick}]{zhang2025the}
Ruochen Zhang, Qinan Yu, Matianyu Zang, Carsten Eickhoff, and Ellie Pavlick. 2025.
\newblock \href {https://openreview.net/forum?id=NCrFA7dq8T} {The same but different: Structural similarities and differences in multilingual language modeling}.
\newblock In \emph{The Thirteenth International Conference on Learning Representations}.

\end{thebibliography}

\clearpage
\appendix

\begin{table}[t]
\vspace{1em}
\centering\footnotesize\setlength{\tabcolsep}{3pt}
\resizebox{\linewidth}{!}{
\begin{tabular}{lccc}
\toprule
Language & Script & Family & SOV/SVO \\ 
\midrule
English & Latin & Germanic & SVO \\ 
Spanish & Latin & Romance & SVO \\ 
Vietnamese & Latin & Austroasiatic & SVO \\ 
Turkish & Latin & Turkic & SOV \\ 
Russian & Cyrillic & Slavic & SVO* \\ 
Ukrainian & Cyrillic & Slavic & SVO* \\ 
Japanese & Kanji & Proto-Japonic & SOV \\ 
Korean & Korean & Koreanic & SOV \\ 
Hebrew & Hebrew & Arabic & VSO \\ 
Farsi (Persian) & Perso-Arabic & Indo-Iranian & SOV \\ 
Arabic & Arabic & Arabic & VSO \\ 
\bottomrule
\end{tabular}
}
\caption{Languages, their scripts, families, and sentence structures (SVO: subject-verb-object, SOV: subject-object-verb, VSO: verb-subject-object, SVO*: SVO dominant but SOV is also possible).}
\label{tab:languages_chars}
\end{table}

\section{Experimental Setup}\label{appendix:experimental_setup}
We use the \mpararel{} triplets and templates to query the factual knowledge of the language models. As mentioned earlier, we perform 3 modifications to the dataset. First, we fetch WikiData for aliases of the target object to be able to match different possible surface forms. Second, we filter out examples where the target object is contained in the query, e.g. \textit{Microsoft Outlook is developed by}. Finally, to better compare across languages we control the variety of the subject-object pairs, by only using a crosslingual version of \mpararel{}. Specifically, for each relation we filter out triplets that are not present in all the languages (in \mpararel{} a subject and object may have not been translated if they were not found in WikiData). Thus, in the crosslingual \mpararel{} version, each subject-object pair in a relation is present in each of the languages. For \xglm{} we additionally restrict the templates to have the object at the end of the sentence, therefore for some languages (\textit{tr}, \textit{ko}, \textit{ja}, and \textit{fa}) both the autorregressive condition on the templates and the crosslingual restriction leads to too few total examples (< 1000), so for these we do not impose the crosslingual restriction. For all the other languages, and for all the languages in \mtfive{} there are enough examples so we restrict them to be crosslingual. In Table \ref{appendix:memorized_counts} we present the total number of examples we consider per language and model, and the correctly predicted fraction.\footnote{In decoder-only models, the total number of examples is higher in some languages due to crosslingual filtering, which is applied per relation. If a language has no examples for a given relation, it doesn’t restrict examples in other languages. Since decoder-only models use only autoregressive templates, some lower-resource languages may have zero examples for certain relations. However, when using all templates (as in \mtfive{}), these lower-resource languages can restrict the triplets available for other languages within the same relation.}

\begin{table}[t]
\vspace{1em}
\centering\footnotesize\setlength{\tabcolsep}{3pt}
\resizebox{\linewidth}{!}{
\begin{tabular}{lcccccccccccc}
\toprule
& \multicolumn{3}{c}{\xglm{} and \eurollm{}} & \multicolumn{3}{c}{\mtfive{}} \\
\cmidrule(lr){2-4} \cmidrule(lr){5-7}
Language
 & Correct & Total & Percentage & Correct & Total & Percentage \\ \midrule
en & 1812 & 4147 & 43.7\% & 1543 & 3853 & 40.0\% \\
es & 1380 & 4167 & 33.1\% & 1192 & 3926 & 30.4\% \\
vi & 1646 & 4068 & 40.5\% & 993 & 3748 & 26.5\% \\
tr & 418 & 2278 & 18.3\% & 1058 & 4033 & 26.2\% \\
ru & 830 & 4133 & 20.1\% & 683 & 3826 & 17.9\% \\
uk & 213 & 4144 & 5.1\% & 456 & 3830 & 11.9\% \\
ko & 116 & 1444 & 8.0\% & 630 & 3783 & 16.7\% \\
ja & 6 & 1790 & 0.3\% & 358 & 4124 & 8.7\% \\
he & 13 & 4403 & 0.3\% & 565 & 4064 & 13.9\% \\
fa & 7 & 4388 & 0.2\% & 406 & 3814 & 10.6\% \\
ar & 811 & 4482 & 18.1\% & 488 & 4110 & 11.9\% \\
 \bottomrule
\end{tabular}
}
\caption{Total number of examples in \mpararel{}.}\label{appendix:memorized_counts}
\end{table}

\subsection{Prompt Details}\label{appendix:prompt_details}
For \mtfive{}, we feed the \mpararel{} input into the encoder with a sentinel token in the object placeholder.
In the decoder, we provide the beginning-of-sequence token followed by the sentinel token.
We only check the tokens generated next to the first sentinel token, as the pre-training task of the model is to generate the text for each sentinel in the input, the decoder usually continues to generate answers for other sentinel tokens.
Any tokens generated preceding the object, are added  to the decoder input since adding these to the encoder does not ensure that the next token predicted will be the object.

\subsection{Computational Resources}
The experiments were run in A100 GPUs. The causal analysis took from 30 minutes to 12 hours depending on the language. The rest of the experiments took 1-2 hours per language.

\clearpage
\section{Causal Tracing}\label{appendix:causal_tracing}

\begin{minipage}{\textwidth}
    \centering
    \includegraphics[width=0.8\textwidth]{images/causal_analysis/gpt2-xl_avg_logits_w=6.pdf}
    \includegraphics[width=0.8\textwidth]{images/causal_analysis/eurollm_avg_logits_w=5.pdf}
    \includegraphics[width=0.8\textwidth]{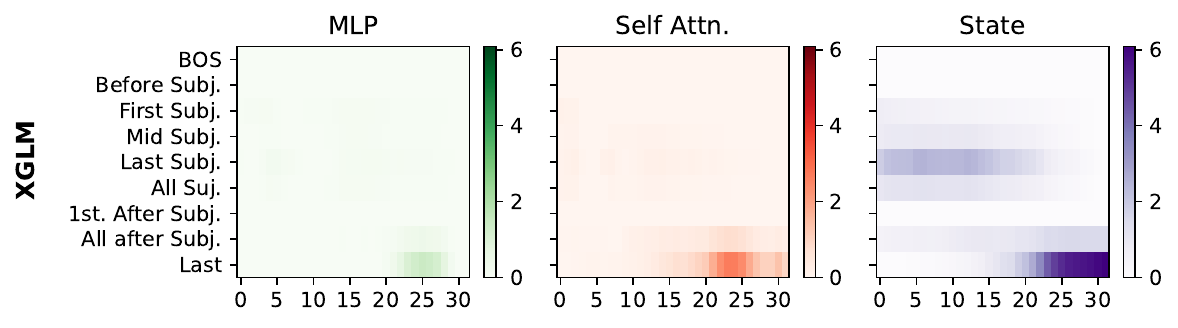}
    \includegraphics[width=\textwidth]{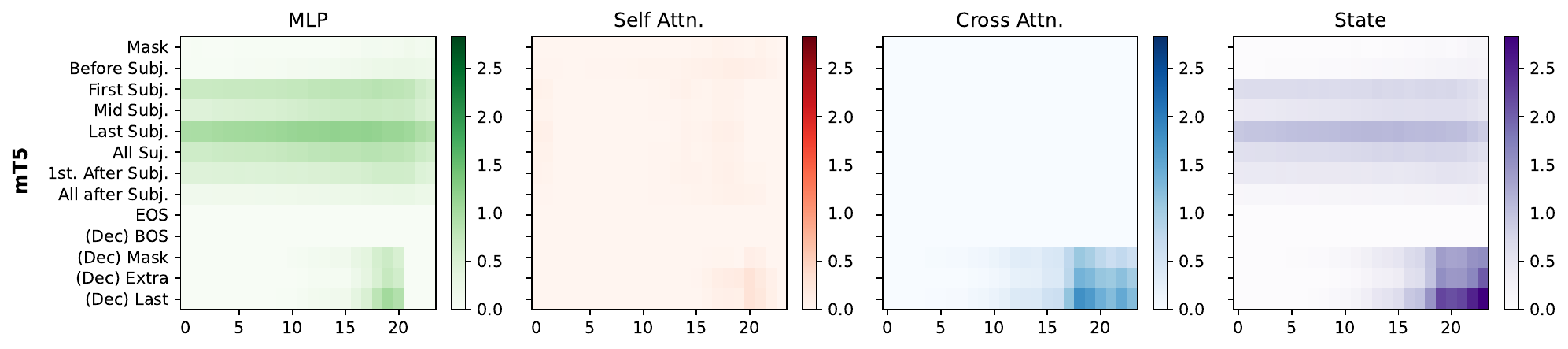}
    \captionof{figure}{Average indirect effect on logit scores, when the subject input has been corrupted and some activations are restored to their uncorrupted values. The MHSA and MLP are restored in windows of 12\% consecutive layers. Top GPT2-XL (only English data).}
    \label{fig_appendix:avg_causal_analysis}
    \vspace{-4mm}
\end{minipage}

\clearpage
\subsection{\xglm{}}
\begin{figure*}[h!]
    \centering
    \includegraphics[width=0.8\linewidth]{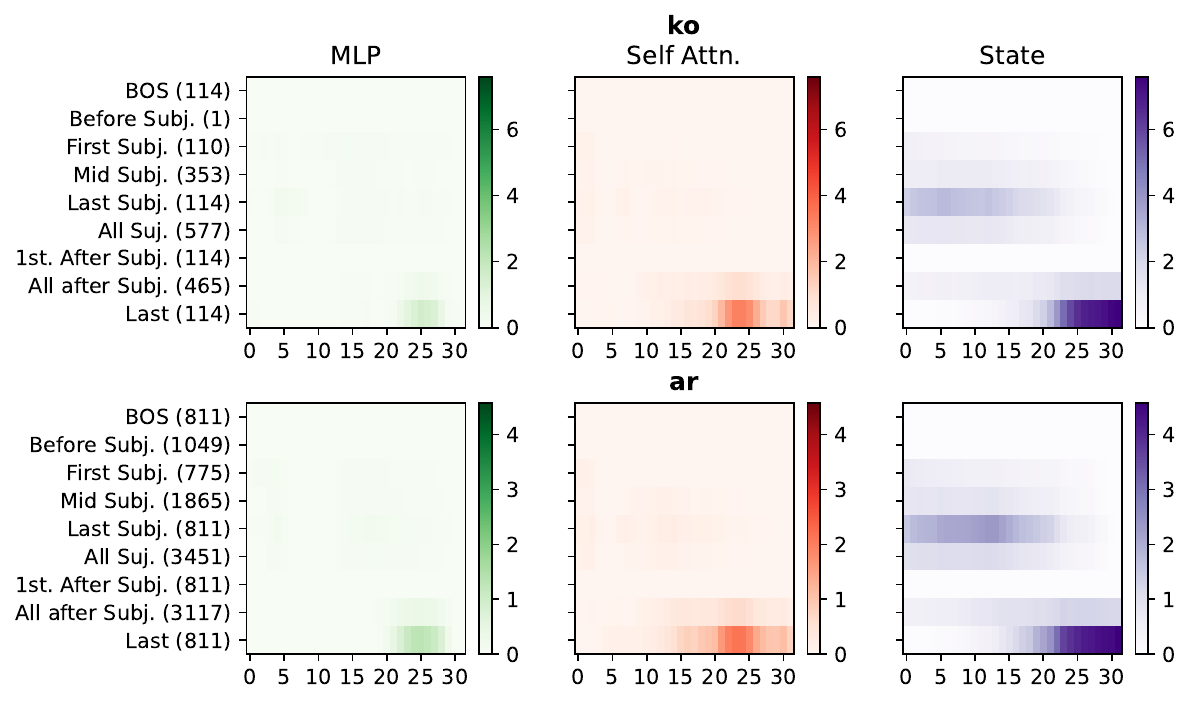} 
    \vspace{-3mm}
    \caption{\xglm{} causal analysis for each language (Rest of the languages in Figure~\ref{fig:xglm_causal_analysis_logits_1}). Average \textbf{logit score} (for the originally predicted token) recovered after corrupting the subject in the input and restoring: the hidden representation at a given layer (State), the MLP in a window of 4 layers, or the Self Attn. in a window of 4 layers.}
    \label{fig:xglm_causal_analysis_logits_2}
\end{figure*}

\begin{minipage}{\textwidth}
    \centering
    \includegraphics[width=0.8\linewidth]{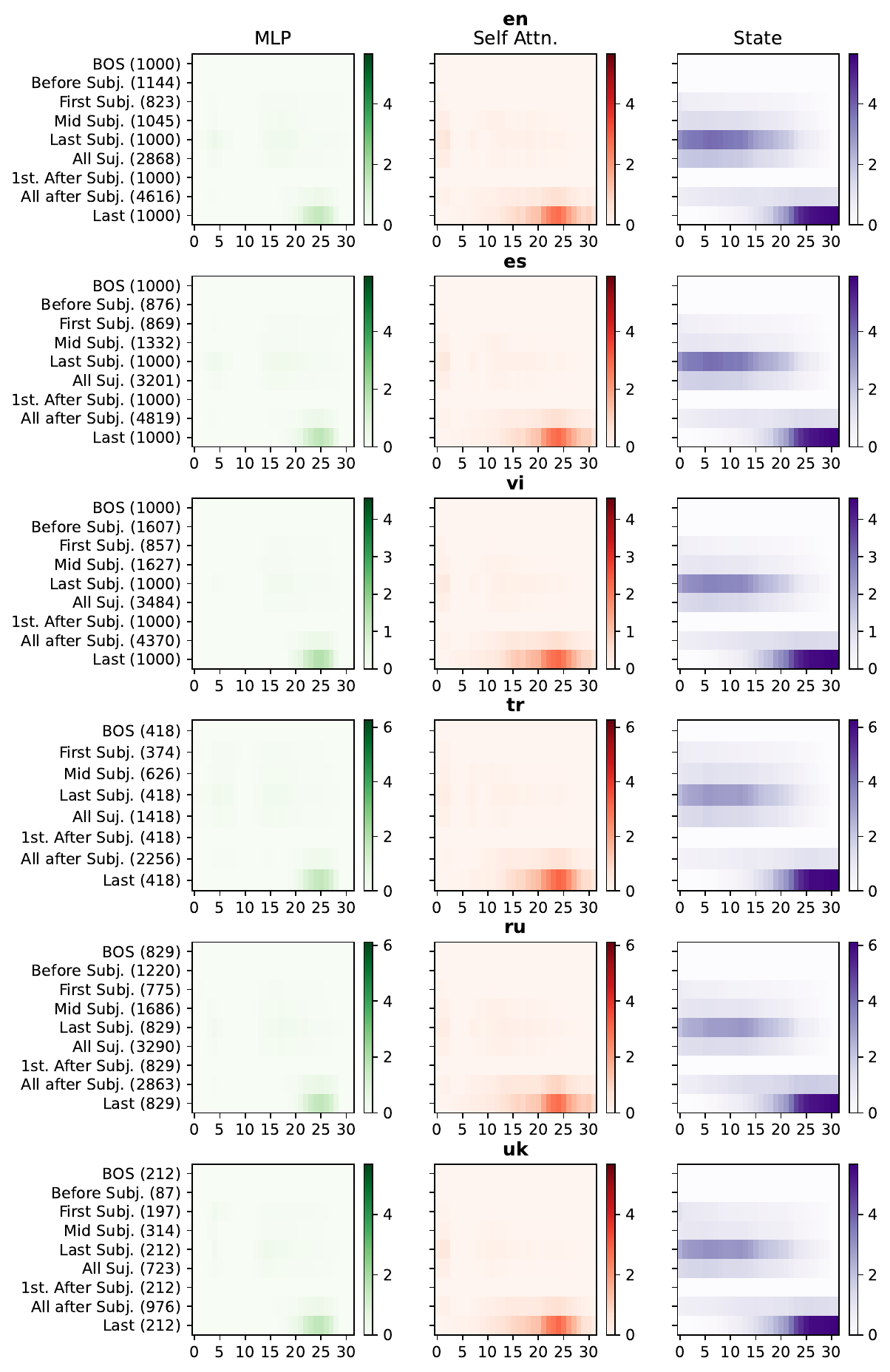} 
    \vspace{-3mm}
    \captionof{figure}{\xglm{} causal analysis for each language (continues in Figure~\ref{fig:xglm_causal_analysis_logits_2}). Average \textbf{logit score} (for the originally predicted token) recovered after corrupting the subject in the input and restoring: the hidden representation at a given layer (State), the MLP in a window of 4 layers, or the Self Attn. in a window of 4 layers.}
    \label{fig:xglm_causal_analysis_logits_1}
\end{minipage}

\begin{figure*}[h!]
    \centering
    \includegraphics[width=0.8\linewidth]{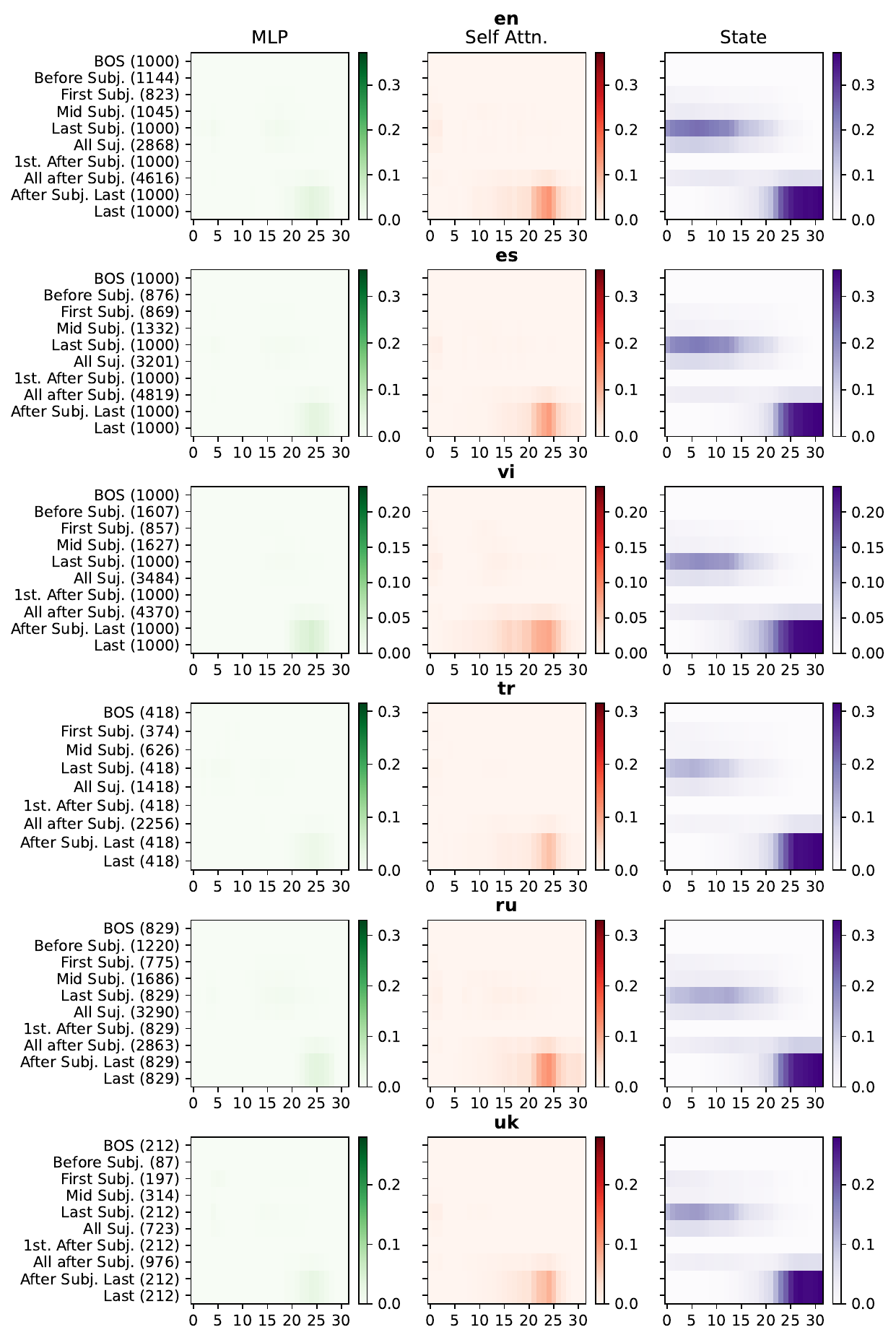} 
    \vspace{-3mm}
    \caption{\xglm{} causal analysis for each language (continues in Figure~\ref{fig:xglm_causal_analysis_probs_2}). Average \textbf{probability} (for the originally predicted token) recovered after corrupting the subject in the input and restoring: the hidden representation at a given layer (State), the MLP in a window of 4 layers, or the Self Attn. in a window of 4 layers.}
    \label{fig:xglm_causal_analysis_probs_1}
\end{figure*}

\begin{figure*}[h!]
    \centering
    \includegraphics[width=0.8\linewidth]{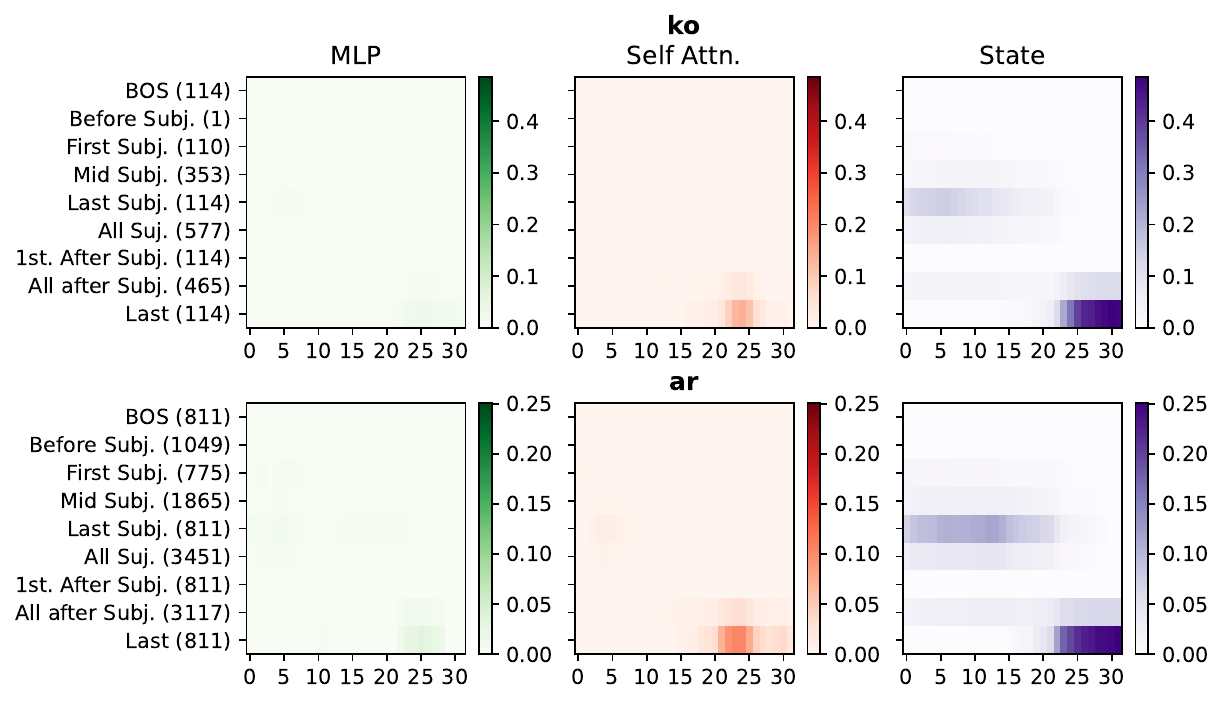} 
    \vspace{-3mm}
    \caption{\xglm{} causal analysis for each language (Rest of the languages in Figure~\ref{fig:xglm_causal_analysis_probs_1}). Average \textbf{probability} (for the originally predicted token) recovered after corrupting the subject in the input and restoring: the hidden representation at a given layer (State), the MLP in a window of 4 layers, or the Self Attn. in a window of 4 layers.}
    \label{fig:xglm_causal_analysis_probs_2}
\end{figure*}

\clearpage
\subsection{\eurollm{}}
\begin{figure*}[h!]
    \centering
    \includegraphics[width=0.9\linewidth]{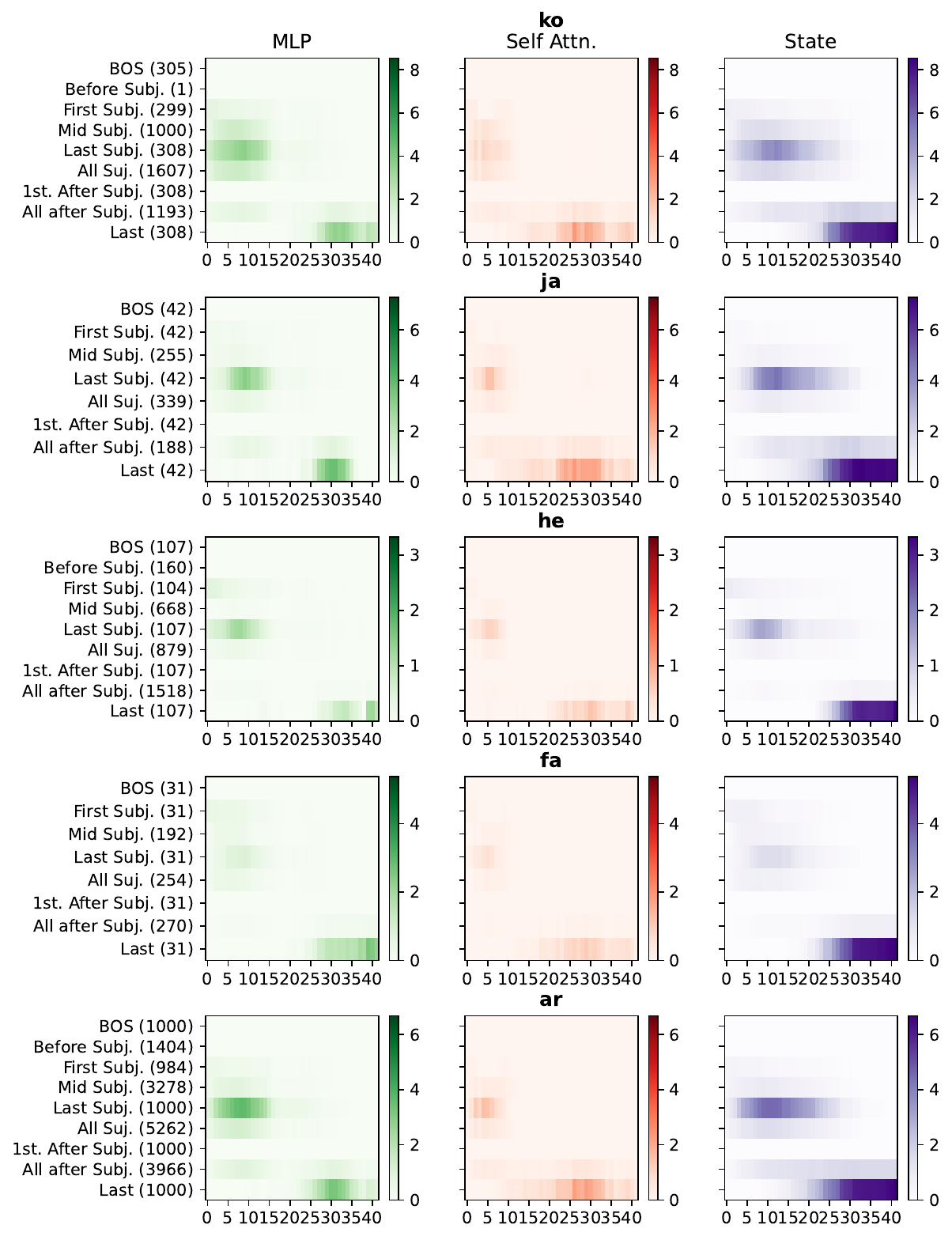} 
    \vspace{-3mm}
    \caption{\eurollm{} causal analysis for each language (Rest of the languages in Figure~\ref{fig:eurollm_causal_analysis_logits_1}). Average \textbf{logit score} (for the originally predicted token) recovered after corrupting the subject in the input and restoring: the hidden representation at a given layer (State), the MLP in a window of 5 layers, or the Self Attn. in a window of 5 layers.}
    \label{fig:eurollm_causal_analysis_logits_2}
\end{figure*}

\begin{minipage}{\textwidth}
    \centering
    \includegraphics[width=0.9\linewidth]{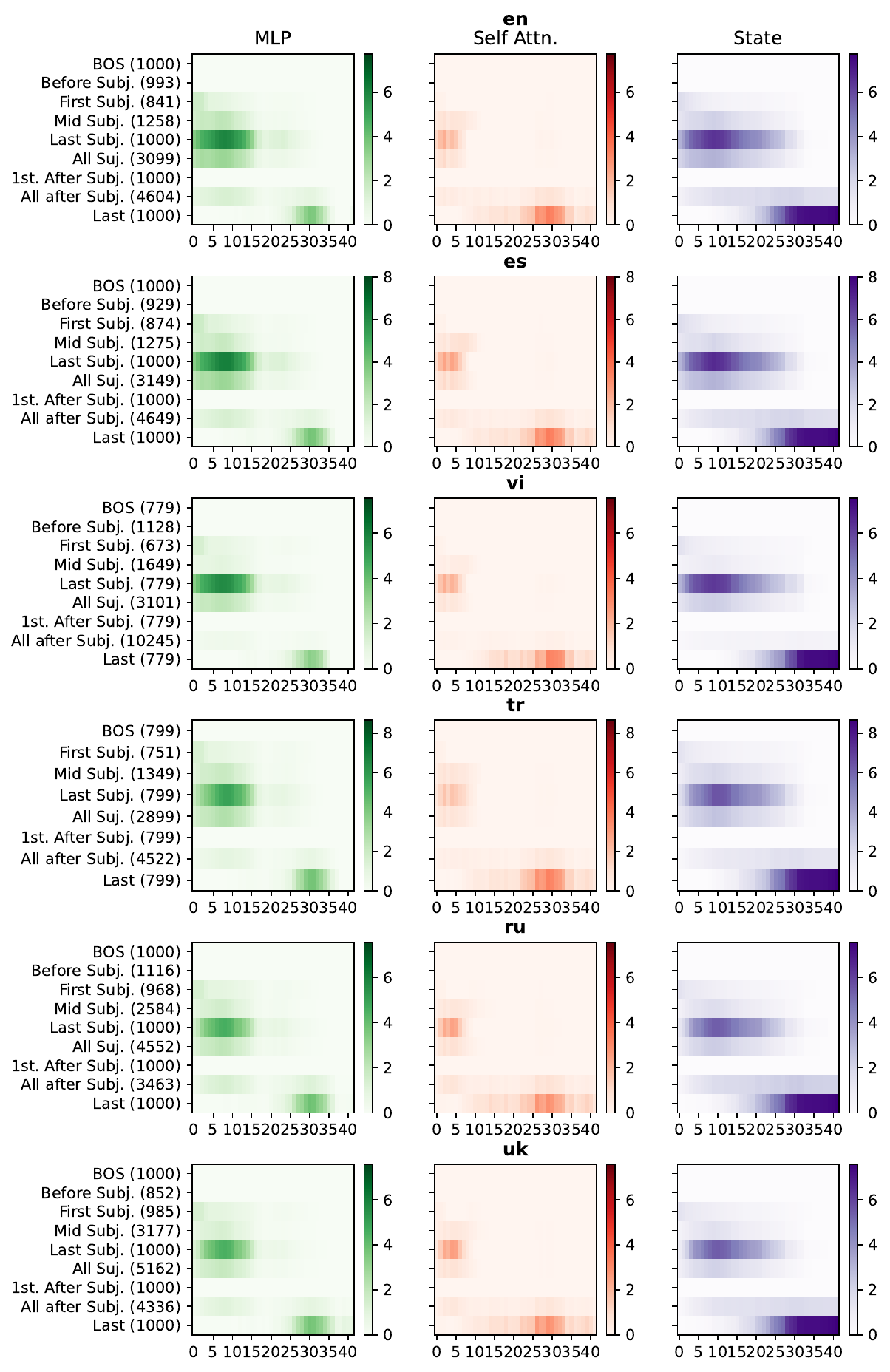} 
    \vspace{-3mm}
    \captionof{figure}{\eurollm{} causal analysis for each language (continues in Figure~\ref{fig:eurollm_causal_analysis_logits_2}). Average \textbf{logit score} (for the originally predicted token) recovered after corrupting the subject in the input and restoring: the hidden representation at a given layer (State), the MLP in a window of 5 layers, or the Self Attn. in a window of 5 layers.}
    \label{fig:eurollm_causal_analysis_logits_1}
\end{minipage}

\begin{figure*}[h!]
    \centering
    \includegraphics[width=0.9\linewidth]{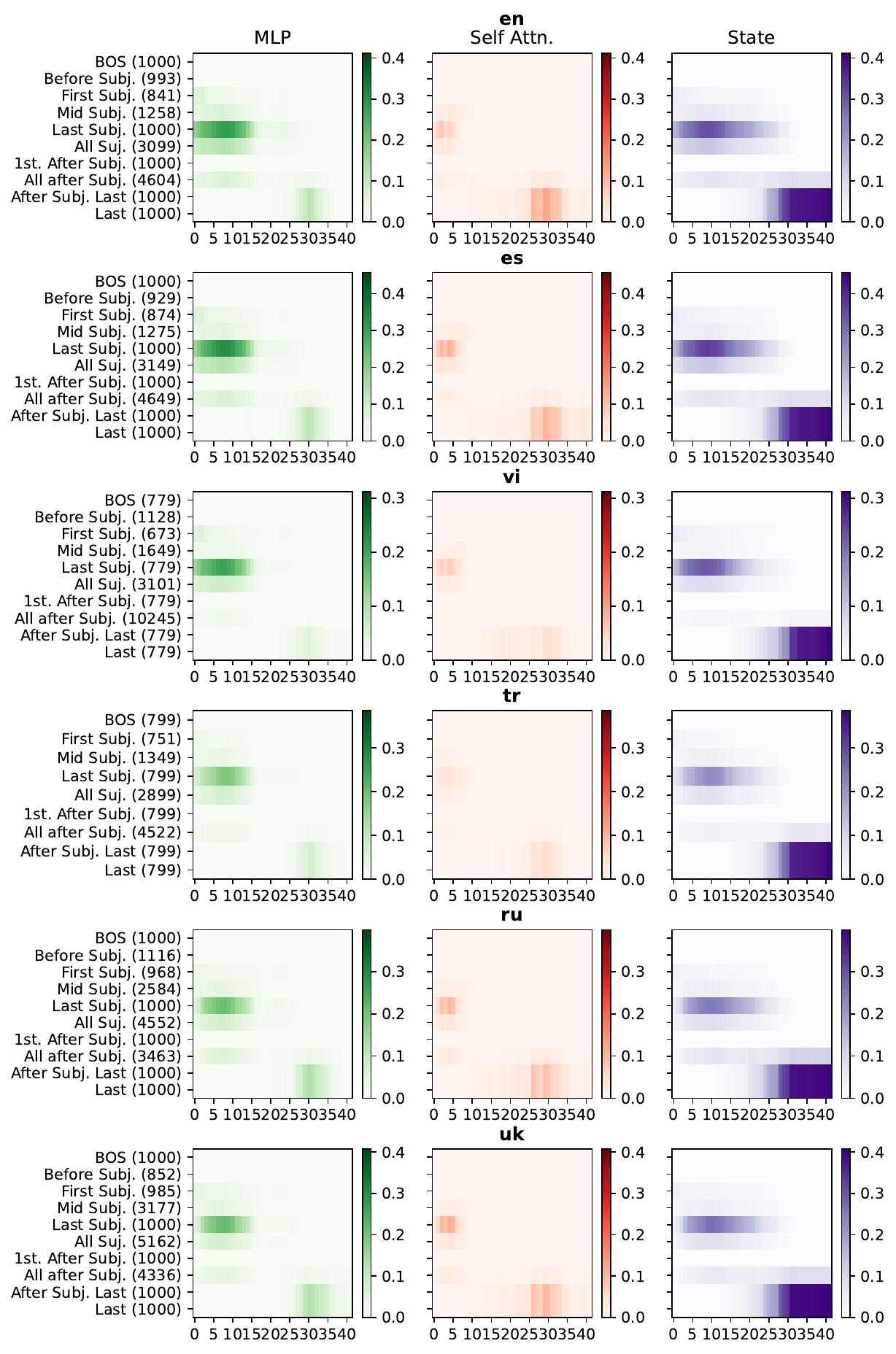} 
    \vspace{-3mm}
    \caption{\eurollm{} causal analysis for each language (continues in Figure~\ref{fig:eurollm_causal_analysis_probs_2}). Average \textbf{probability} (for the originally predicted token) recovered after corrupting the subject in the input and restoring: the hidden representation at a given layer (State), the MLP in a window of 5 layers, or the Self Attn. in a window of 5 layers.}
    \label{fig:eurollm_causal_analysis_probs_1}
\end{figure*}

\begin{figure*}[h!]
    \centering
    \includegraphics[width=0.9\linewidth]{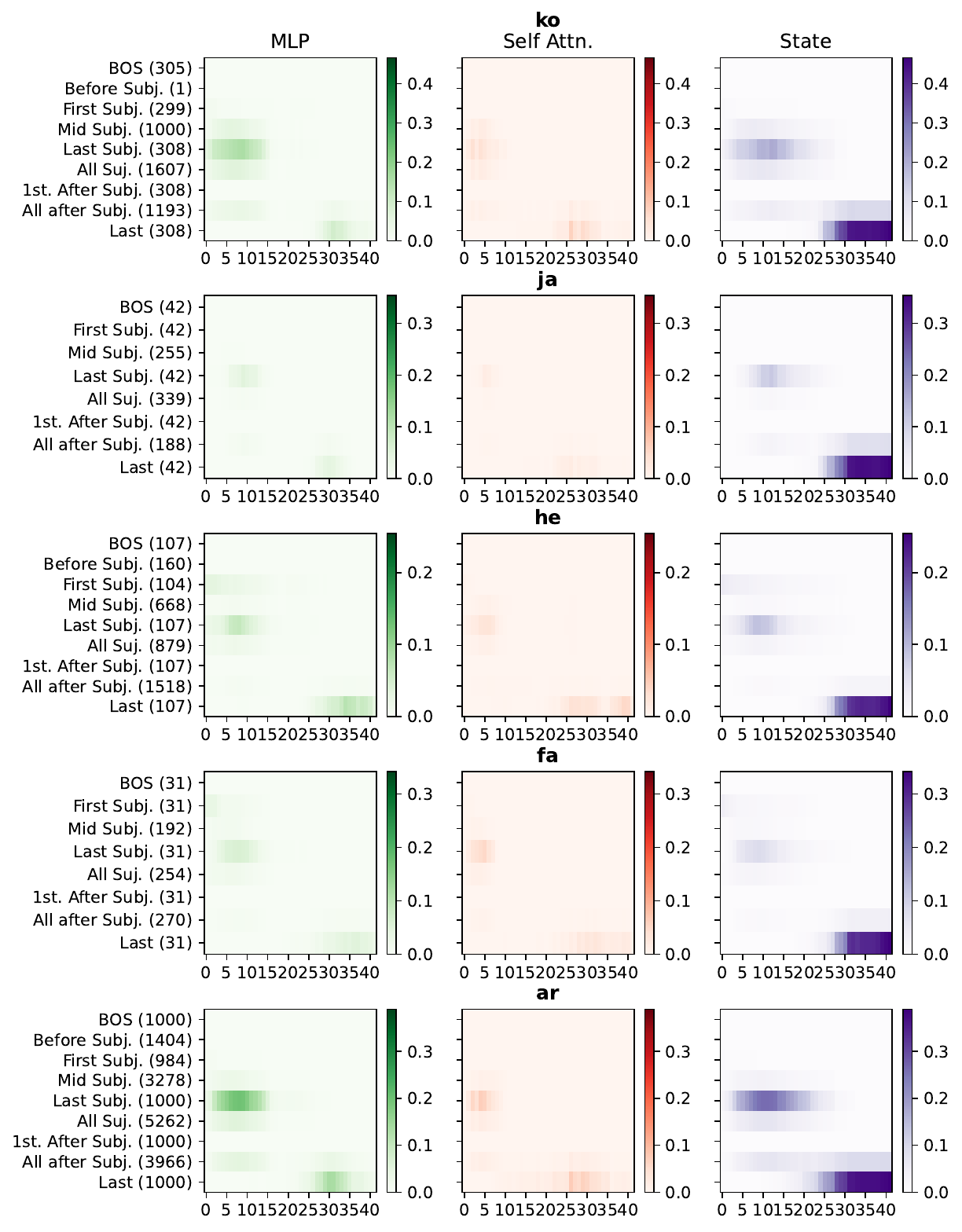} 
    \vspace{-3mm}
    \caption{\mtfive{} causal analysis for each language (Rest of the languages in Figure~\ref{fig:eurollm_causal_analysis_probs_1}). Average \textbf{probability} (for the originally predicted token) recovered after corrupting the subject in the input and restoring: the hidden representation at a given layer (State), the MLP in a window of 5 layers, or the Self Attn. in a window of 5 layers.}
    \label{fig:eurollm_causal_analysis_probs_2}
\end{figure*}

\clearpage
\subsection{\mtfive{}}
\begin{figure*}[h!]
    \centering
    \includegraphics[width=\linewidth]{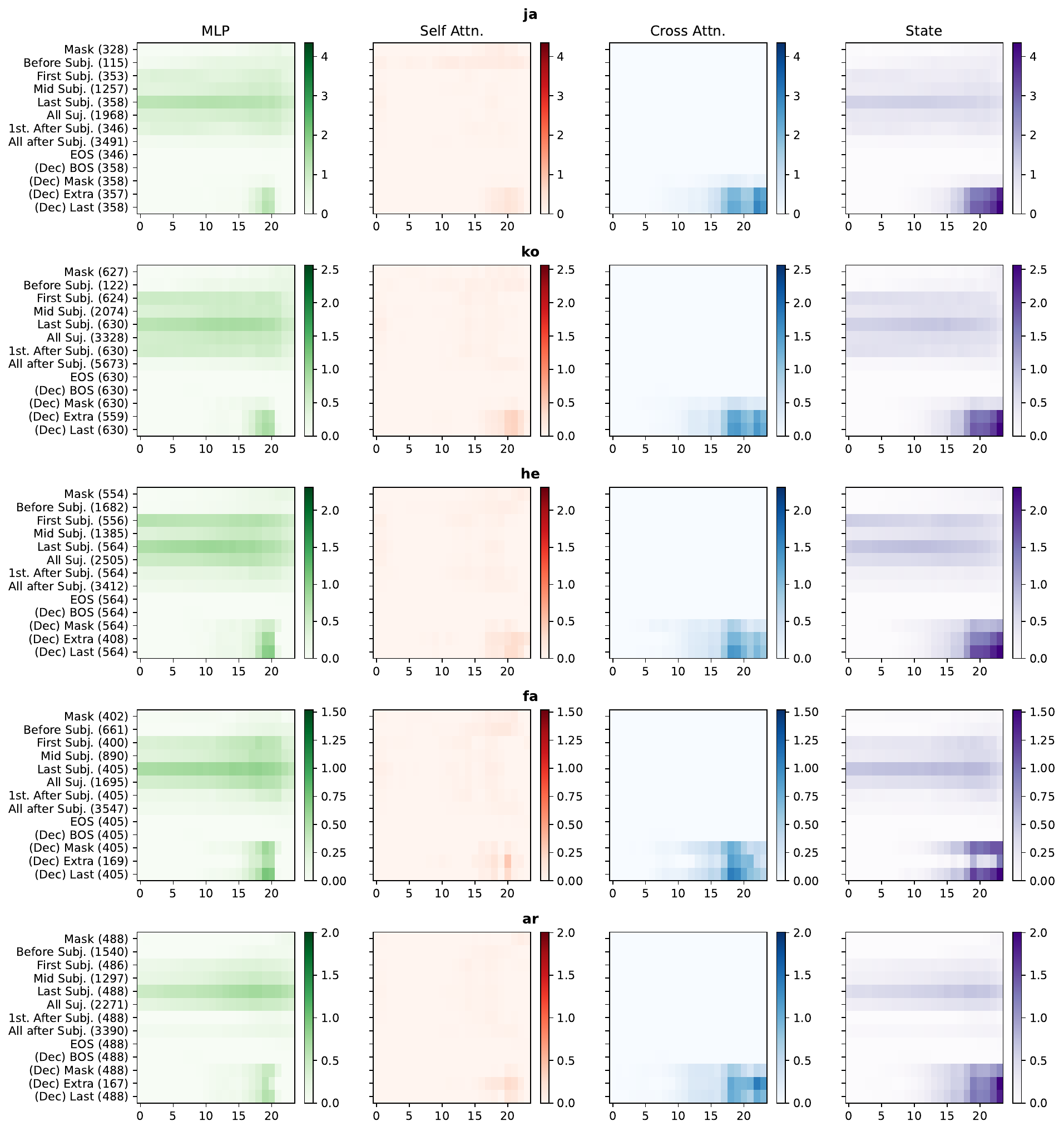} 
    \vspace{-3mm}
    \caption{\mtfive{} causal analysis for each language (Rest of the languages in Figure~\ref{fig:mt5_causal_analysis_logits_1}). Average \textbf{logit score} (for the originally predicted token) recovered after corrupting the subject in the input and restoring: the hidden representation at a given layer (State), the MLP in a window of 3 layers, or the Self Attn. in a window of 3 layers.}
    \label{fig:mt5_causal_analysis_logits_2}
\end{figure*}

\begin{minipage}{\textwidth}
    \centering
    \includegraphics[width=\linewidth]{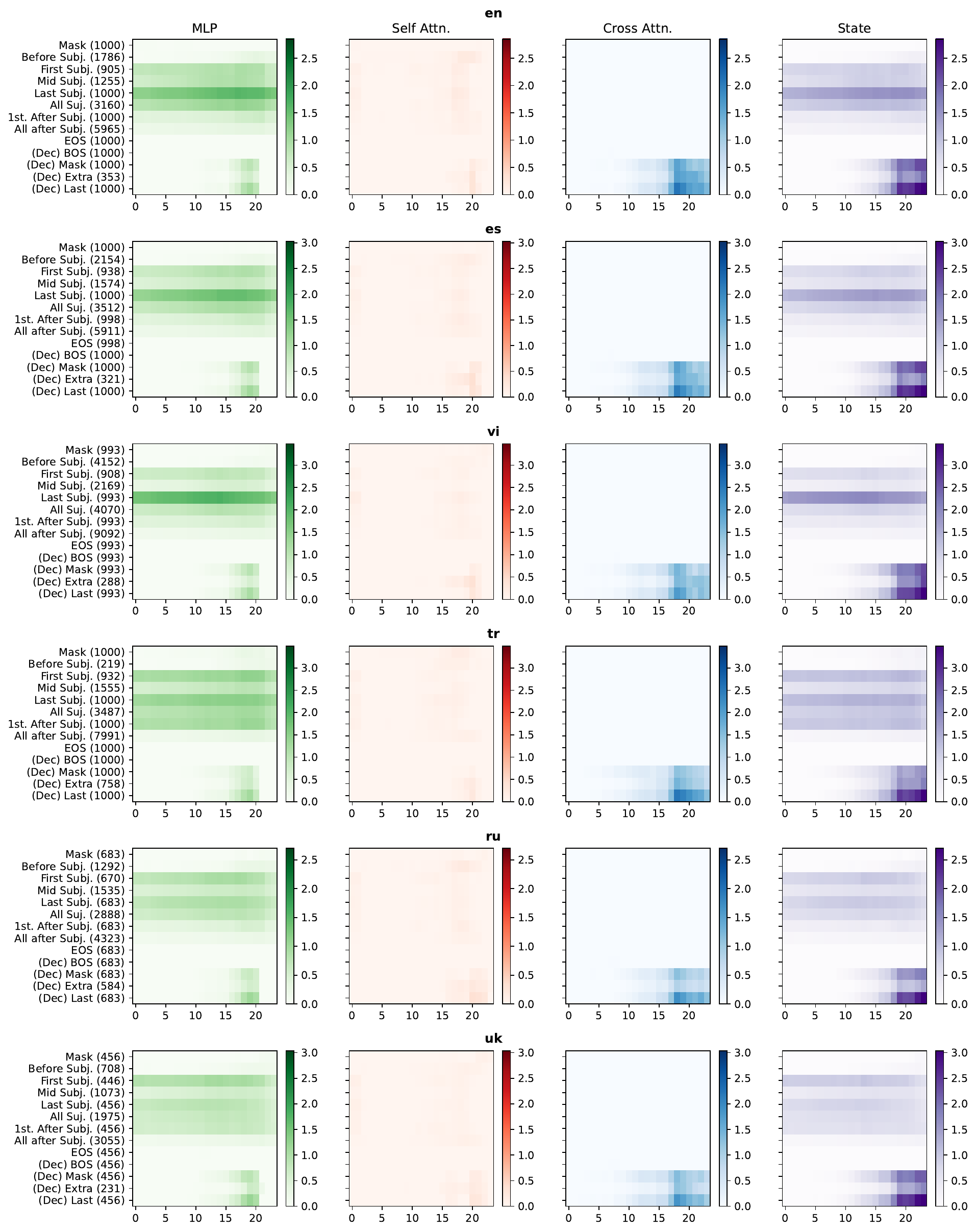} 
    \vspace{-3mm}
    \captionof{figure}{\mtfive{} causal analysis for each language (continues in Figure~\ref{fig:mt5_causal_analysis_logits_2}). Average \textbf{logit score} (for the originally predicted token) recovered after corrupting the subject in the input and restoring: the hidden representation at a given layer (State), the MLP in a window of 3 layers, or the Self Attn. in a window of 3 layers.}
    \label{fig:mt5_causal_analysis_logits_1}
\end{minipage}

\begin{figure*}[h!]
    \centering
    \includegraphics[width=\linewidth]{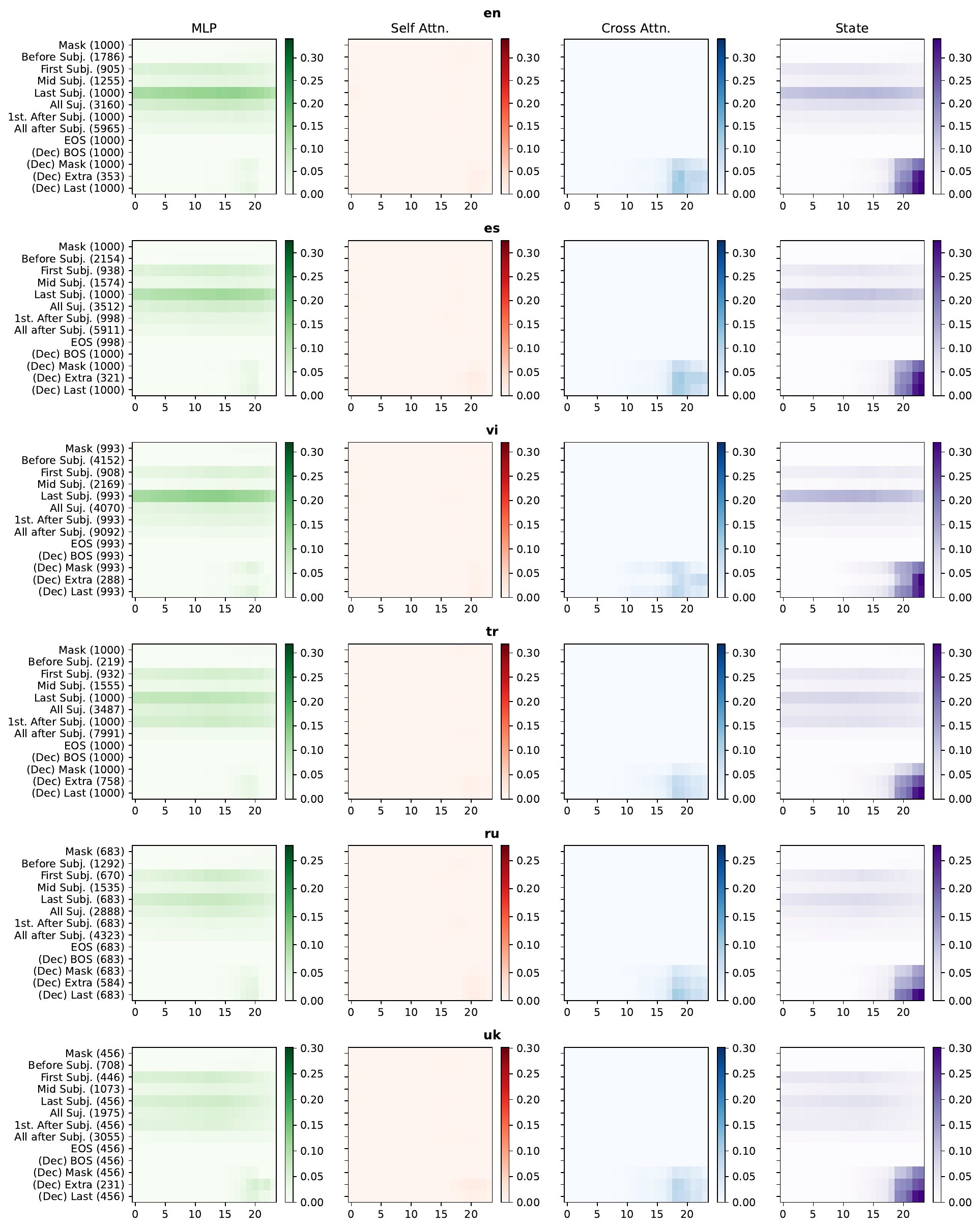} 
    \vspace{-3mm}
    \caption{\mtfive{} causal analysis for each language (continues in Figure~\ref{fig:mt5_causal_analysis_probs_2}). Average \textbf{probability} (for the originally predicted token) recovered after corrupting the subject in the input and restoring: the hidden representation at a given layer (State), the MLP in a window of 3 layers, or the Self Attn. in a window of 3 layers.}
    \label{fig:mt5_causal_analysis_probs_1}
\end{figure*}

\begin{figure*}[h!]
    \centering
    \includegraphics[width=\linewidth]{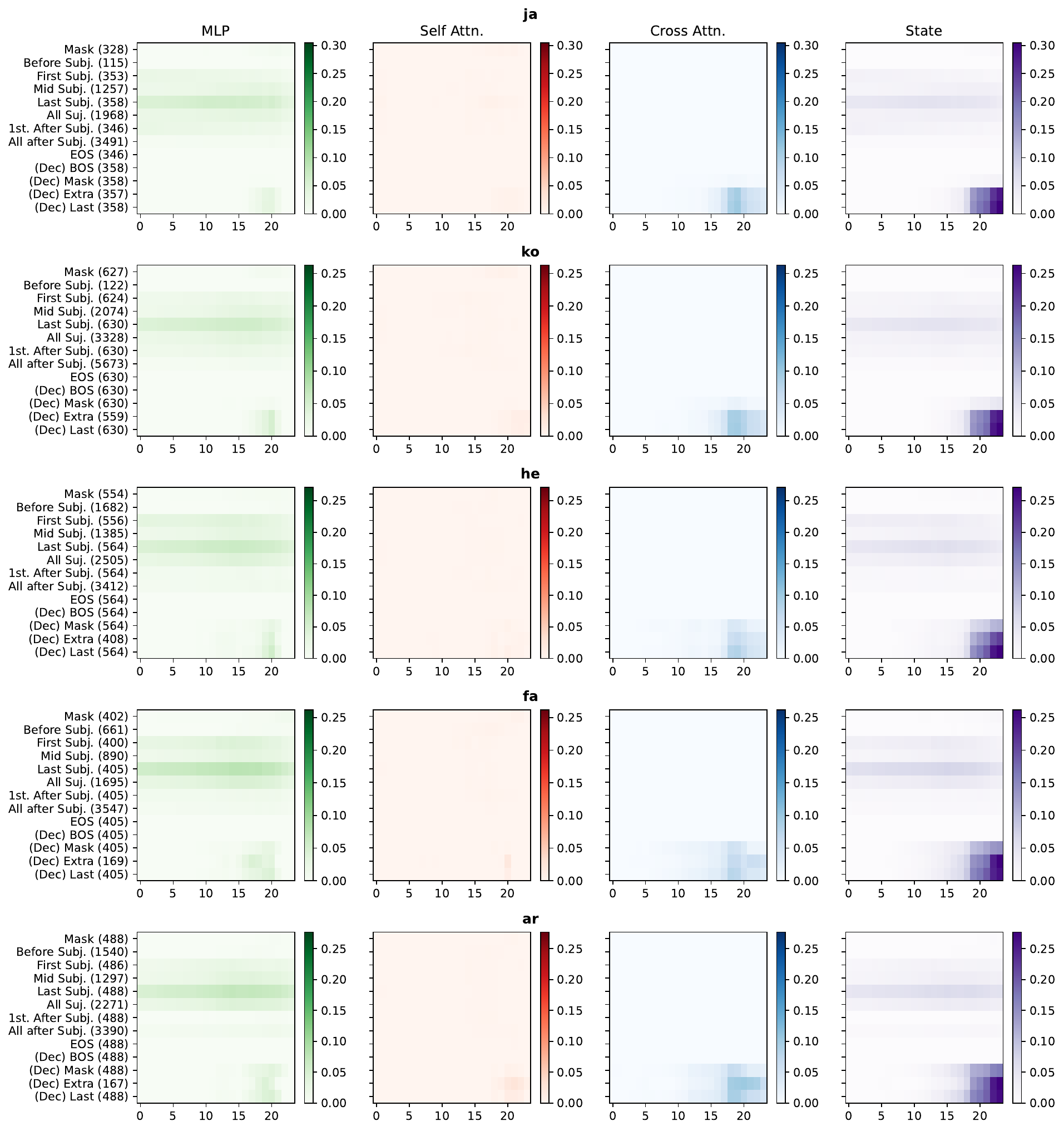} 
    \vspace{-3mm}
    \caption{\mtfive{} causal analysis for each language (Rest of the languages in Figure~\ref{fig:mt5_causal_analysis_probs_1}). Average \textbf{probability} (for the originally predicted token) recovered after corrupting the subject in the input and restoring: the hidden representation at a given layer (State), the MLP in a window of 3 layers, or the Self Attn. in a window of 3 layers.}
    \label{fig:mt5_causal_analysis_probs_2}
\end{figure*}

\clearpage
\section{Attention Knockout}\label{appendix:attn_knockout}

\begin{minipage}{\textwidth}
    \centering
    \vspace{0.5cm}
    \includegraphics[width=\textwidth]{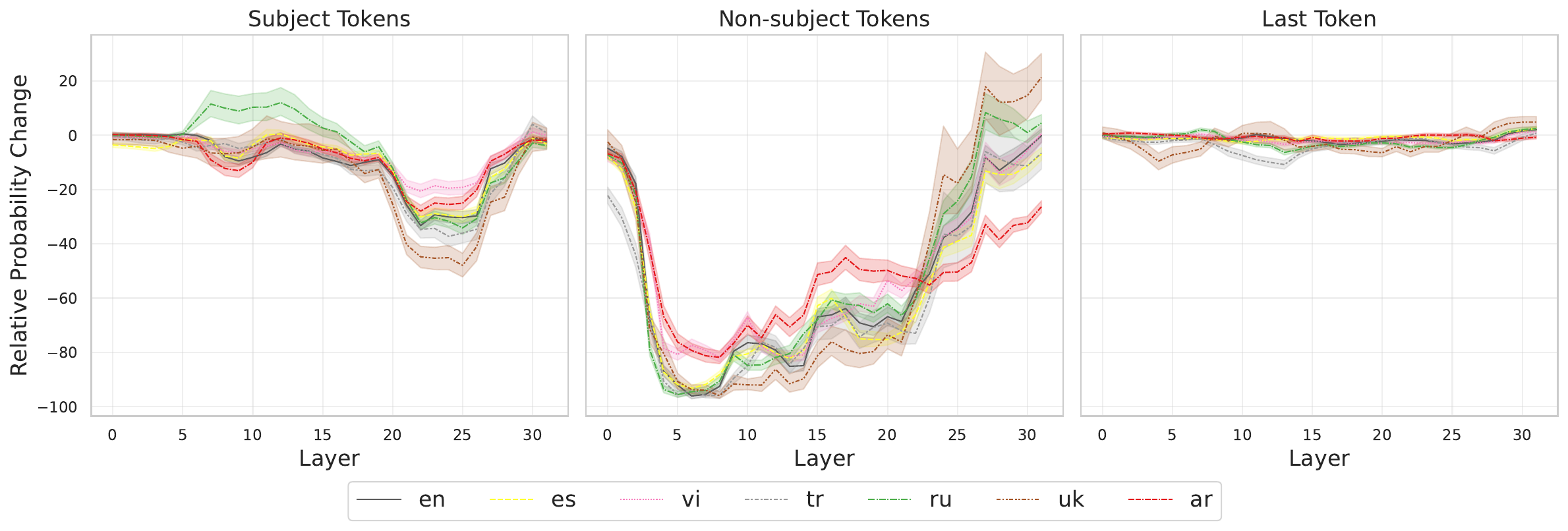}
    \includegraphics[width=\textwidth]{images/eurollm_attn_knockout.pdf}
    \includegraphics[width=\textwidth]{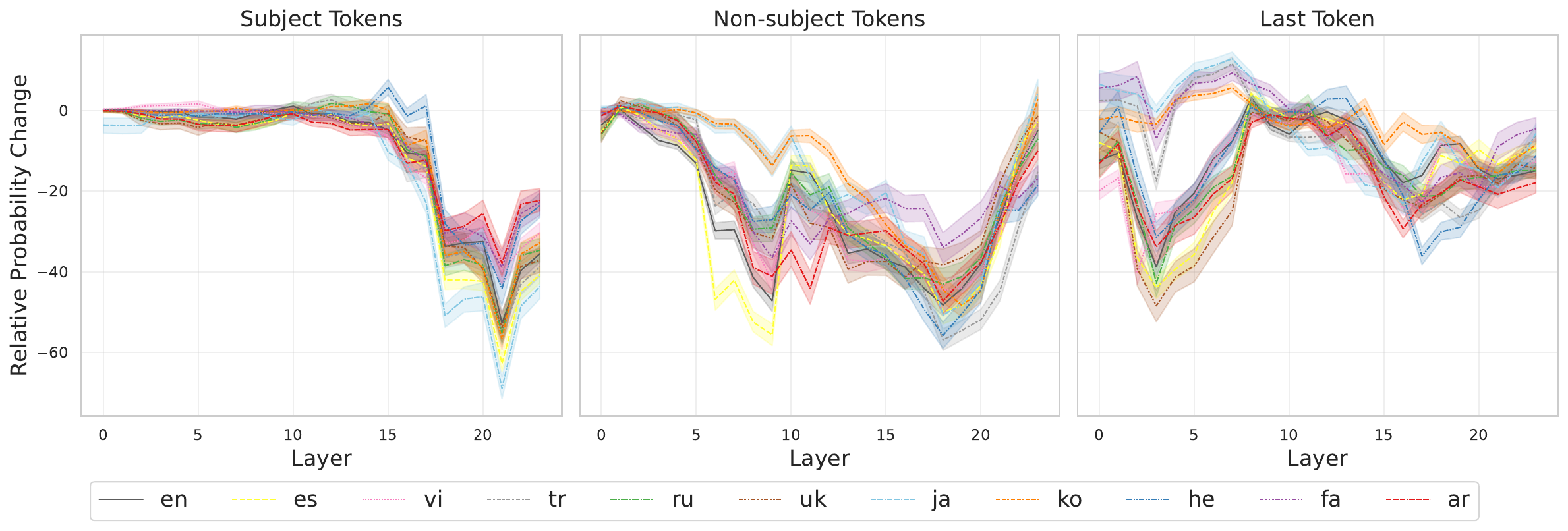}
    \vspace{-1em}
    \captionof{figure}{Attention knockout between the last token and a given set of tokens. Each layer represents the effect of the knockout on a window of \(w\) layers. Models from top to bottom: \xglm{} (\(w=6\)), \eurollm{} (\(w=7\)), \mtfive{} (\(w=4\)).}
    \label{appendix:attention_knockout}
\end{minipage}

\begin{figure*}[h!]
    \centering
    \includegraphics[width=0.7\textwidth]{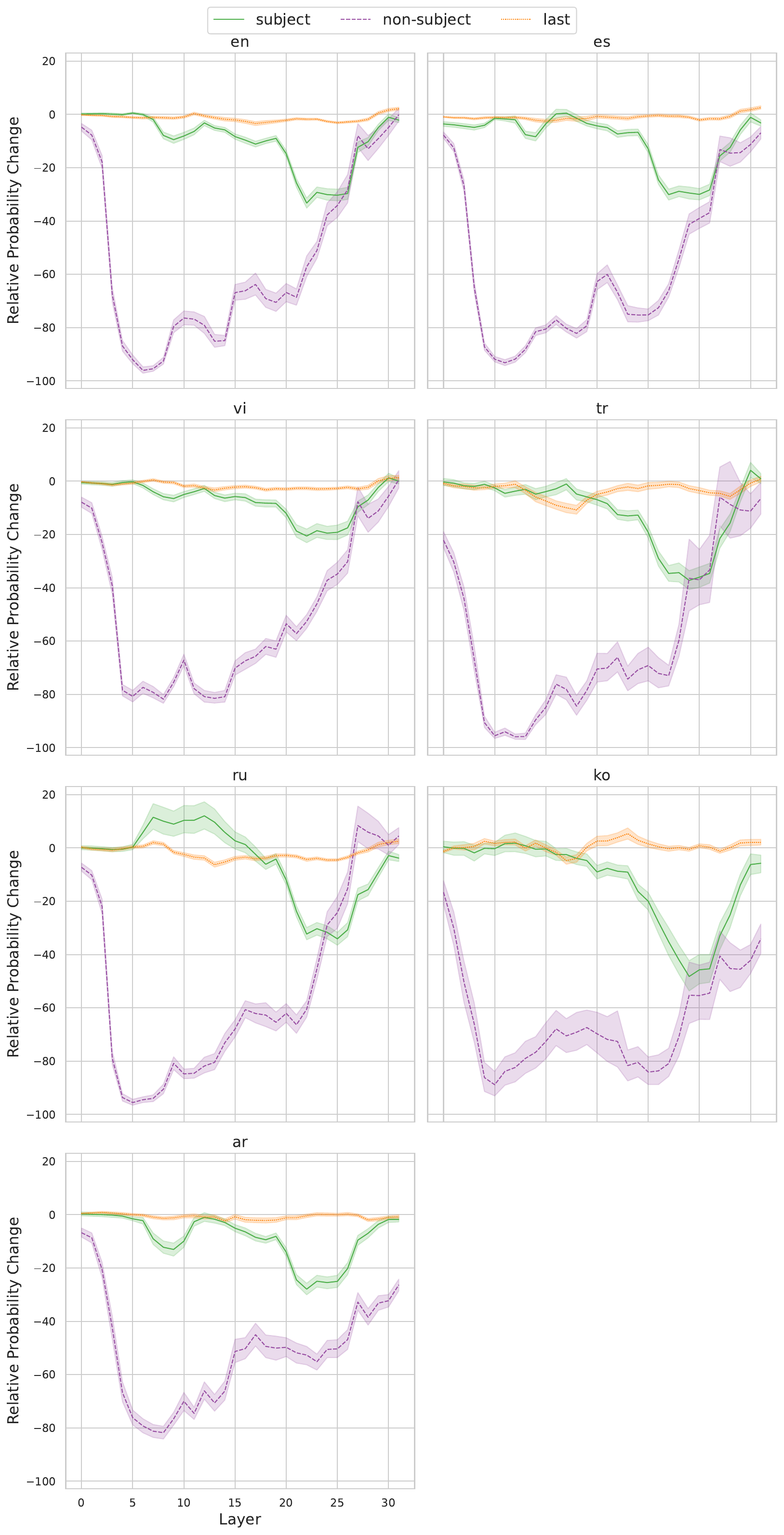} 
    \caption{\xglm{} attention knockout for each language.}
    \label{fig:xglm_attn_knockout_each_lang}
\end{figure*}

\begin{figure*}[h!]
    \centering
    \includegraphics[width=0.7\textwidth]{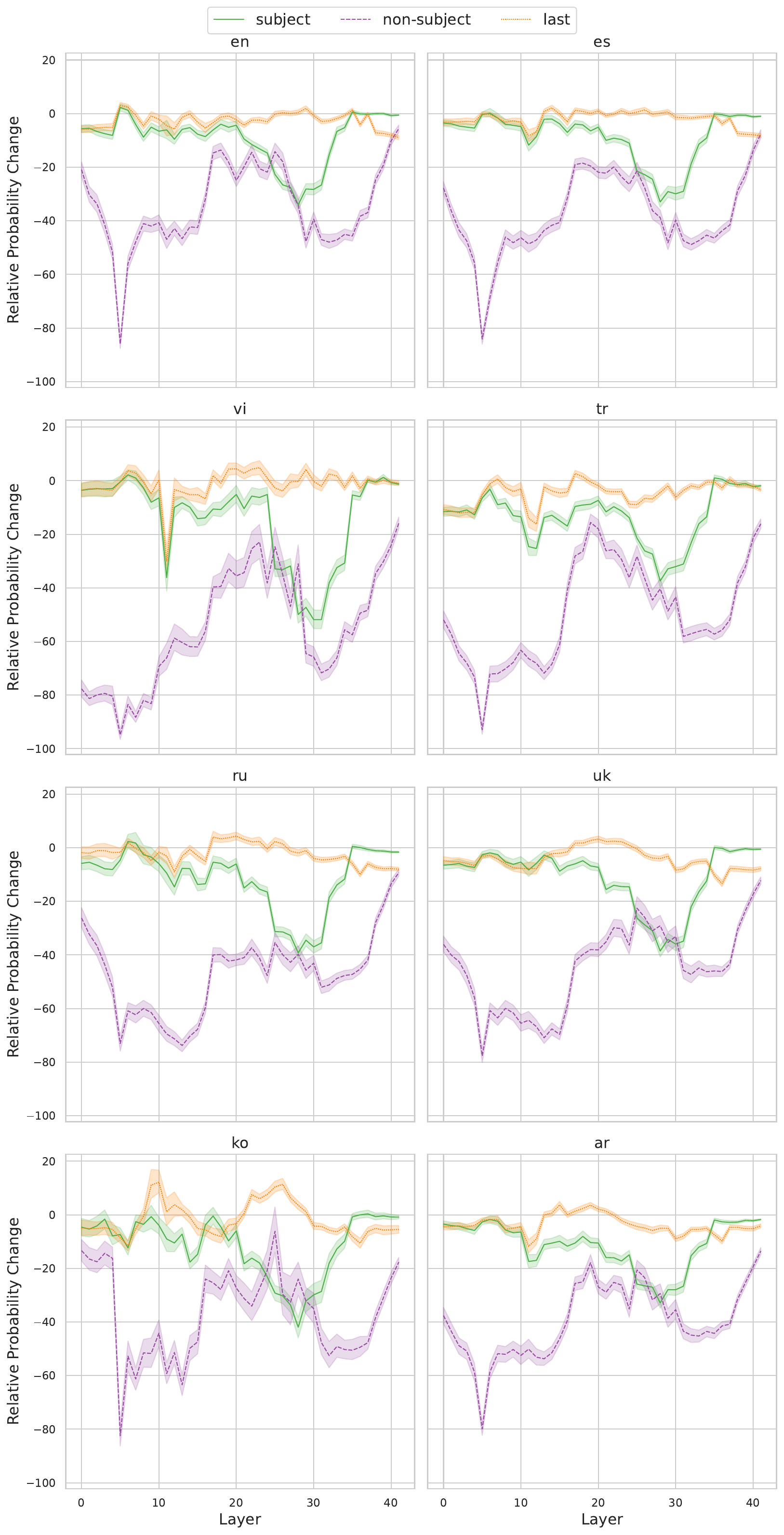} 
    \caption{\eurollm{} attention knockout for each language.}
    \label{fig:eurollm_attn_knockout_each_lang}
\end{figure*}

\begin{figure*}[h!]
    \centering
    \includegraphics[width=\textwidth]{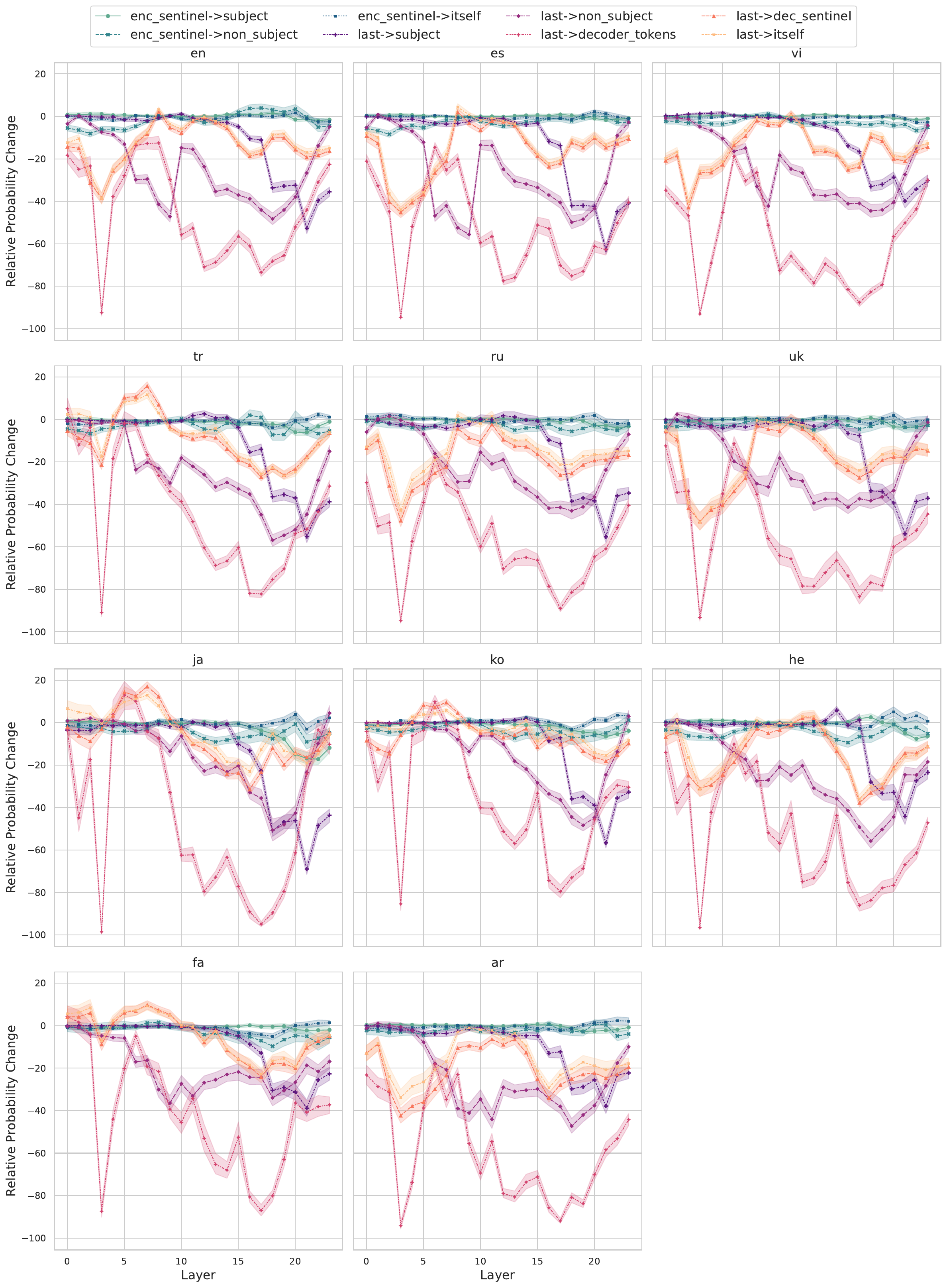} 
    \caption{Attention knockout for each language in \mtfive{}. The knockout is performed between the last token in the decoder (``last'') to a given set of tokens \(\{t\}\), and between the masked token in the encoder (``enc\_sentinel'') and \(\{t\}\). From the encoder sentinel there is no much flow of information so these were not included in the main body.}
    \label{fig:mt5_attn_knockout_each_lang}
\end{figure*}

\clearpage
\section{Extraction Event}\label{appendix:prediction_extraction}

\begin{minipage}{\textwidth}
    \centering
    \includegraphics[width=0.6\textwidth]{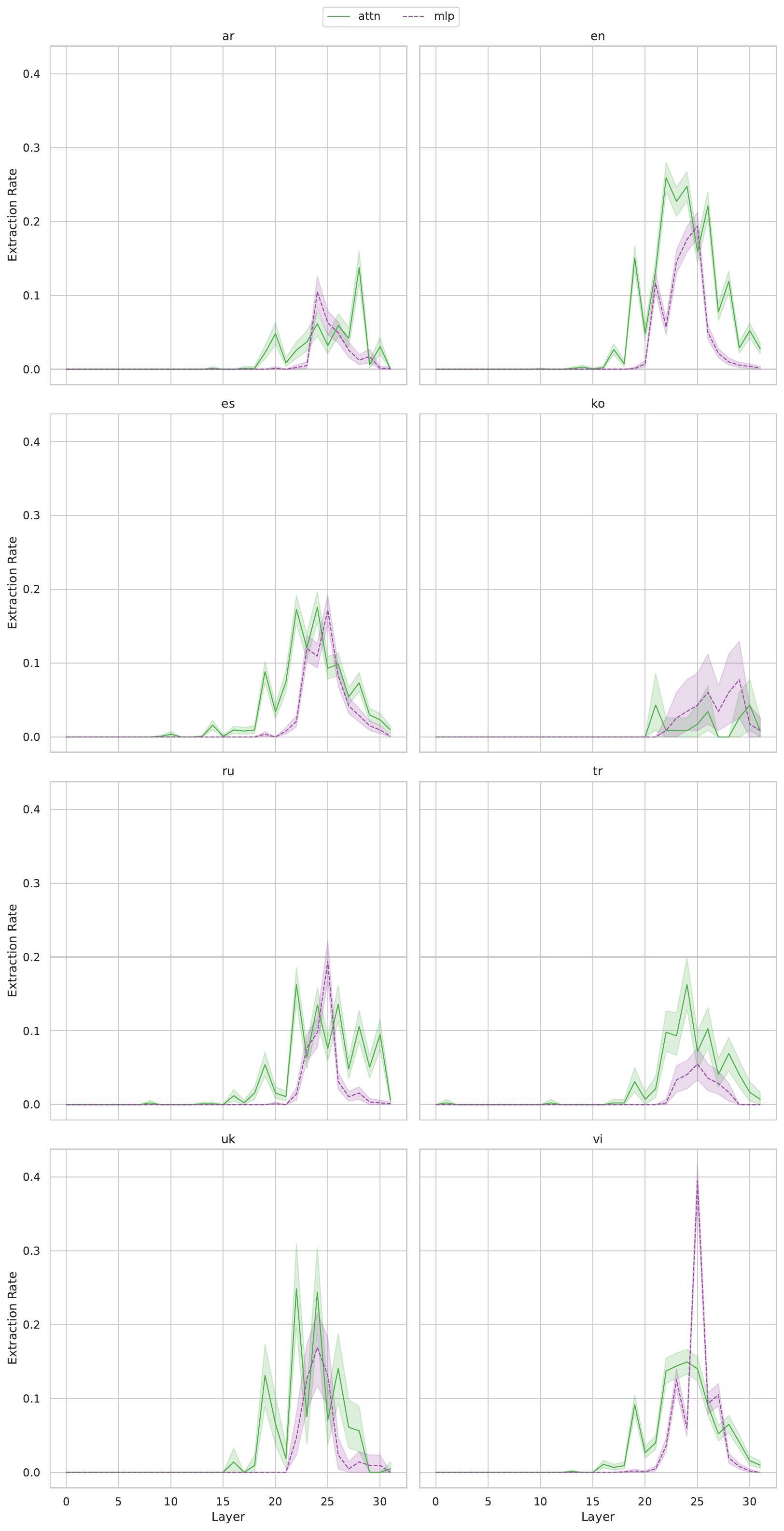} 
    \captionof{figure}{\xglm{} extraction rates for each language.}
    \label{fig:xglm_extraction_rates_per_language}
\end{minipage}

\begin{figure*}[h!]
    \centering
    \includegraphics[width=0.6\textwidth]{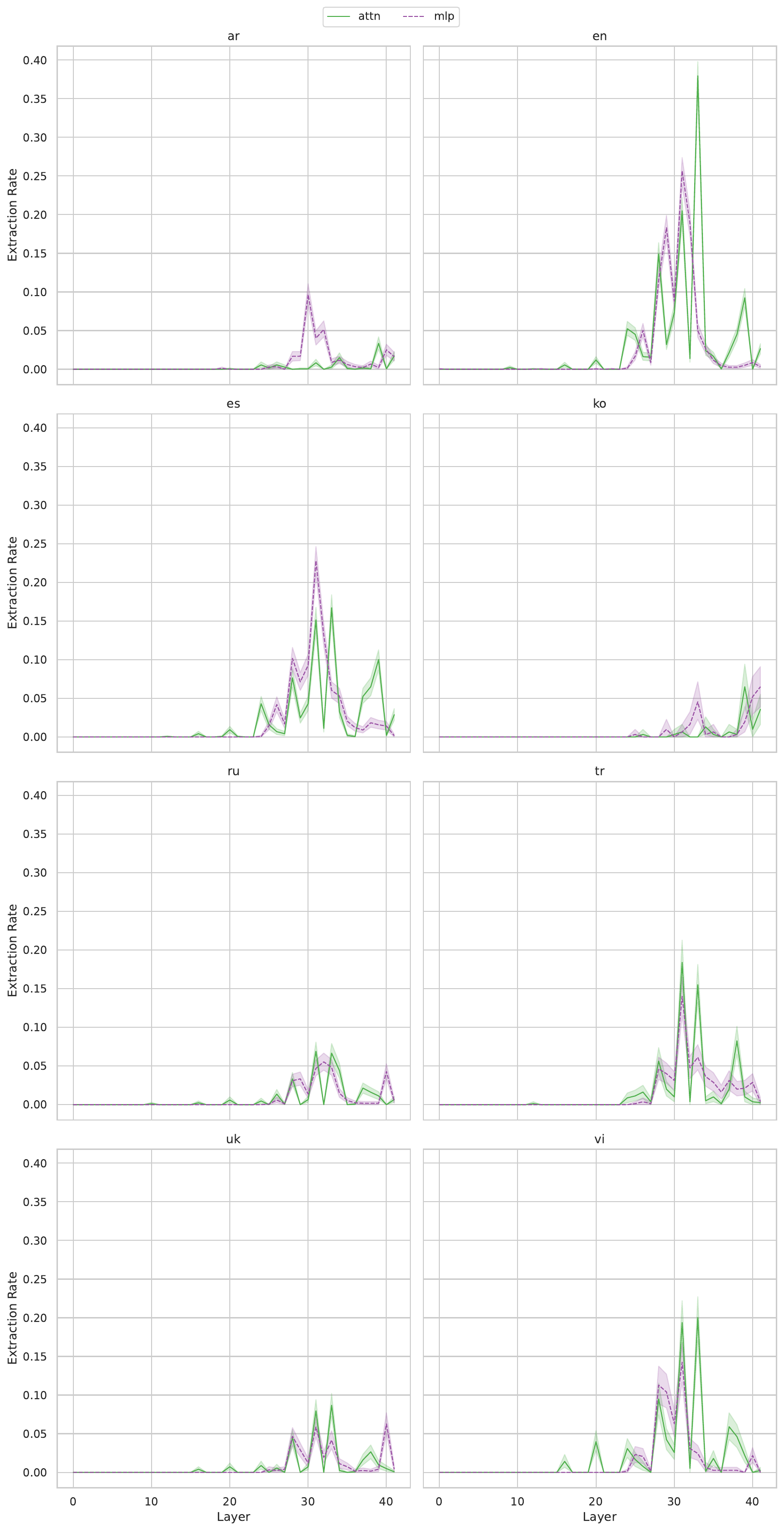} 
    \caption{\eurollm{} extraction rates for each language.}
    \label{fig:eurollm_extraction_rates_per_language}
\end{figure*}

\begin{figure*}[h!]
    \centering
    \includegraphics[width=0.8\textwidth]{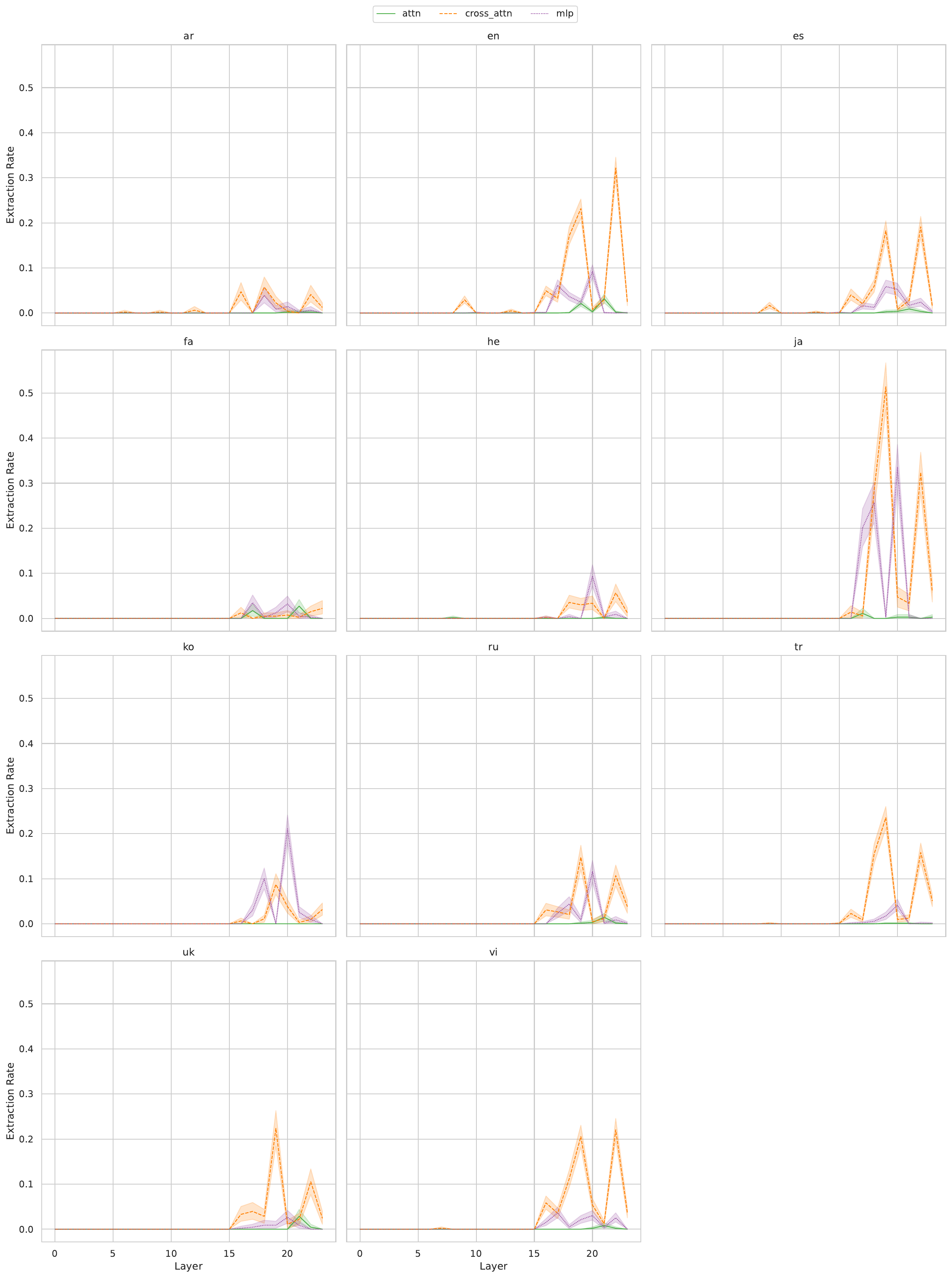} 
    \caption{\mtfive{} extraction rates for each language.}
    \label{fig:mt5_extraction_rates_per_language}
\end{figure*}

\begin{figure*}[h!]
    \centering
    \begin{minipage}{\linewidth}
        \centering
        \includegraphics[width=0.45\linewidth]{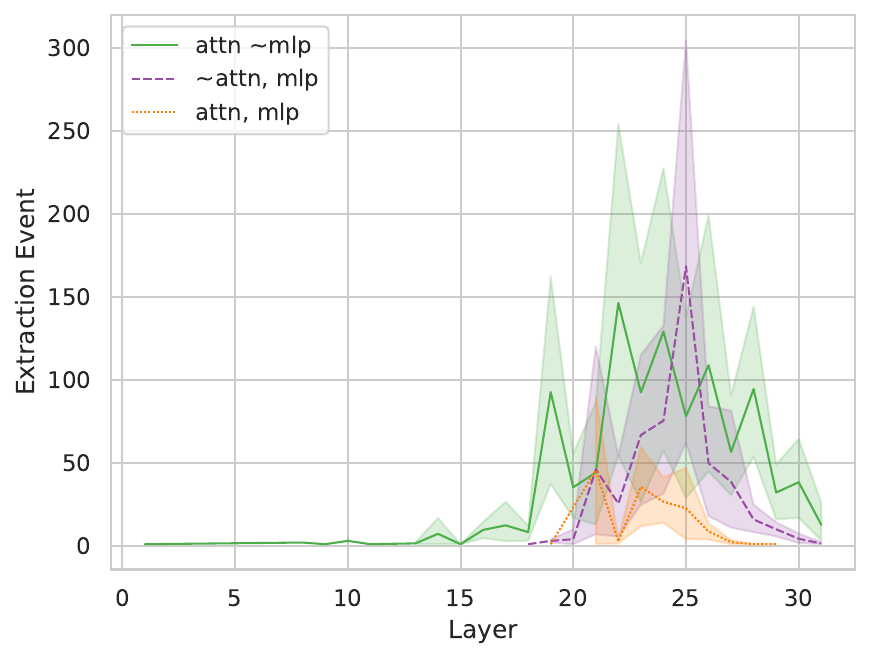}
    \end{minipage}
    \vspace{5mm}
    \begin{minipage}{\linewidth}
        \centering
        \includegraphics[width=0.45\linewidth]{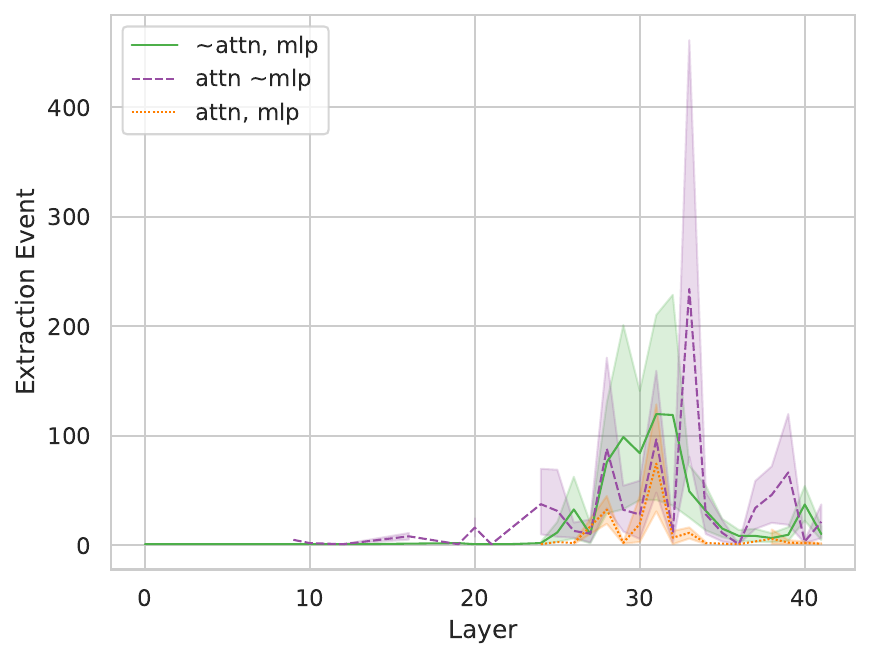}
    \end{minipage}
    \vspace{5mm} 
    \begin{minipage}{\linewidth}
        \centering
        \includegraphics[width=0.45\linewidth]{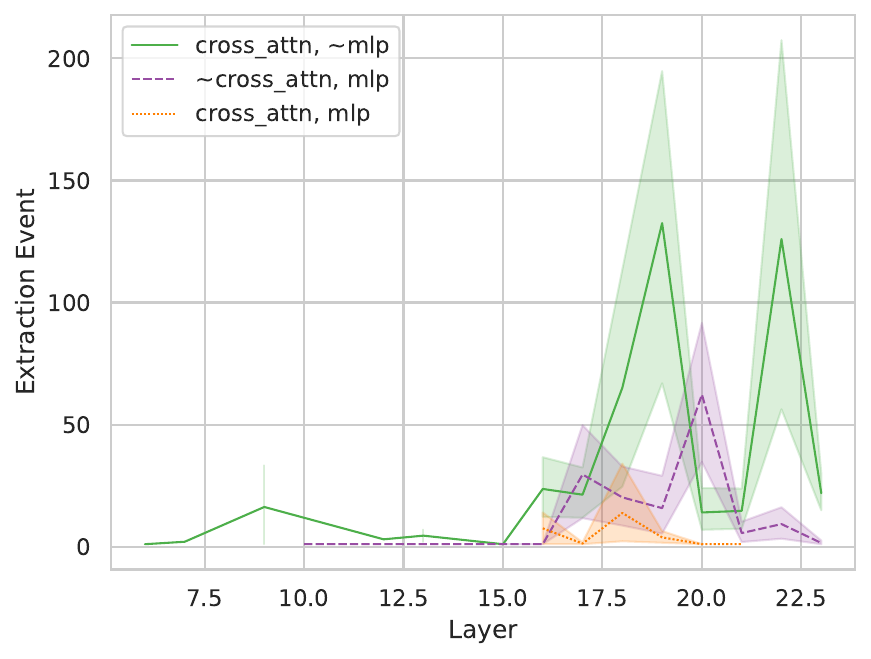}
    \end{minipage}
    \caption{Number of extraction events split by precedence (or not) of an extraction event in the self-attn or cross-attn. Models from top to bottom: \xglm{}, \eurollm, \mtfive{}.}
    \label{fig:extraction_rates_mlp_vs_attn}
\end{figure*}

\begin{figure*}[h!]
    \centering
    \includegraphics[width=0.6\textwidth]{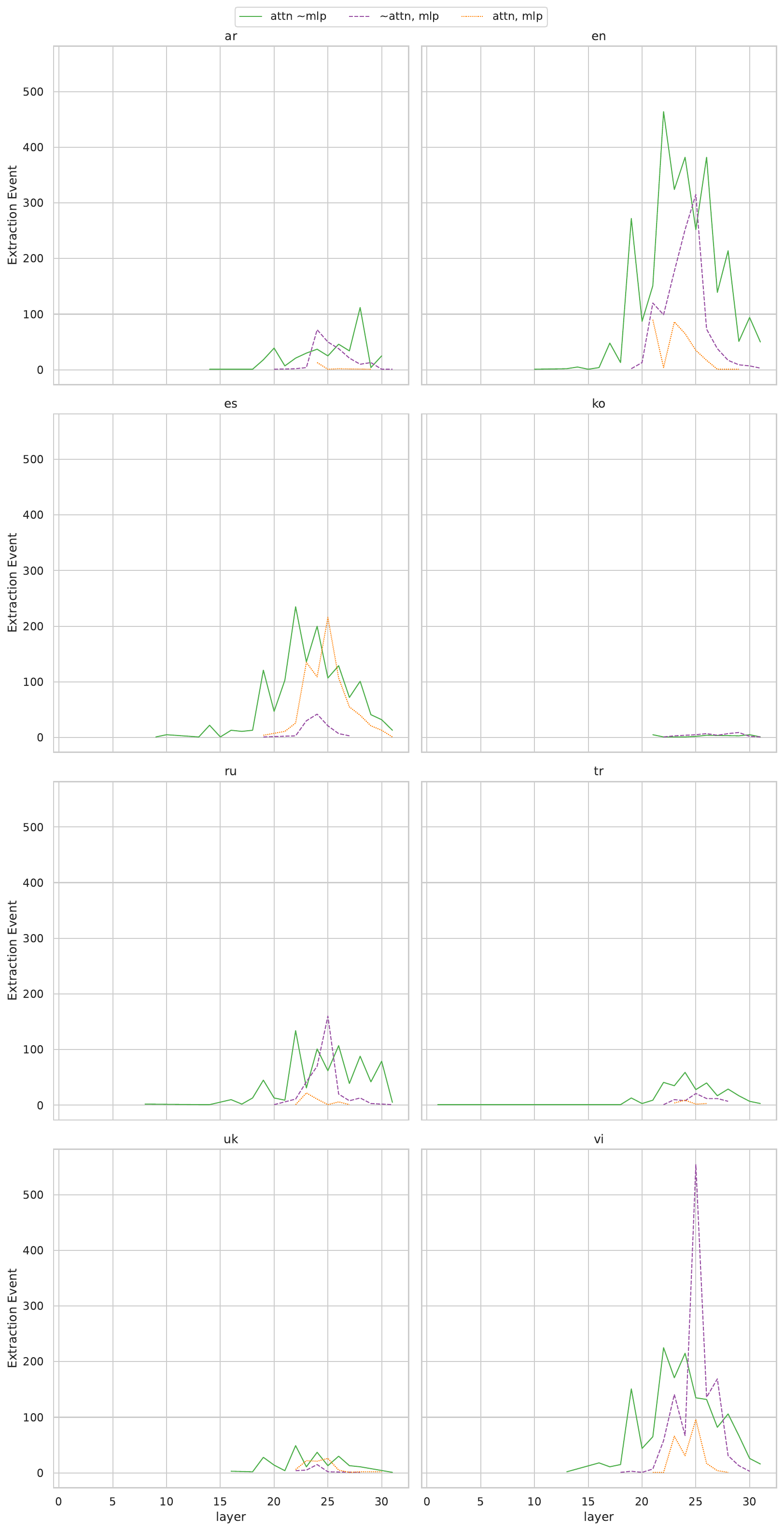} 
    \caption{Number of extraction events split by precedence (or not) of an extraction event in the self-attn in \xglm{} for each language.}
\end{figure*}

\begin{figure*}[h!]
    \centering
    \includegraphics[width=0.6\textwidth]{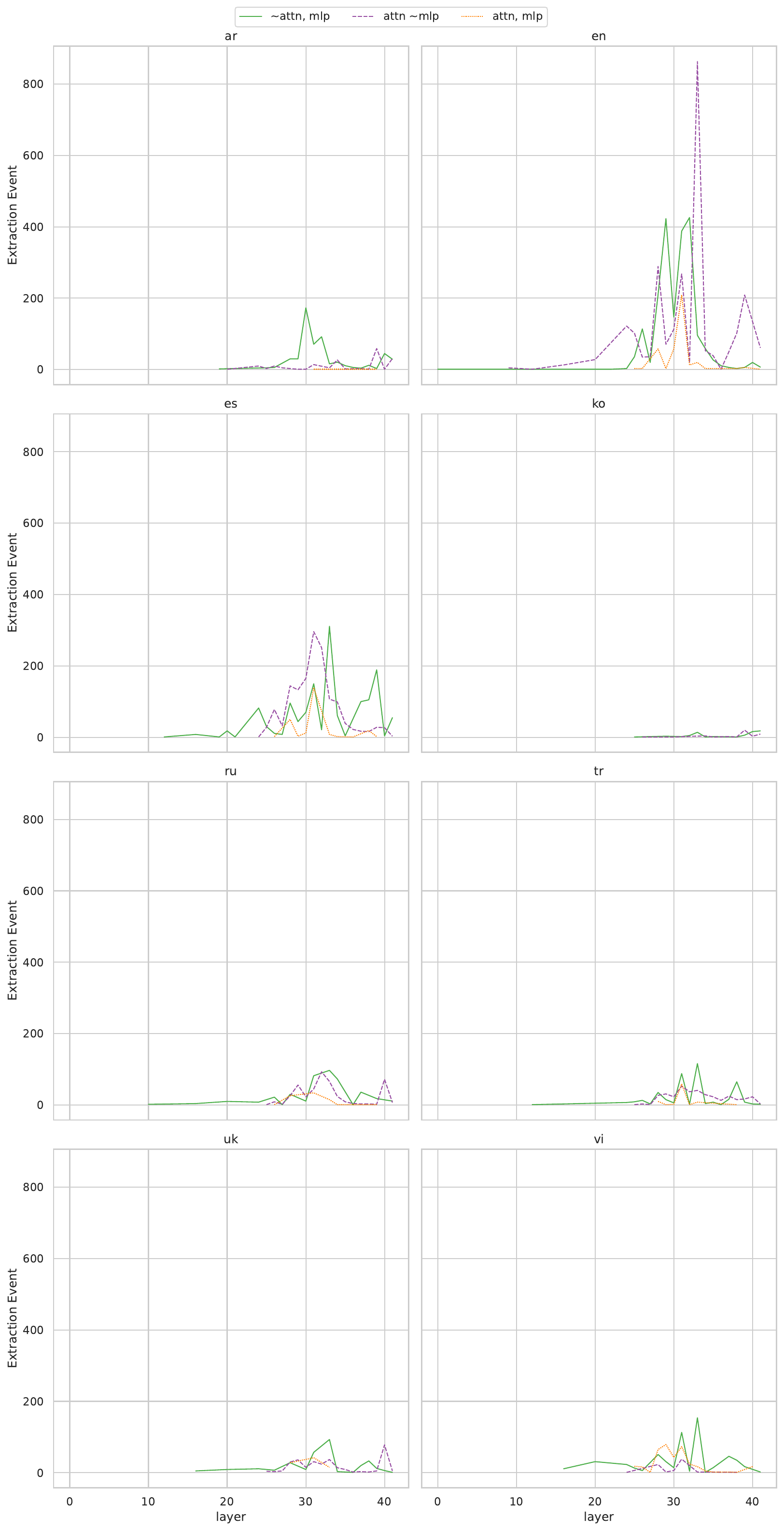} 
    \caption{Number of extraction events split by precedence (or not) of an extraction event in the self-attn in \eurollm{} for each language.}
\end{figure*}

\begin{figure*}[h!]
    \centering
    \includegraphics[width=0.8\textwidth]{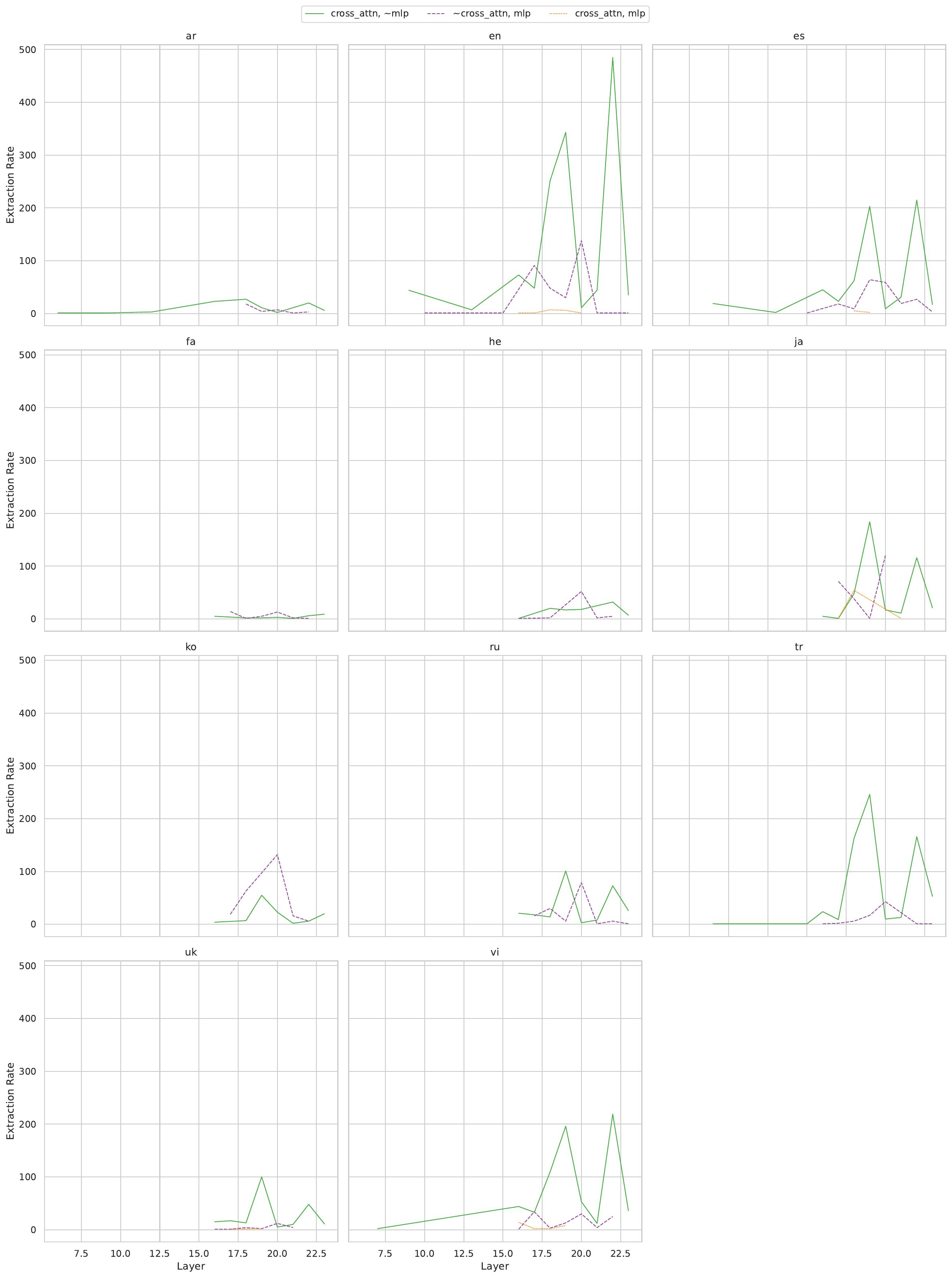} 
    \caption{Number of extraction events split by precedence (or not) of an extraction event in the cross-attn in \mtfive{} for each language.}
\end{figure*}

\clearpage
\section{Patching}

\begin{minipage}{\textwidth}
\centering
\begin{minipage}[t]{0.6\linewidth}
    \centering
    \includegraphics[width=\linewidth]{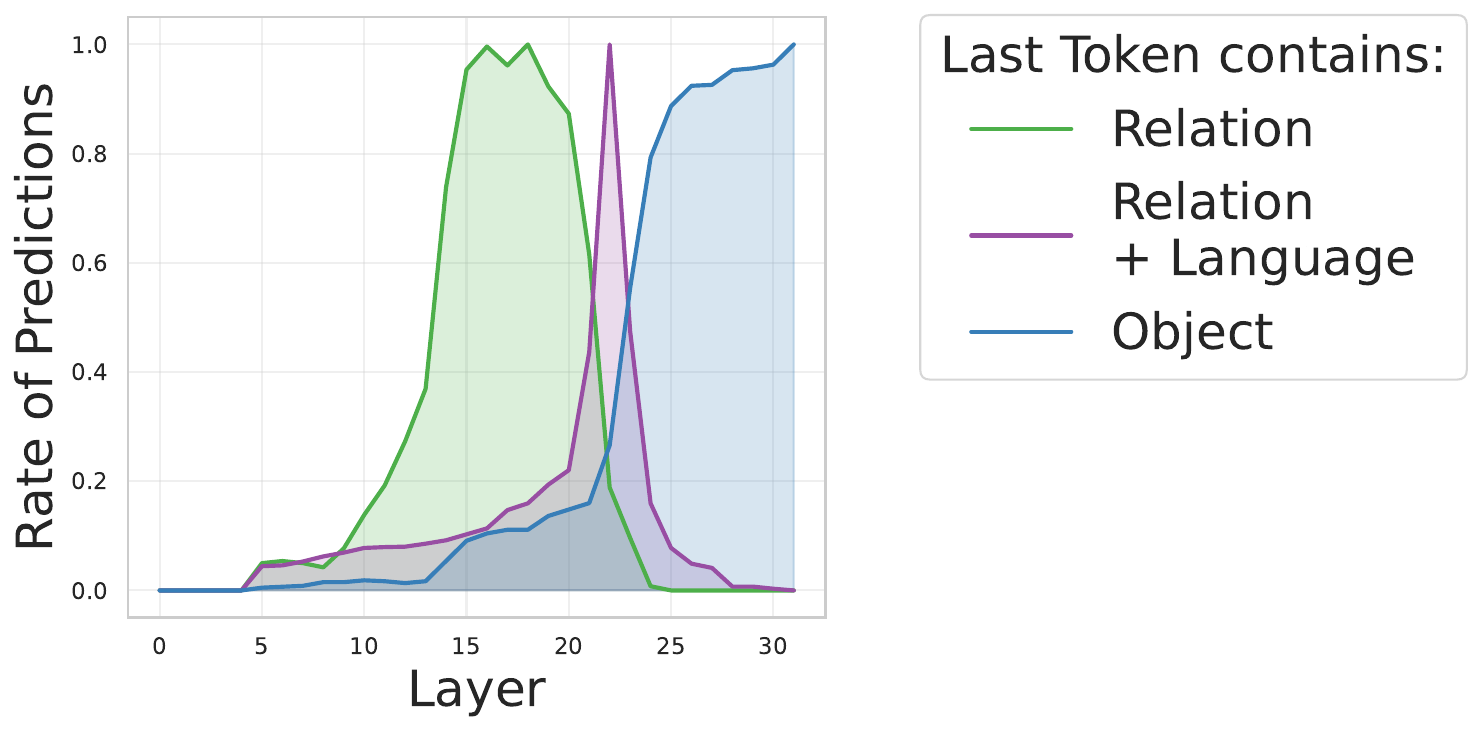}
\end{minipage}
  \captionof{figure}{Aggregated patching results (\S\ref{sec:patching}) for \xglm{}. The last token representation contains the function that solves the task. This function is formed in two stages: first, the relation to extract is encoded (green), and then the language is composed into the function (purple). Finally, the function is applied to the subject, and the predicted object is resolved (blue).}\label{fig:summary_patching_xglm}
\end{minipage}

\begin{minipage}{\textwidth}
\vspace{2em}
\centering\footnotesize\setlength{\tabcolsep}{3pt}
\begin{tabular}{lccccccccc}
\toprule
& \multicolumn{3}{c}{\xglm{}} & \multicolumn{3}{c}{\eurollm{}} & \multicolumn{3}{c}{\mtfive{}} \\
\cmidrule(lr){2-4} \cmidrule(lr){5-7} \cmidrule(lr){8-10}
Patch - Context & All & \(\mathcal{L}_c(o_p)\) & \(\mathcal{L}_p(o_c)\) & All & \(\mathcal{L}_c(o_p)\) & \(\mathcal{L}_p(o_c)\) & All & \(\mathcal{L}_c(o_p)\) & \(\mathcal{L}_p(o_c)\) \\
\midrule
en-es & 849 & 834 & 759 & 880 & 862 & 755 & 842 & 831 & 780 \\
en-vi & 862 & 857 & 836 & 930 & 910 & 843 & 854 & 844 & 825 \\
en-tr & 974 & 961 & 961 & 966 & 945 & 914 & 815 & 803 & 794 \\
en-ru & 823 & 821 & 817 & 846 & 846 & 827 & 796 & 795 & 787 \\
en-uk & 914 & 914 & 911 & 915 & 915 & 915 & 853 & 850 & 852 \\
en-ko & 995 & 995 & 991 & 991 & 991 & 991 & 744 & 744 & 743 \\
en-he & 962 & 962 & 961 & 810 & 810 & 775 & 778 & 778 & 778 \\
en-fa & 537 & 537 & 532 & 771 & 771 & 734 & 714 & 714 & 712 \\
en-ar & 829 & 829 & 828 & 862 & 859 & 858 & 775 & 775 & 774 \\
en-ja & 134 & 134 & 134 & 895 & 895 & 872 & 774 & 774 & 773 \\
\bottomrule
\end{tabular}
\captionof{table}{Total number of patch-context examples considered in the patching experiments with \(\{\neq \mathcal{L}, = r , \neq s\}\). The \(\mathcal{L}_c(o_p)\) column is the total number of examples where the detection of \(\mathcal{L}_c(o_p)\) would be unambiguous, that is, \(\mathcal{L}_c(o_p) \ne \mathcal{L}_p(o_p)\), conversely for the \(\mathcal{L}_p(o_c)\) column.}
\label{appendix:table_counts_same_r}
\vspace{4em}
\end{minipage}

\begin{minipage}{\textwidth}
\centering\footnotesize\setlength{\tabcolsep}{3pt}
\begin{tabular}{lcccccc}
\toprule
& \multicolumn{2}{c}{\xglm{}} & \multicolumn{2}{c}{\eurollm{}} & \multicolumn{2}{c}{\mtfive{}} \\
\cmidrule(lr){2-3} \cmidrule(lr){4-5} \cmidrule(lr){6-7}
Patch - Context & \(\mathcal{L}_c(o_p)\) & \(\mathcal{L}_p(o_c)\) & \(\mathcal{L}_c(o_p)\) & \(\mathcal{L}_p(o_c)\) & \(\mathcal{L}_c(o_p)\) & \(\mathcal{L}_p(o_c)\) \\
\midrule
en-es & 0.7\% (6) & \textbf{26.0\% (197)} & 5.0\% (43) & \textbf{34.2\% (258)} & 0.5\% (4) & \textbf{21.5\% (168)}\\
en-vi & 0.5\% (4) & \textbf{34.4\% (288)} & 0.5\% (5) & \textbf{16.1\% (136)} & 0.1\% (1) & \textbf{20.2\% (167)}\\
en-tr & 0.2\% (2) & \textbf{16.4\% (158)} & 0.2\% (2) & \textbf{25.4\% (232)} & 0.4\% (3) & \textbf{26.2\% (208)}\\
en-ru & 2.2\% (18) & \textbf{51.4\% (420)} & 4.6\% (39) & \textbf{63.5\% (525)} & 1.1\% (9) & \textbf{36.1\% (284)}\\
en-uk & 0.5\% (5) & \textbf{21.3\% (194)} & 6.9\% (63) & \textbf{62.5\% (572)} & 0.2\% (2) & \textbf{31.8\% (271)}\\
en-ko & 0.3\% (3) & \textbf{47.1\% (467)} & 0.1\% (1) & \textbf{75.7\% (750)} & 0.3\% (2) & \textbf{32.3\% (240)}\\
en-he & 0.0\% (0) & \textbf{27.3\% (262)} & 0.0\% (0) & \textbf{65.9\% (511)} & 0.4\% (3) & \textbf{38.4\% (299)}\\
en-fa & 0.0\% (0) & \textbf{41.7\% (222)} & 0.3\% (2) & \textbf{64.0\% (470)} & 1.0\% (7) & \textbf{25.1\% (179)}\\
en-ar & 8.7\% (72) & \textbf{39.5\% (327)} & 4.5\% (39) & \textbf{65.7\% (564)} & 0.6\% (5) & \textbf{31.9\% (247)}\\
en-ja & 0.0\% (0) & \textbf{3.0\% (4)} & 1.9\% (17) & \textbf{11.5\% (100)} & 0.0\% (0) & \textbf{26.0\% (201)}\\
\bottomrule
\end{tabular}
\captionof{table}{Proportion of times an object is predicted in the other language in the patching experiments with \(\{\neq \mathcal{L}, = r , \neq s\}\). In parenthesis the number of examples corresponding to the percentage. In bold when \(\mathcal{L}_c(o_p)\) or \(\mathcal{L}_p(o_c)\) are detected more often for each of the experiments. The total number of examples varies, see total numbers in Table~\ref{appendix:table_counts_same_r}.}
\label{appendix:table_swapped_counts_same_r}
\vspace{4em}
\end{minipage}

\clearpage
\begin{minipage}{\textwidth}
\vspace{1em}
\centering\footnotesize\setlength{\tabcolsep}{3pt}
\begin{tabular}{lccccccccc}
\toprule
& \multicolumn{3}{c}{\xglm{}} & \multicolumn{3}{c}{\eurollm{}} & \multicolumn{3}{c}{\mtfive{}} \\
\cmidrule(lr){2-4} \cmidrule(lr){5-7} \cmidrule(lr){8-10}
Patch - Context & All & \(\mathcal{L}_c(o_p)\) & \(\mathcal{L}_p(o_c)\) & All & \(\mathcal{L}_c(o_p)\) & \(\mathcal{L}_p(o_c)\) & All & \(\mathcal{L}_c(o_p)\) & \(\mathcal{L}_p(o_c)\) \\
\midrule
en-es & 324 & 284 & 298 & 413 & 360 & 391 & 263 & 253 & 249 \\
en-vi & 336 & 329 & 333 & 105 & 102 & 100 & 209 & 207 & 204 \\
en-tr & 88 & 77 & 84 & 111 & 102 & 110 & 209 & 207 & 204 \\
en-ru & 218 & 216 & 218 & 420 & 415 & 419 & 191 & 191 & 190 \\
en-uk & 35 & 35 & 35 & 351 & 347 & 350 & 136 & 136 & 133 \\
en-ko & 46 & 46 & 46 & 98 & 98 & 98 & 165 & 165 & 165 \\
en-ar & 212 & 212 & 212 & 391 & 390 & 391 & 159 & 159 & 159 \\
en-he & - & - & - & 19 & 19 & 19 & 166 & 166 & 166 \\
en-ja & - & - & - & 1 & 1 & 1 & 85 & 85 & 85 \\
en-fa & - & - & - & - & - & - & 114 & 114 & 114 \\
\bottomrule
\end{tabular}
\captionof{table}{Total number of patch-context examples considered in the patching experiments with \(\{\neq \mathcal{L}, \neq r , = s\}\). The \(\mathcal{L}_c(o_p)\) column is the total number of examples where the detection of \(\mathcal{L}_c(o_p)\) would be unambiguous, that is, \(\mathcal{L}_c(o_p) \ne \mathcal{L}_p(o_p)\), conversely for the \(\mathcal{L}_p(o_c)\) column.}
\label{appendix:table_counts_diff_r}
\vspace{4em}
\end{minipage}

\begin{minipage}{\textwidth}
\vspace{1em}
\centering\footnotesize\setlength{\tabcolsep}{3pt}
\begin{tabular}{lcccccc}
\toprule
& \multicolumn{2}{c}{\xglm{}} & \multicolumn{2}{c}{\eurollm{}} & \multicolumn{2}{c}{\mtfive{}} \\
\cmidrule(lr){2-3} \cmidrule(lr){4-5} \cmidrule(lr){6-7}
Patch - Context & \(\mathcal{L}_c(o_p)\) & \(\mathcal{L}_p(o_c)\) & \(\mathcal{L}_c(o_p)\) & \(\mathcal{L}_p(o_c)\) & \(\mathcal{L}_c(o_p)\) & \(\mathcal{L}_p(o_c)\) \\
\midrule
en-es & \textbf{18.0\% (51)} & 3.7\% (11) & \textbf{41.7\% (150)} & 2.8\% (11) & 0.8\% (2) & \textbf{7.6\% (19)} \\
en-vi & \textbf{9.1\% (30)} & 6.6\% (22) & 3.9\% (4) & \textbf{8.0\% (8)} & 1.0\% (2) & \textbf{12.3\% (25)} \\
en-tr & \textbf{36.4\% (28)} & 6.0\% (5) & \textbf{56.9\% (58)} & 4.5\% (5) & 2.9\% (6) & \textbf{9.3\% (19)} \\
en-ru & \textbf{64.8\% (140)} & 4.1\% (9) & \textbf{61.7\% (256)} & 9.3\% (39) & 6.3\% (12) & \textbf{8.9\% (17)} \\
en-uk & 17.1\% (6) & \textbf{17.1\% (6)} & \textbf{66.0\% (229)} & 2.3\% (8) & 3.7\% (5) & \textbf{9.0\% (12)} \\
en-ko & \textbf{41.3\% (19)} & 21.7\% (10) & \textbf{71.4\% (70)} & 5.1\% (5) & 2.4\% (4) & \textbf{12.1\% (20)} \\
en-ar & \textbf{40.1\% (85)} & 1.9\% (4) & \textbf{52.8\% (206)} & 0.8\% (3) & 1.9\% (3) & \textbf{7.5\% (12)} \\
en-he & - & - & 0.0\% (0) & 0.0\% (0) & 3.6\% (6) & \textbf{15.7\% (26)} \\
en-ja & - & - & 0.0\% (0) & 0.0\% (0) & 0.0\% (0) & \textbf{25.9\% (22)} \\
en-fa & - & - & - & - & 2.6\% (3) & \textbf{15.8\% (18)} \\
\bottomrule
\end{tabular}
\captionof{table}{Proportion of times an object is predicted in the other language in the patching experiments with \(\{\neq \mathcal{L}, \neq r , = s\}\). In parenthesis the number of examples corresponding to the percentage. In bold when \(\mathcal{L}_c(o_p)\) or \(\mathcal{L}_p(o_c)\) are detected more often for each of the experiments. The total number of examples varies, see total numbers in Table~\ref{appendix:table_counts_diff_r}.}
\label{appendix:table_swapped_counts_diff_r}
\vspace{4em}
\end{minipage}

\clearpage
\subsection{Different Relation, Different Subject}

\begin{minipage}{\textwidth}
    \centering
    \vspace{5mm}
    \begin{minipage}{\linewidth}
        \centering
        \includegraphics[width=0.45\linewidth]{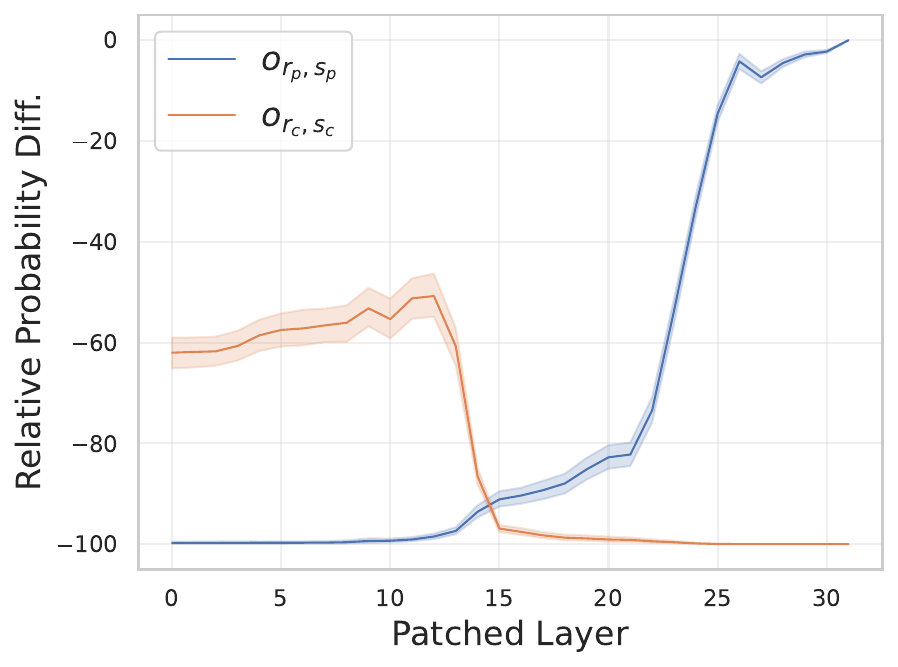}
    \end{minipage}
    \vspace{5mm}
    \begin{minipage}{\linewidth}
        \centering
        \includegraphics[width=0.45\linewidth]{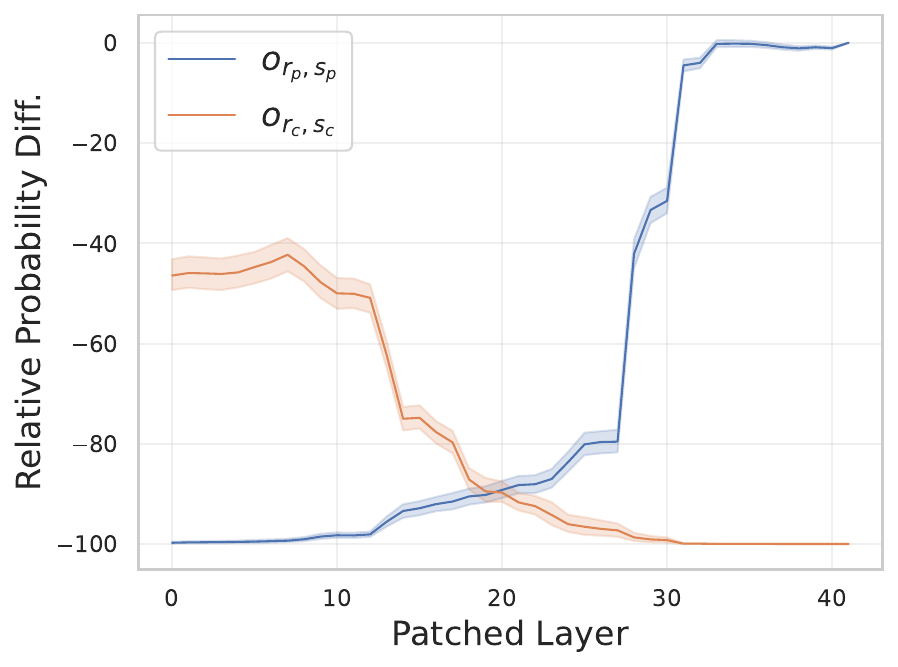}
    \end{minipage}
    \vspace{5mm} 
    \begin{minipage}{\linewidth}
        \centering
        \includegraphics[width=0.45\linewidth]{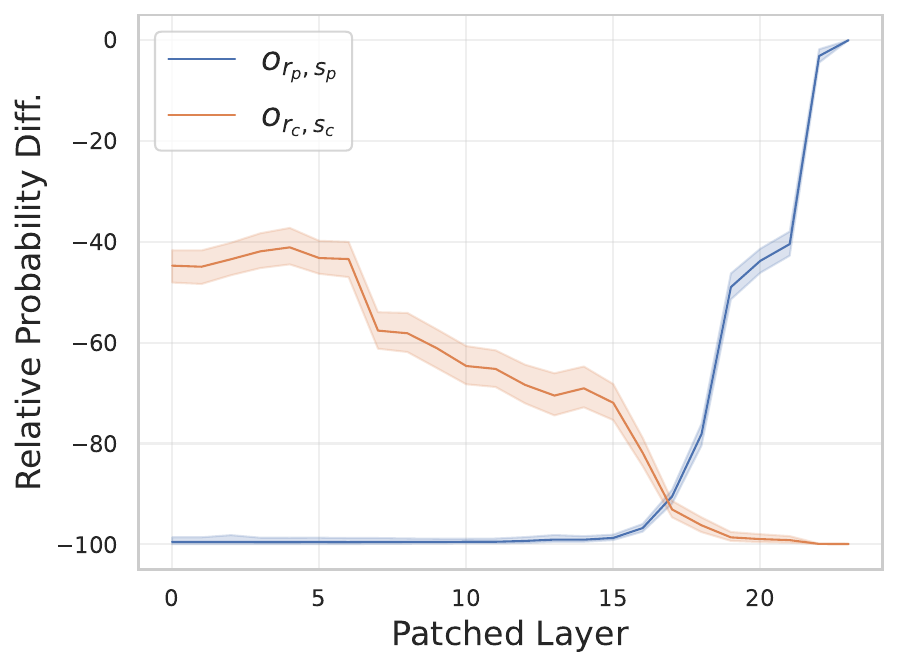}
    \end{minipage}
    \vspace{-2mm}
\captionof{figure}{Probability of the patch answer and the context answer when patching at different layers. Models from top to bottom: \xglm{}, \eurollm, \mtfive{}.}\label{fig:en_all_diff_results_probs}
\vspace{4mm}
\end{minipage}

\clearpage
\subsection{Same Relation, Different Subject}\label{appendix:same_r_different_s}

\begin{minipage}{\textwidth}
    \centering
    \vspace{5mm}
    \begin{minipage}{\linewidth}
        \centering
        \includegraphics[width=0.45\linewidth]{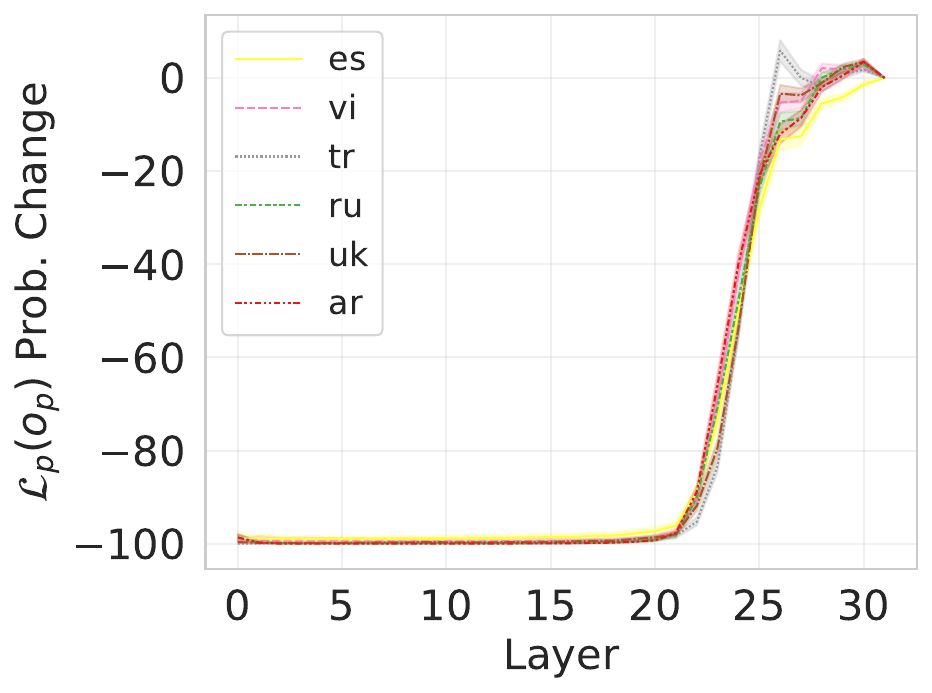}
    \end{minipage}
    \vspace{5mm}
    \begin{minipage}{\linewidth}
        \centering
        \includegraphics[width=0.45\linewidth]{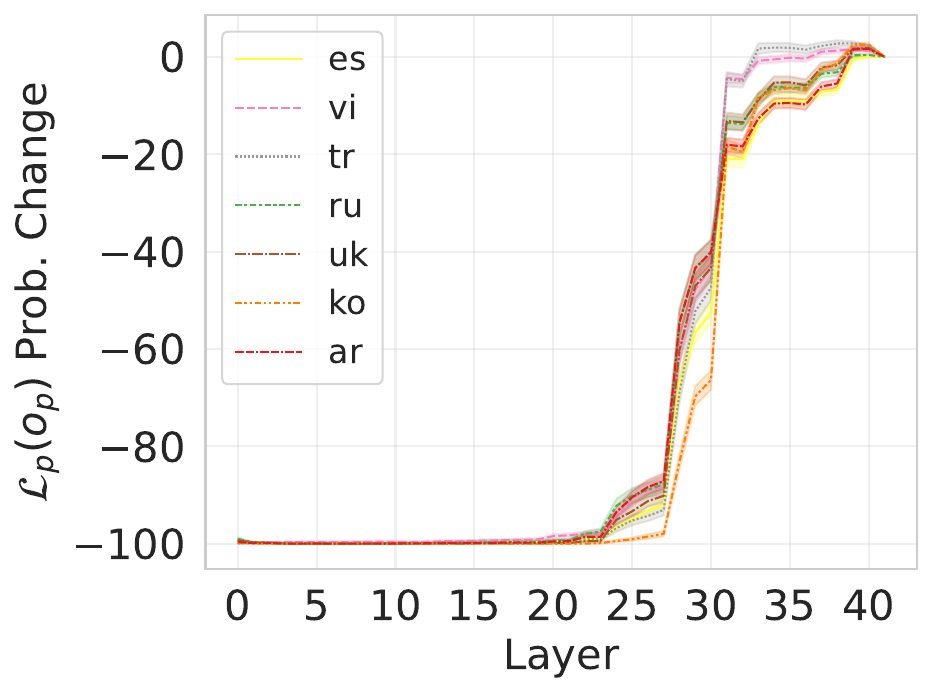}
    \end{minipage}
    \vspace{5mm} 
    \begin{minipage}{\linewidth}
        \centering
        \includegraphics[width=0.45\linewidth]{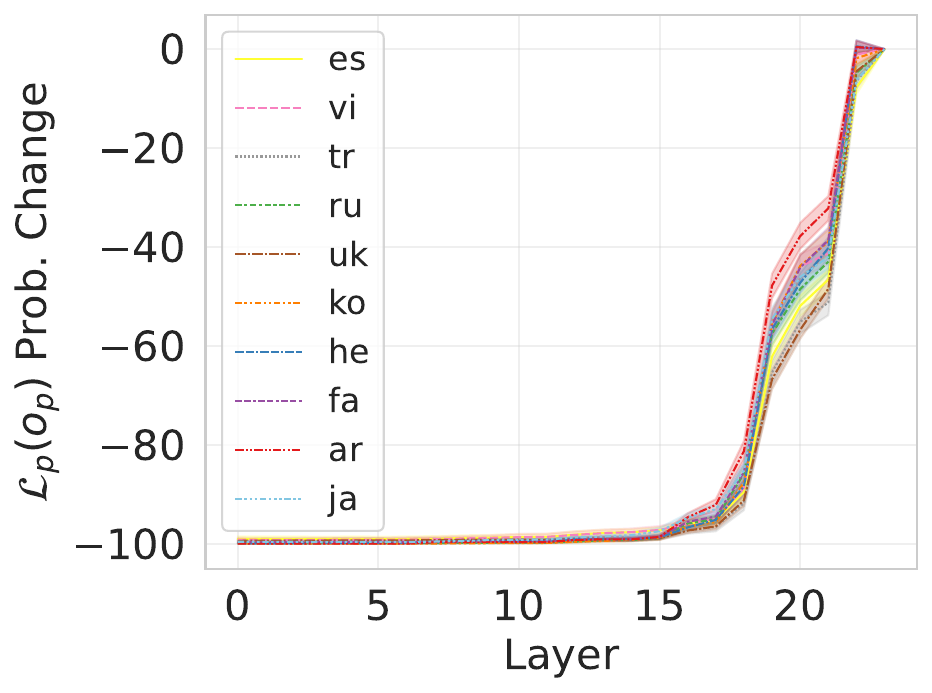}
    \end{minipage}
    \vspace{-2mm}
\captionof{figure}{Probability of the patch answer \(\mathcal{L}_p(o_p)\) when patching at different layers. Models from top to bottom: \xglm{}, \eurollm, \mtfive{}.}
\vspace{4mm}
\end{minipage}

\begin{figure*}[h!]
    \centering
    \includegraphics[width=0.9\textwidth]{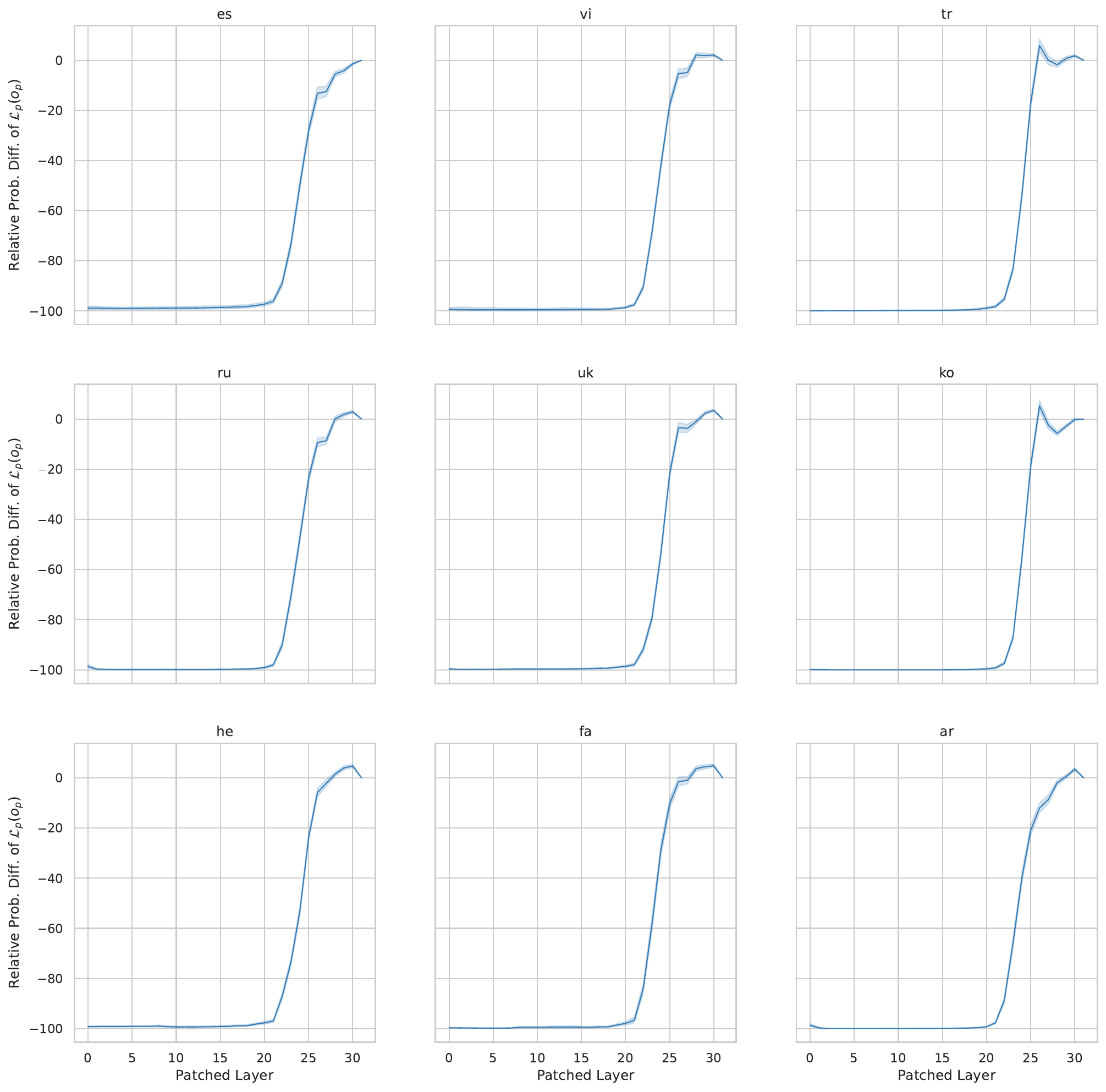} 
    \caption{Probability of the patch answer \(\mathcal{L}_p(o_p)\) when patching at different layers in \xglm{}, for examples with \(\{\ne \mathcal{L}, = r, \ne s\}\).}
    \label{fig:prob_patch_each_lang}
\end{figure*}

\begin{figure*}[h!]
    \centering
    \includegraphics[width=0.9\textwidth]{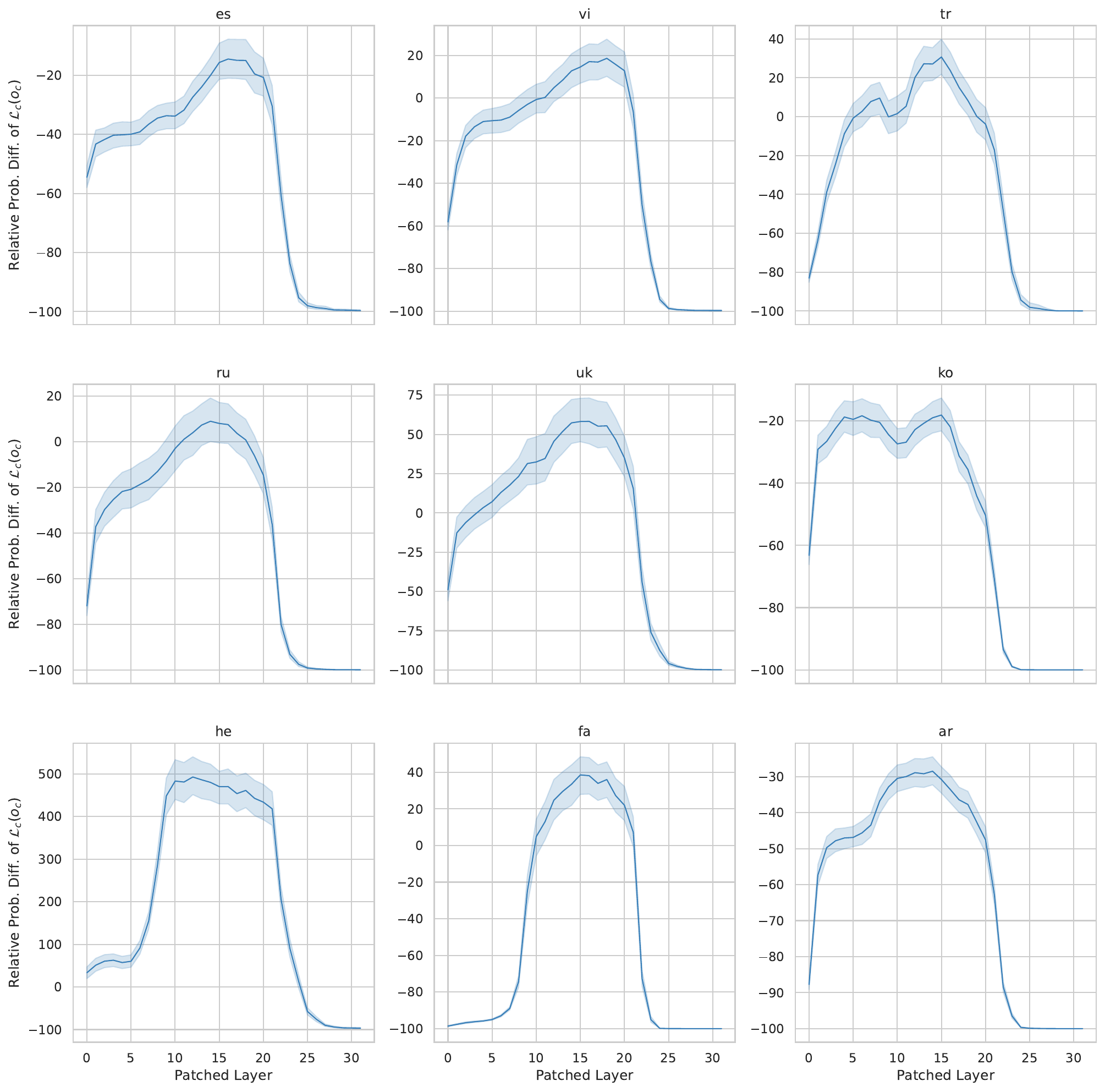} 
    \caption{Probability of the \(\mathcal{L}_c(o_c)\) when patching at different layers in \xglm{}, for examples with \(\{\ne \mathcal{L}, = r, \ne s\}\). Note that the plots do not share the y-axis.}
    \label{fig:prob_context_each_lang}
\end{figure*}

\begin{figure*}[h!]
    \centering
    \includegraphics[width=0.9\textwidth]{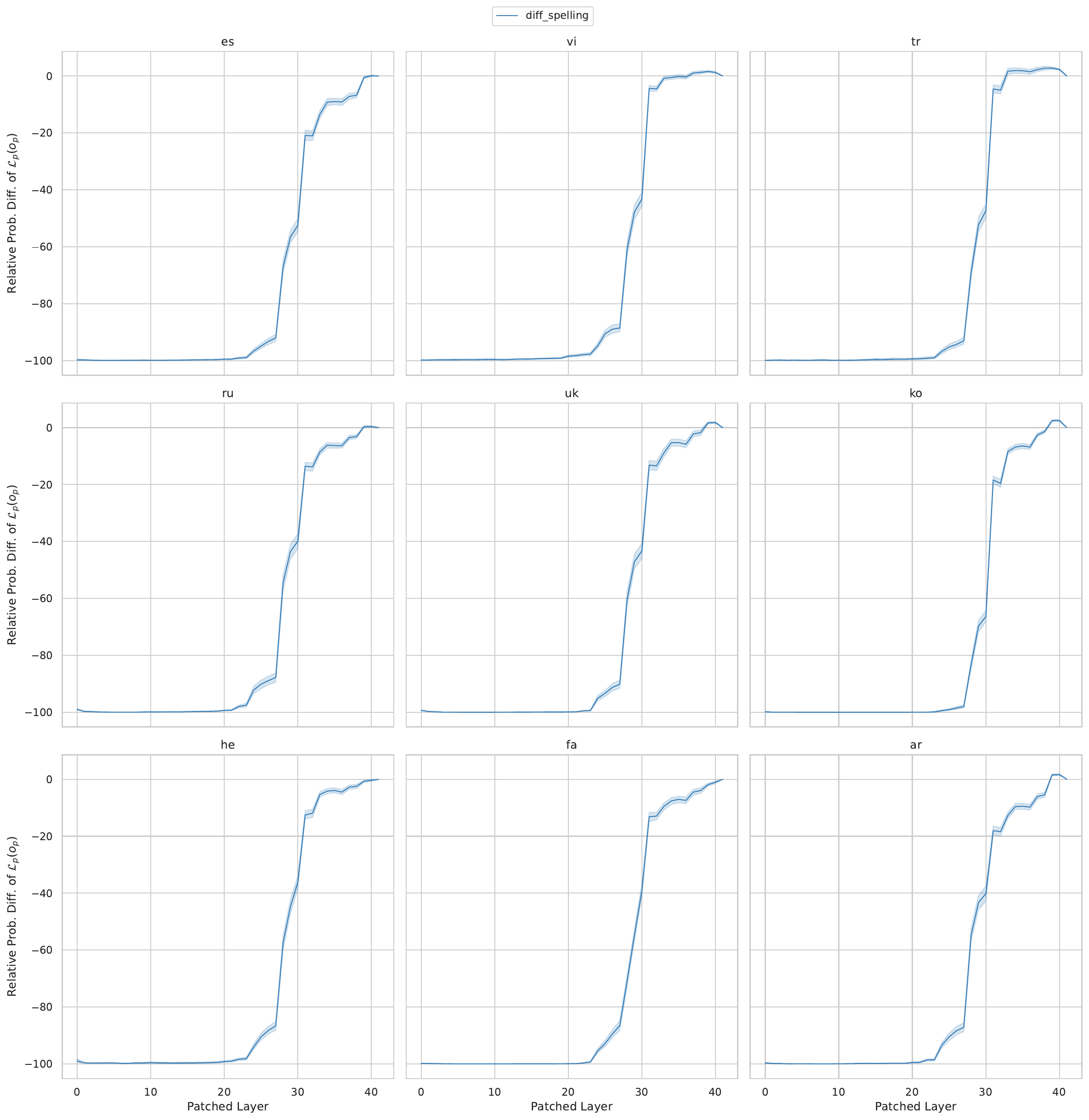} 
    \caption{Probability of the patch answer \(\mathcal{L}_p(o_p)\) when patching at different layers in \eurollm{}, for examples with \(\{\ne \mathcal{L}, = r, \ne s\}\).}
    \label{fig:eurollm_prob_patch_each_lang}
\end{figure*}

\begin{figure*}[h!]
    \centering
    \includegraphics[width=0.9\textwidth]{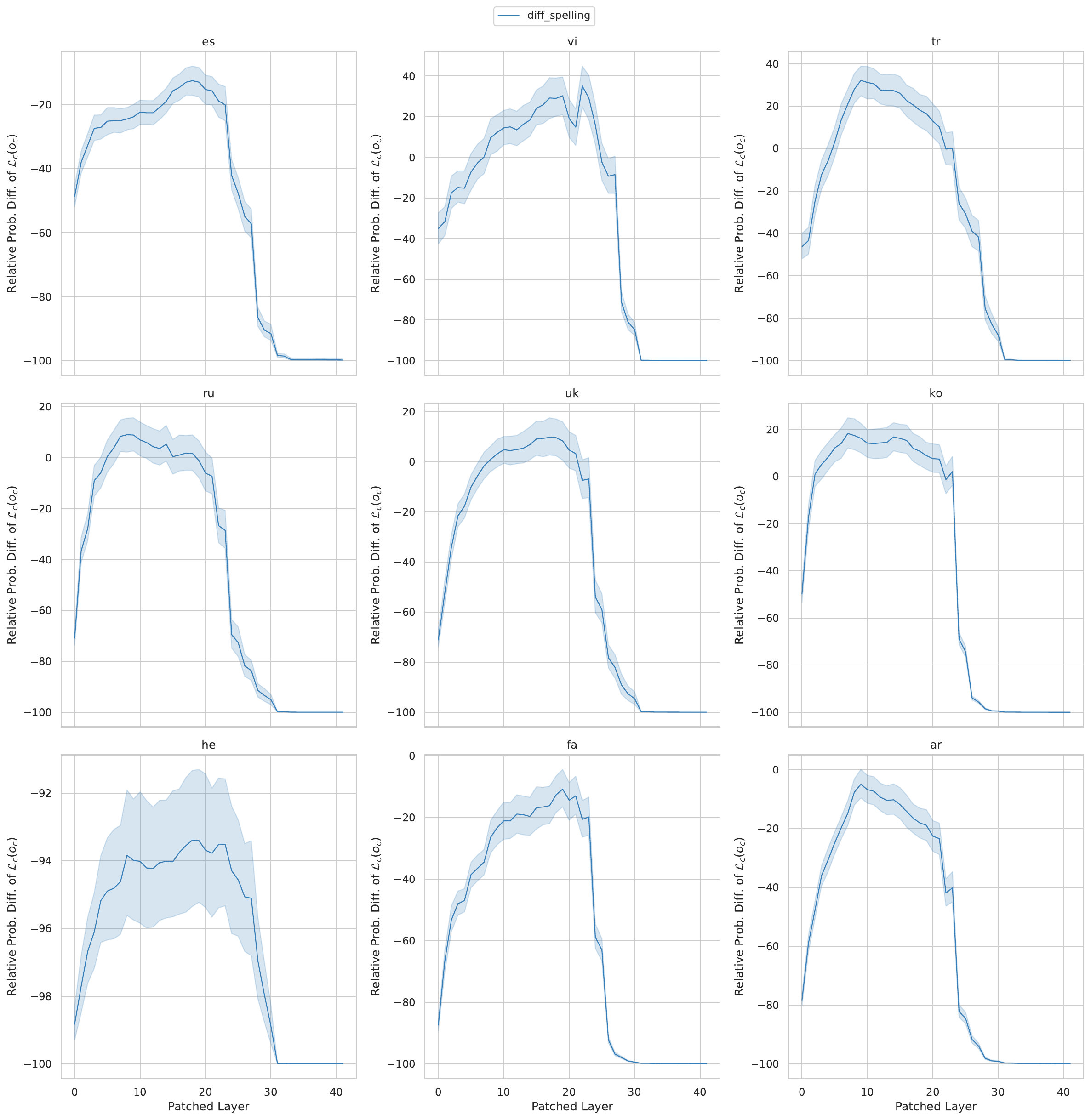} 
    \caption{Probability of the \(\mathcal{L}_c(o_c)\) when patching at different layers in \eurollm{}, for examples with \(\{\ne \mathcal{L}, = r, \ne s\}\). Note that the plots do not share the y-axis.}
    \label{fig:eurollm_prob_context_each_lang}
\end{figure*}

\begin{figure*}[h!]
    \centering
    \includegraphics[width=0.9\textwidth]{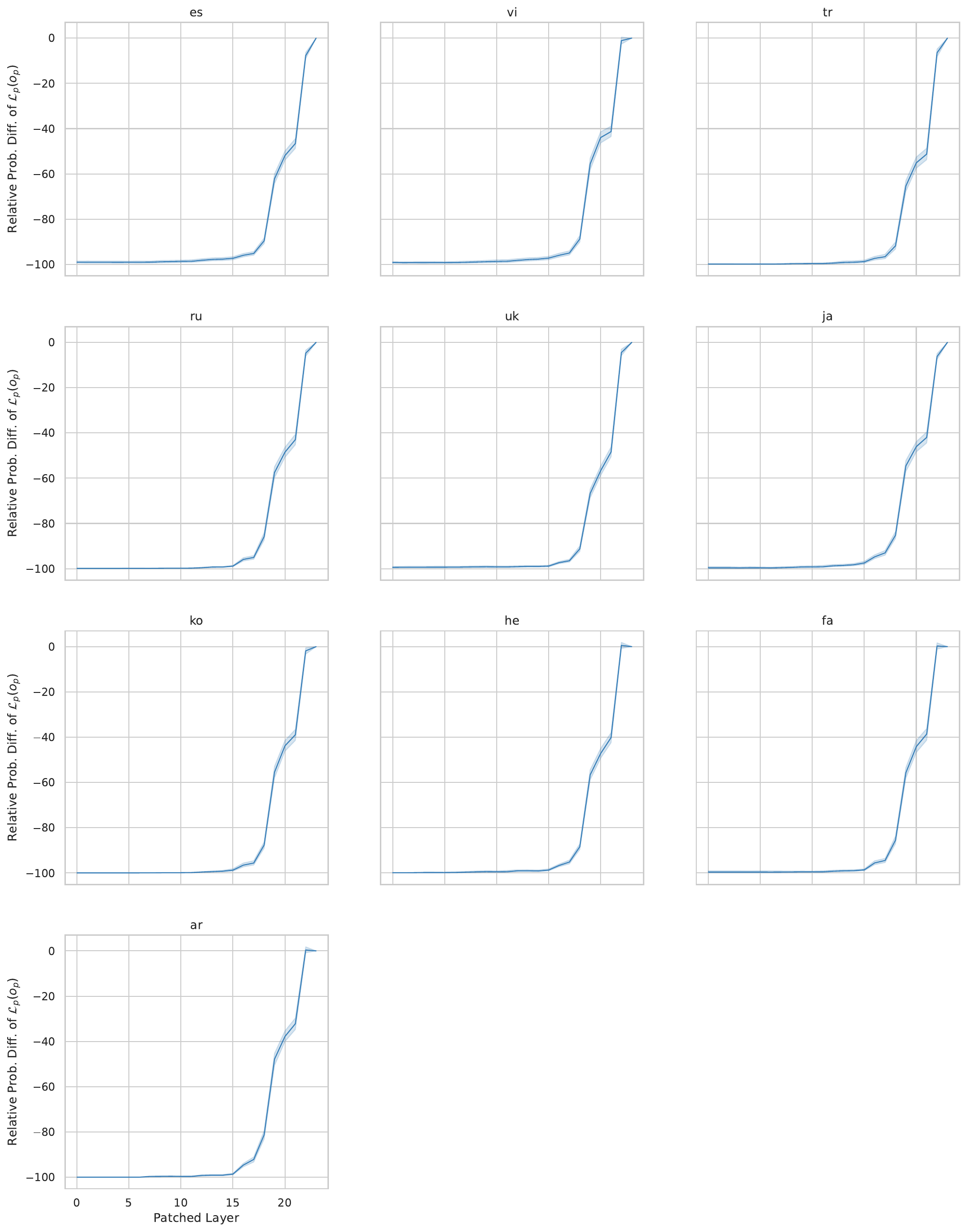} 
    \caption{Probability of the \(\mathcal{L}_p(o_p)\) when patching at different layers in \mtfive{}, for examples with \(\{\ne \mathcal{L}, = r, \ne s\}\).}
    \label{fig:mt5_prob_patch_each_lang}
\end{figure*}

\begin{figure*}[h!]
    \centering
    \includegraphics[width=0.9\textwidth]{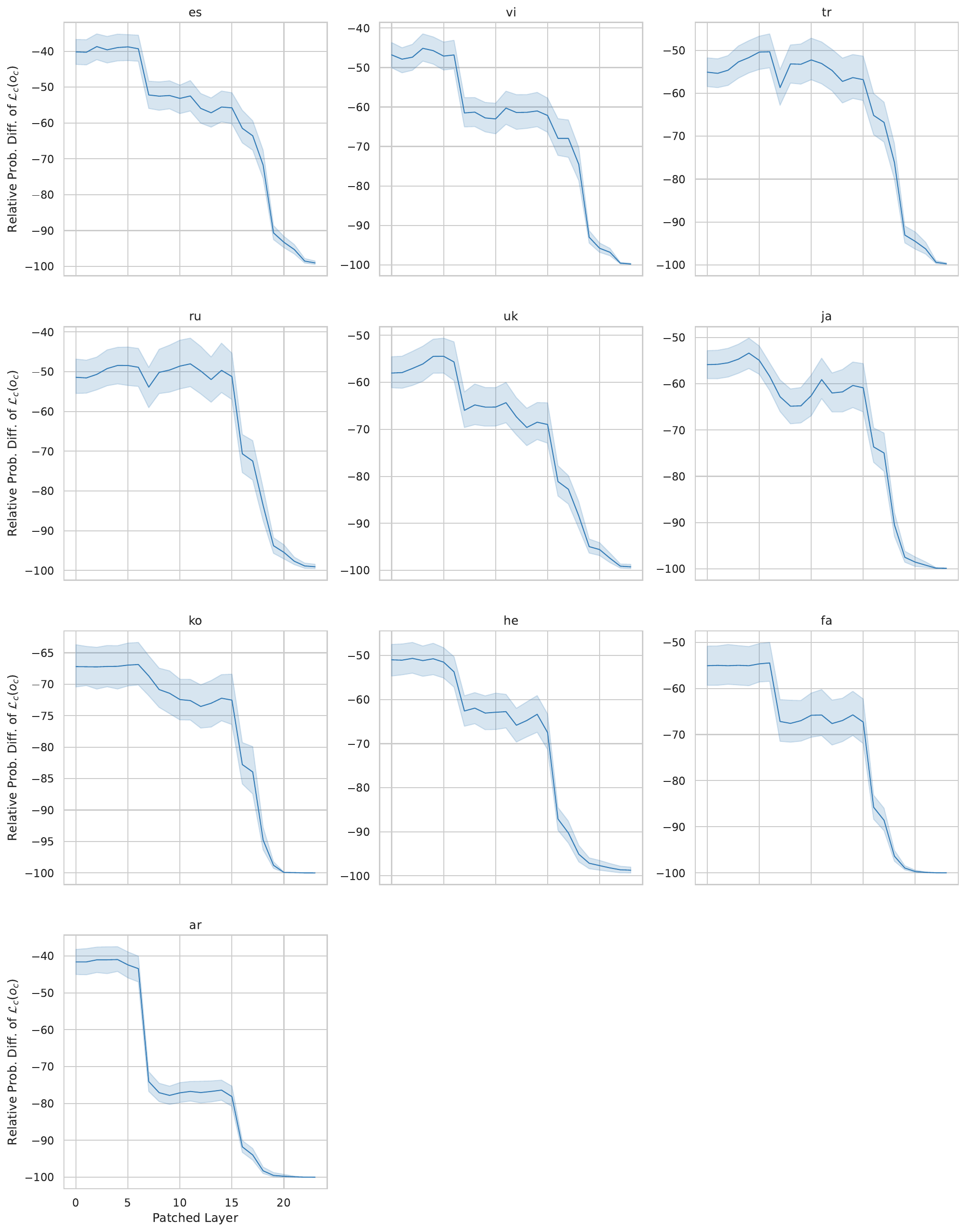} 
    \caption{Probability of the \(\mathcal{L}_c(o_c)\) when patching at different layers in \mtfive{}, for examples with \(\{\ne \mathcal{L}, = r, \ne s\}\). Note that the plots do not share the y-axis.}
    \label{fig:mt5_prob_context_each_lang}
\end{figure*}

\clearpage

\subsection{Different Relation, Same Subject}\label{appendix:different_r_same_s}

\begin{minipage}{\textwidth}
    \centering
    \vspace{5mm}
    \begin{minipage}{\linewidth}
        \centering
        \includegraphics[width=0.45\linewidth]{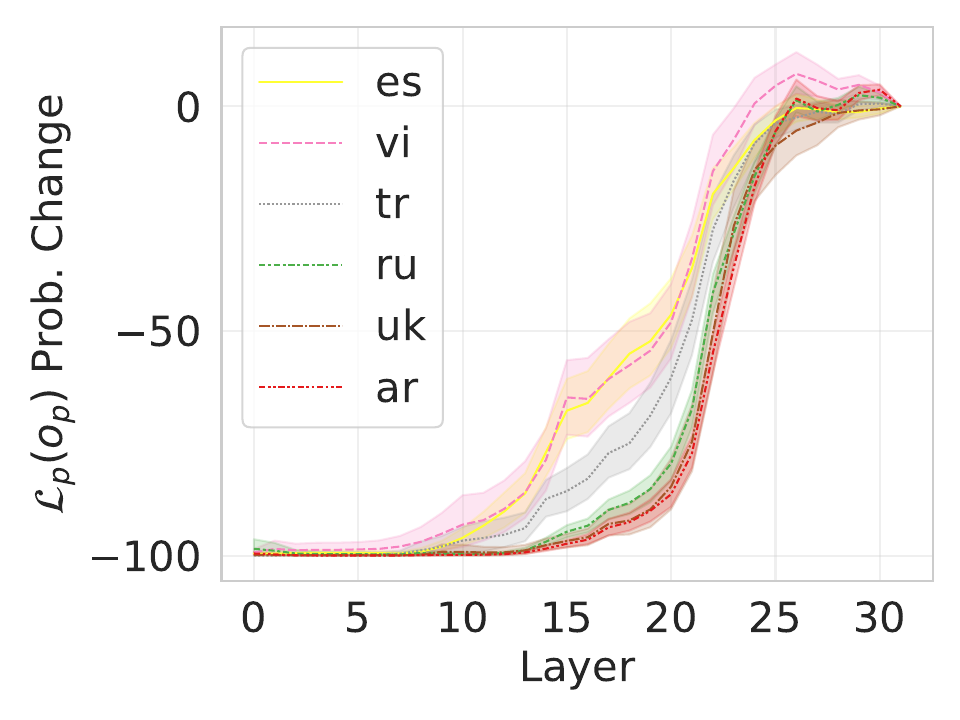}
    \end{minipage}
    \vspace{5mm}
    \begin{minipage}{\linewidth}
        \centering
        \includegraphics[width=0.45\linewidth]{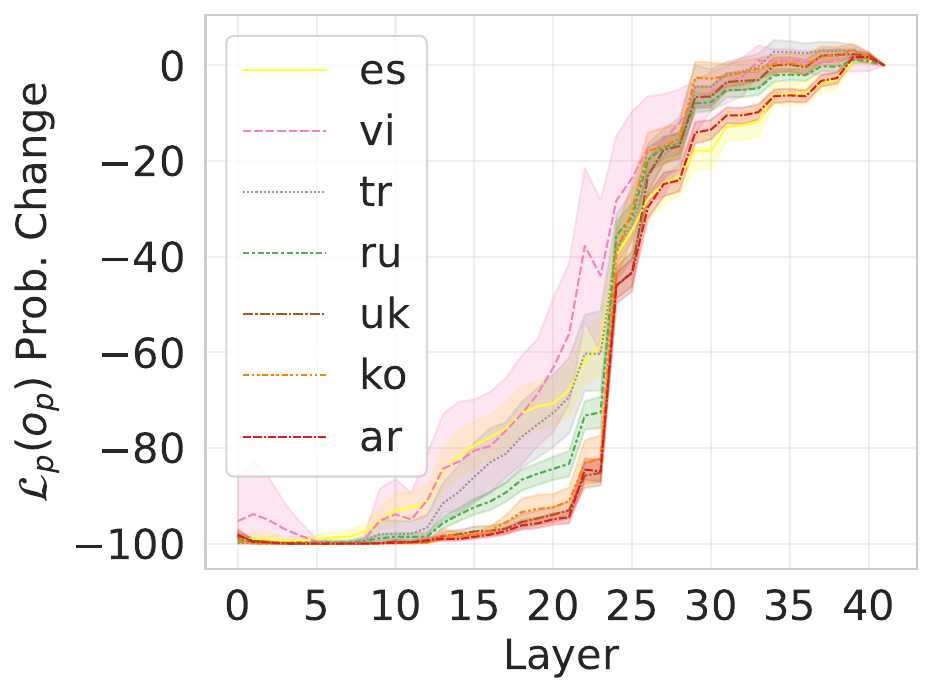}
    \end{minipage}
    \vspace{5mm} 
    \begin{minipage}{\linewidth}
        \centering
        \includegraphics[width=0.45\linewidth]{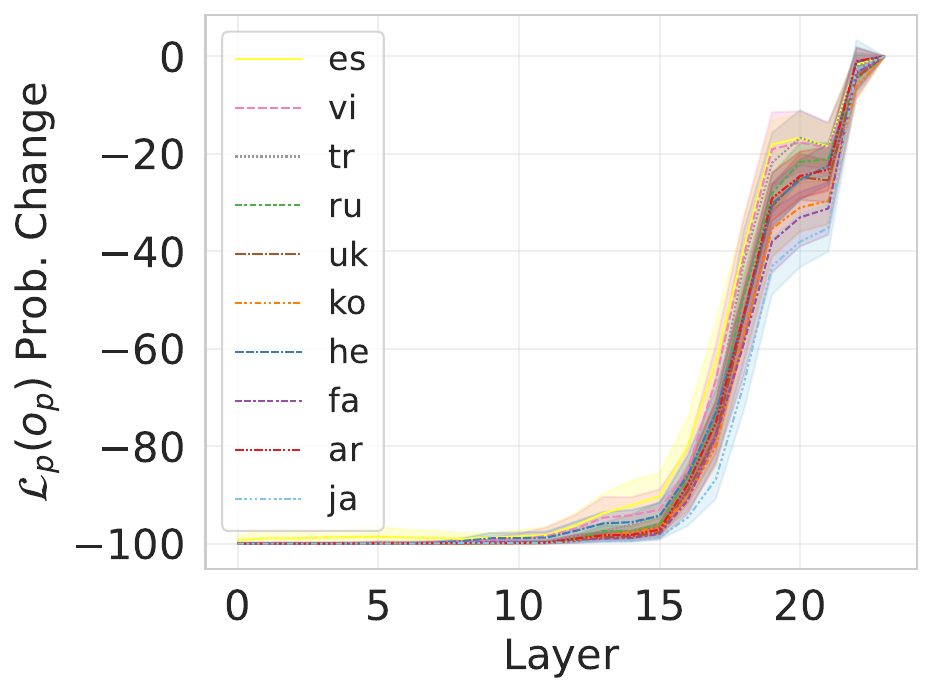}
    \end{minipage}
    \vspace{-2mm}
\captionof{figure}{Probability of the patch answer \(\mathcal{L}_p(o_p)\) when patching at different layers for examples with \(\{\ne \mathcal{L}, \ne r, = s\}\). Models from top to bottom: \xglm{}, \eurollm, \mtfive{}.}\label{fig:diff_r_same_s_patch_prob}
\vspace{4mm}
\end{minipage}

\begin{figure*}[h!]
    \centering
    \includegraphics[width=0.9\textwidth]{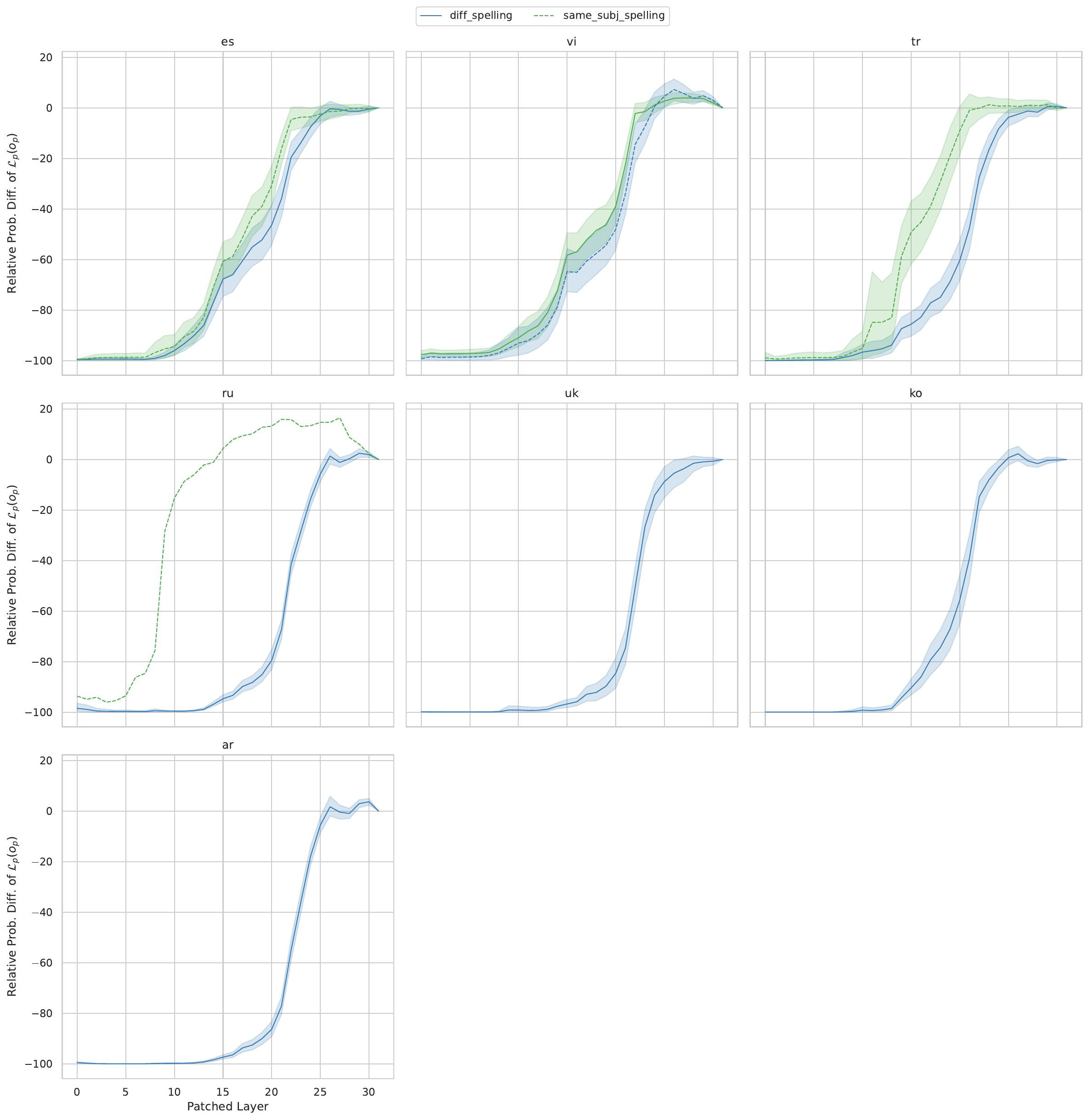} 
    \caption{Probability of the \(\mathcal{L}_p(o_p)\) when patching at different layers in \xglm{}, for examples with \(\{\ne \mathcal{L}, \ne r, = s\}\).}
    \label{fig:xglm_diff_r_prob_patch_each_lang}
\end{figure*}

\begin{figure*}[h!]
    \centering
    \includegraphics[width=0.9\textwidth]{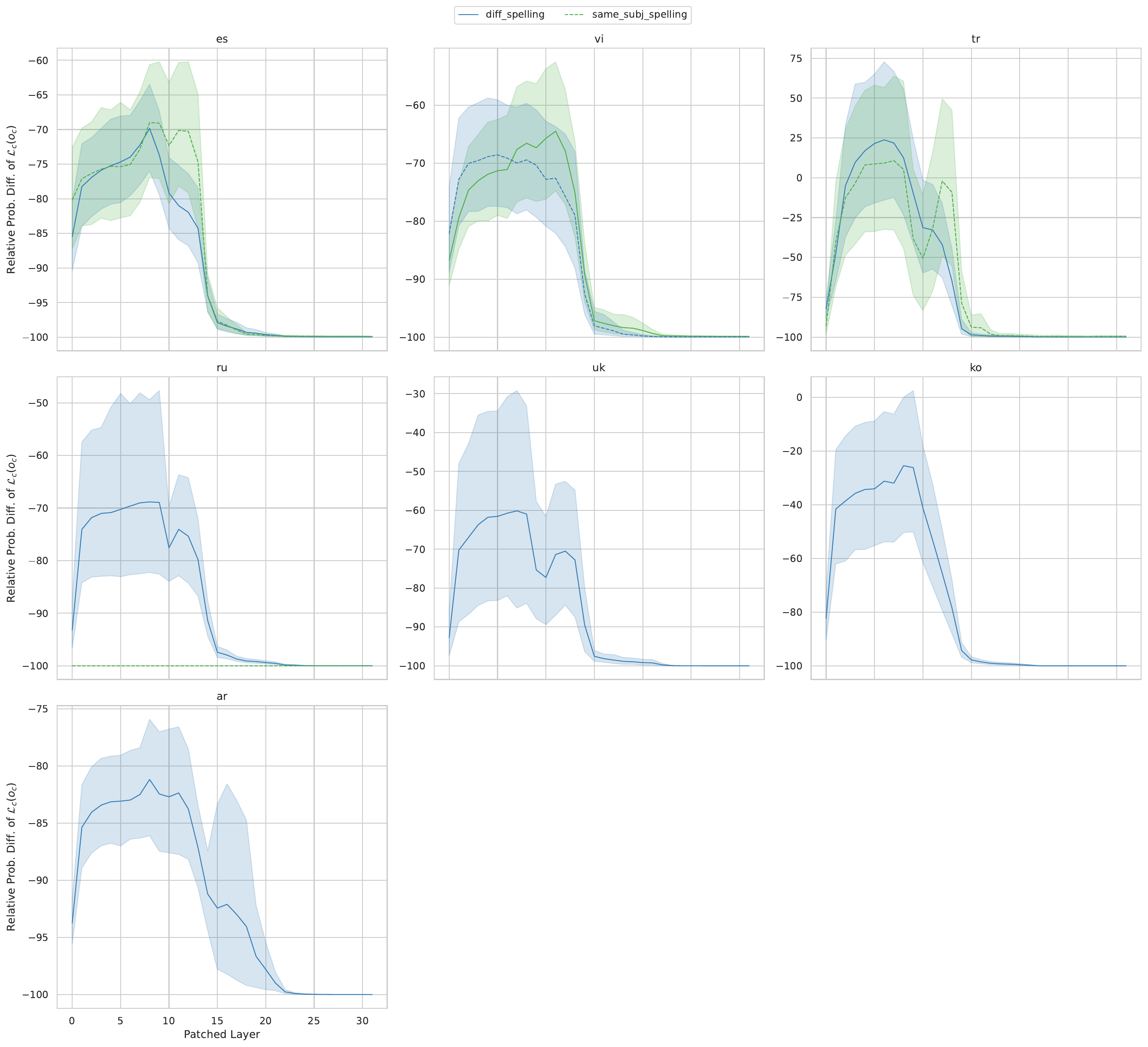} 
    \caption{Probability of the \(\mathcal{L}_c(o_c)\) when patching at different layers in \xglm{}, for examples with \(\{\ne \mathcal{L}, \ne r, = s\}\). Note that the plots do not share the y-axis.}
    \label{fig:xglm_diff_r_prob_context_each_lang}
\end{figure*}

\begin{figure*}[h!]
    \centering
    \includegraphics[width=0.9\textwidth]{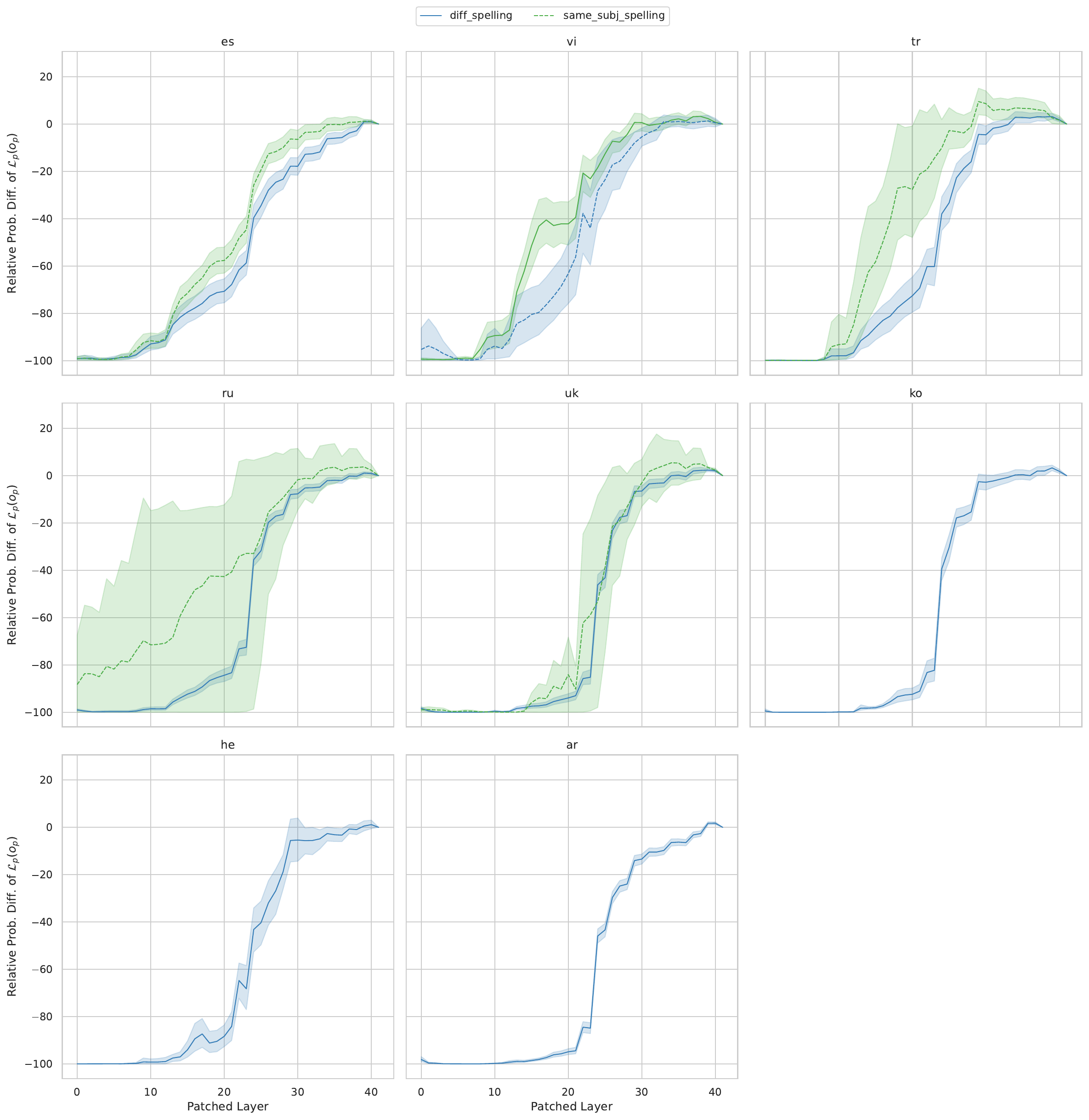} 
    \caption{Probability of the \(\mathcal{L}_p(o_p)\) when patching at different layers in \eurollm{}, , for examples with \(\{\ne \mathcal{L}, \ne r, = s\}\).}
    \label{fig:eurollm_diff_r_prob_patch_each_lang}
\end{figure*}

\begin{figure*}[h!]
    \centering
    \includegraphics[width=0.9\textwidth]{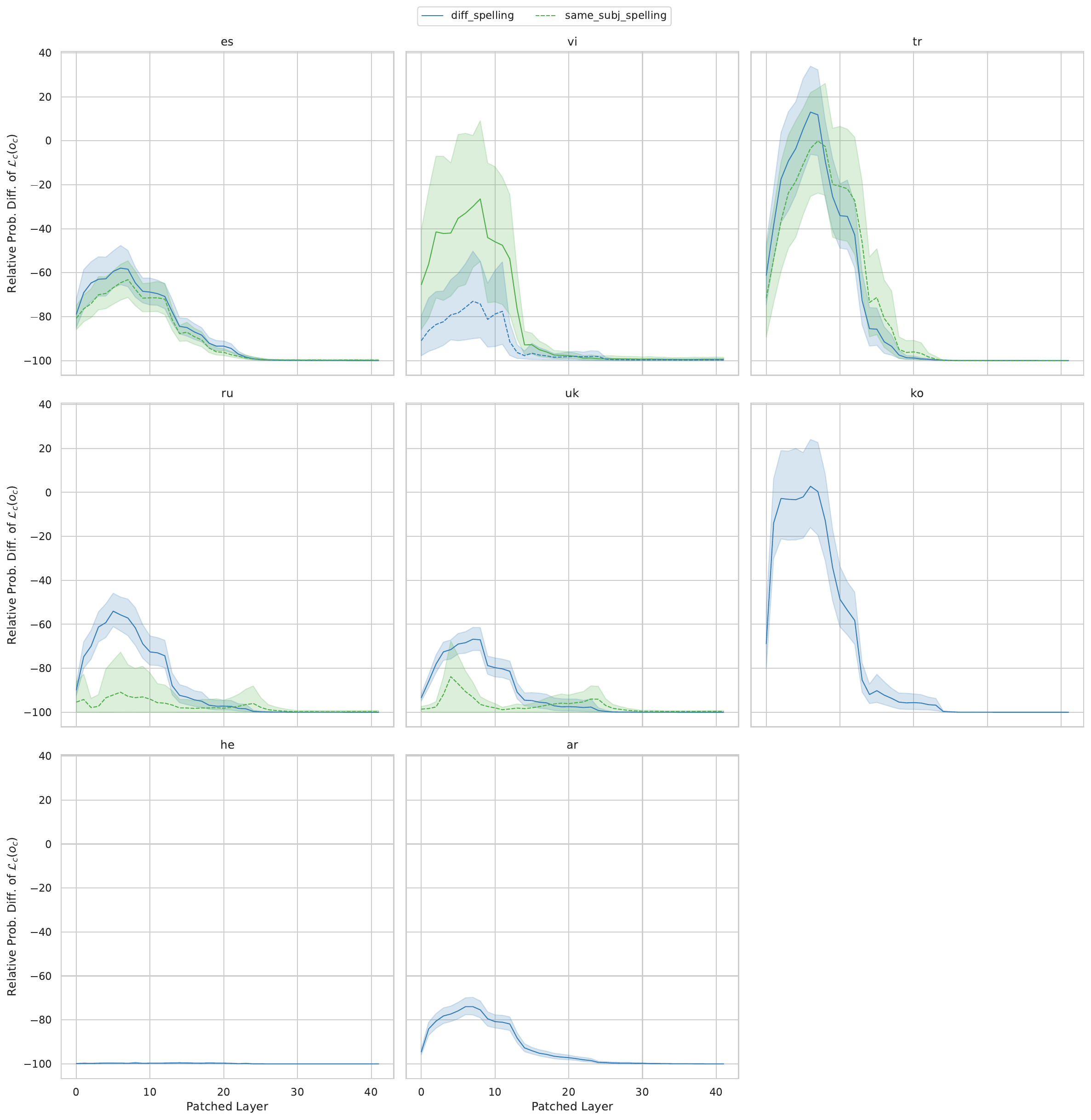} 
    \caption{Probability of the \(\mathcal{L}_c(o_c)\) when patching at different layers in \eurollm{}, for examples with \(\{\ne \mathcal{L}, \ne r, = s\}\). Note that the plots do not share the y-axis.}
    \label{fig:eurollm_diff_r_prob_context_each_lang}
\end{figure*}

\begin{figure*}[h!]
    \centering
    \includegraphics[width=0.9\textwidth]{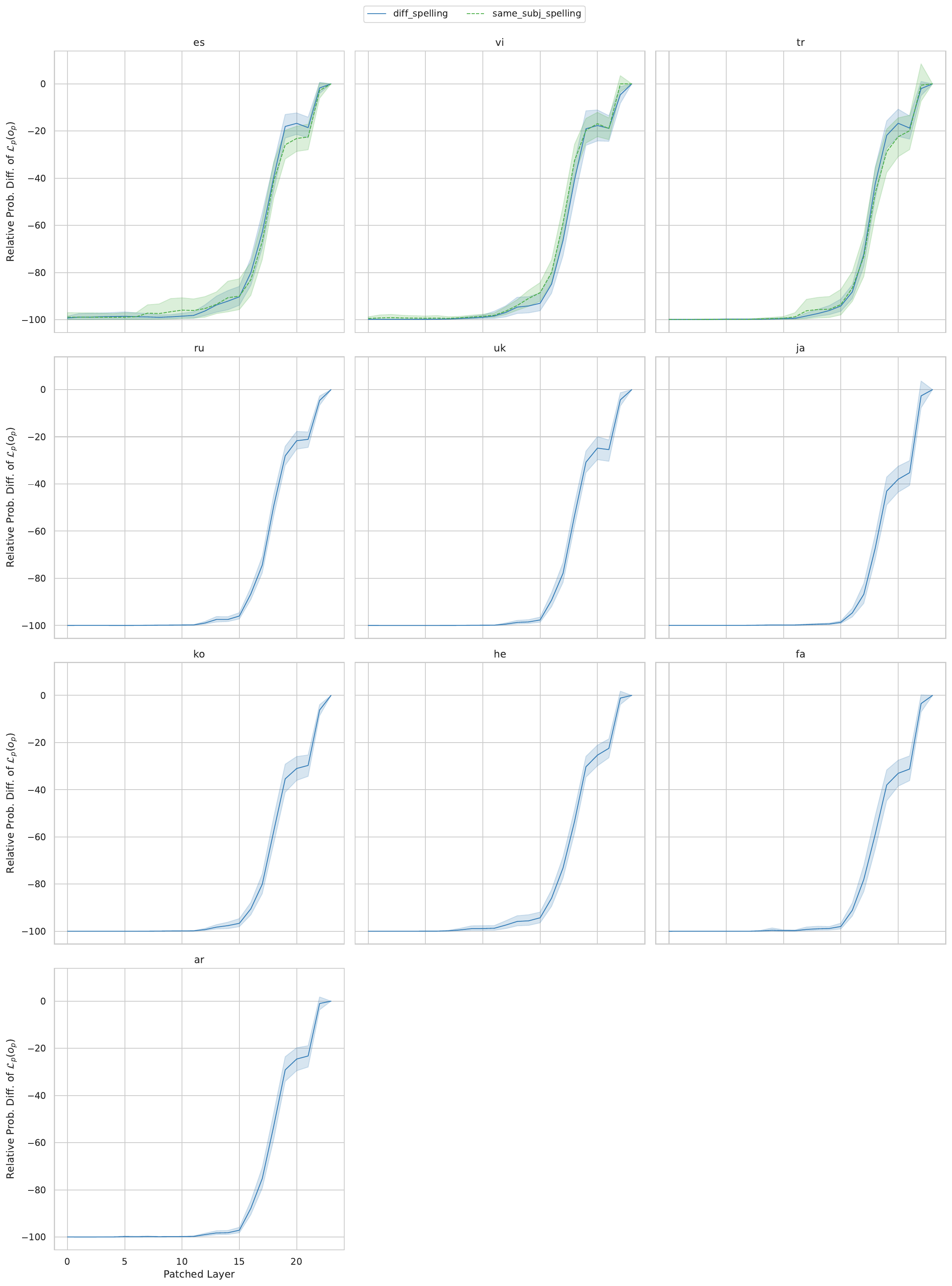} 
    \caption{Probability of the \(\mathcal{L}_p(o_p)\) when patching at different layers in \mtfive{}, , for examples with \(\{\ne \mathcal{L}, \ne r, = s\}\).}
    \label{fig:mt5_diff_r_prob_patch_each_lang}
\end{figure*}

\begin{figure*}[h!]
    \centering
    \includegraphics[width=0.9\textwidth]{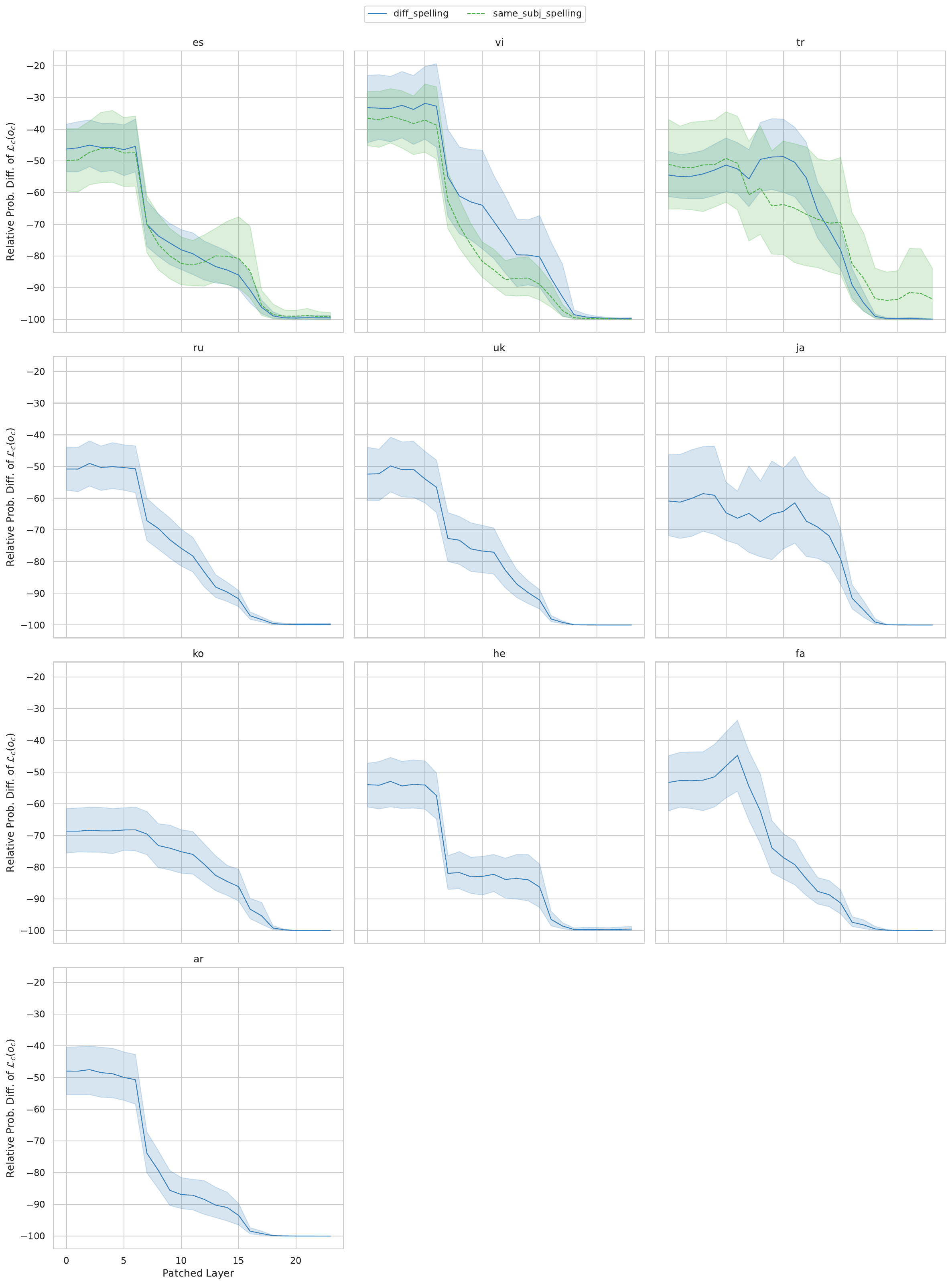} 
    \caption{Probability of the \(\mathcal{L}_c(o_c)\) when patching at different layers in \mtfive{}, for examples with \(\{\ne \mathcal{L}, \ne r, = s\}\). Note that the plots do not share the y-axis.}
    \label{fig:mt5_diff_r_prob_context_each_lang}
\end{figure*}

\begin{figure*}[t]
    \centering
    \vspace{5mm}
    \begin{minipage}{\linewidth}
        \centering
        \includegraphics[width=0.45\linewidth]{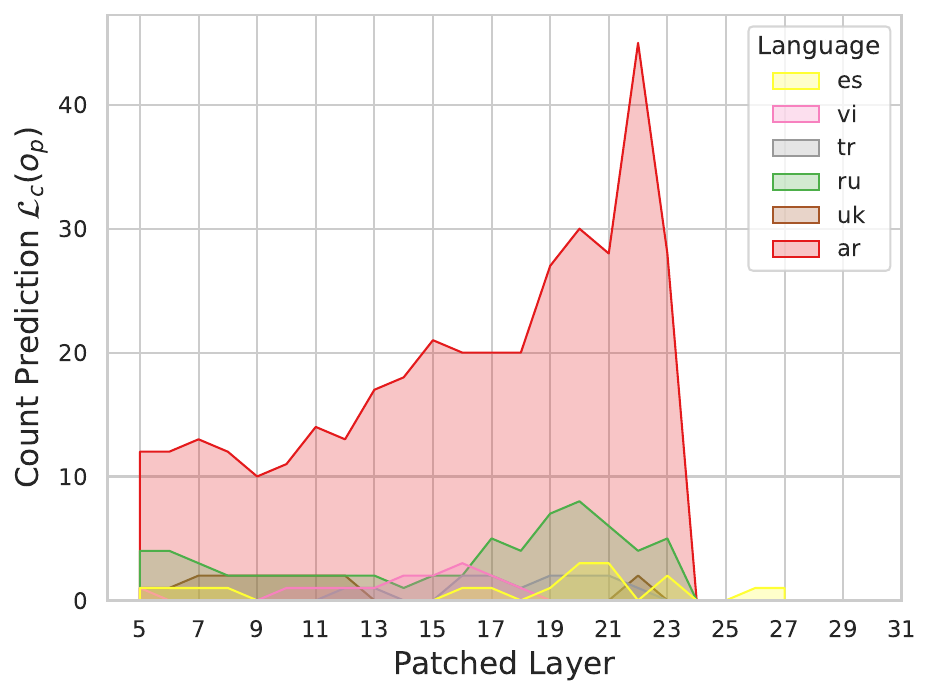}
    \end{minipage}
    \vspace{5mm}
    \begin{minipage}{\linewidth}
        \centering
        \includegraphics[width=0.45\linewidth]{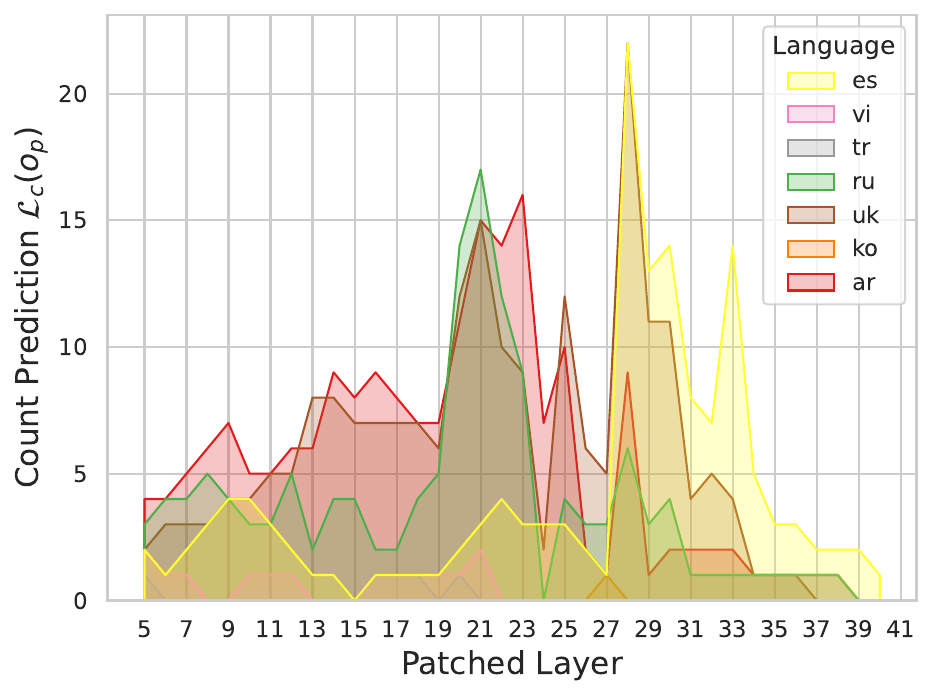}
    \end{minipage}
    \vspace{5mm} 
    \begin{minipage}{\linewidth}
        \centering
        \includegraphics[width=0.45\linewidth]{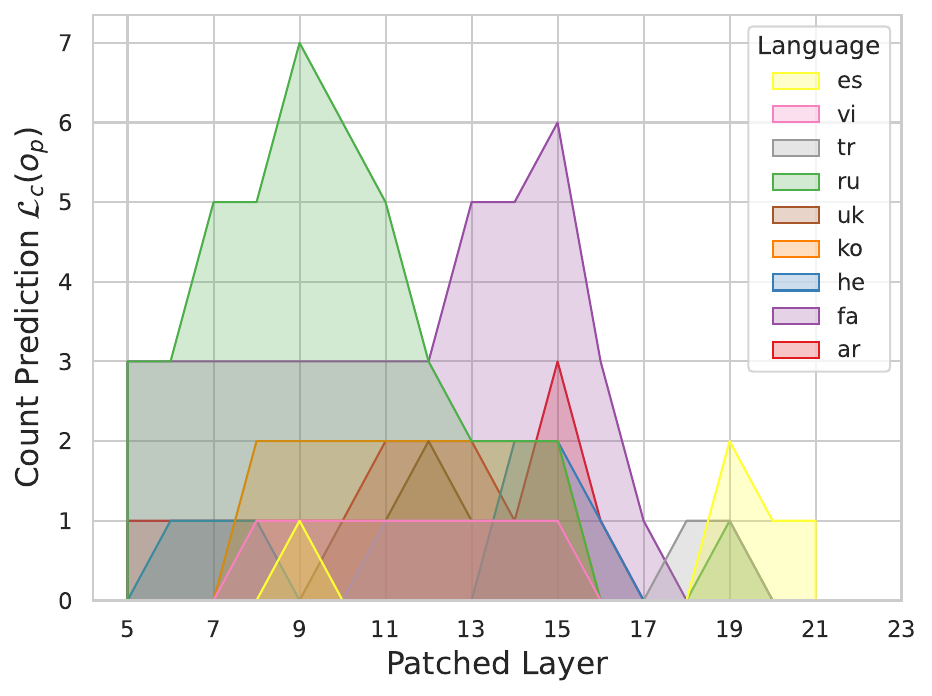}
    \end{minipage}
    \vspace{-2mm}
\caption{Number of times the patch object is predicted in the context language for the experiment of same relation different subject. Models from top to bottom: \xglm{}, \eurollm, \mtfive{}.}\label{fig:same_r_diff_s_layers_appendix}
\vspace{4mm}
\end{figure*}

\begin{figure*}[t]
\centering
\begin{minipage}[t]{0.48\linewidth}
    \centering
    \includegraphics[width=\linewidth]{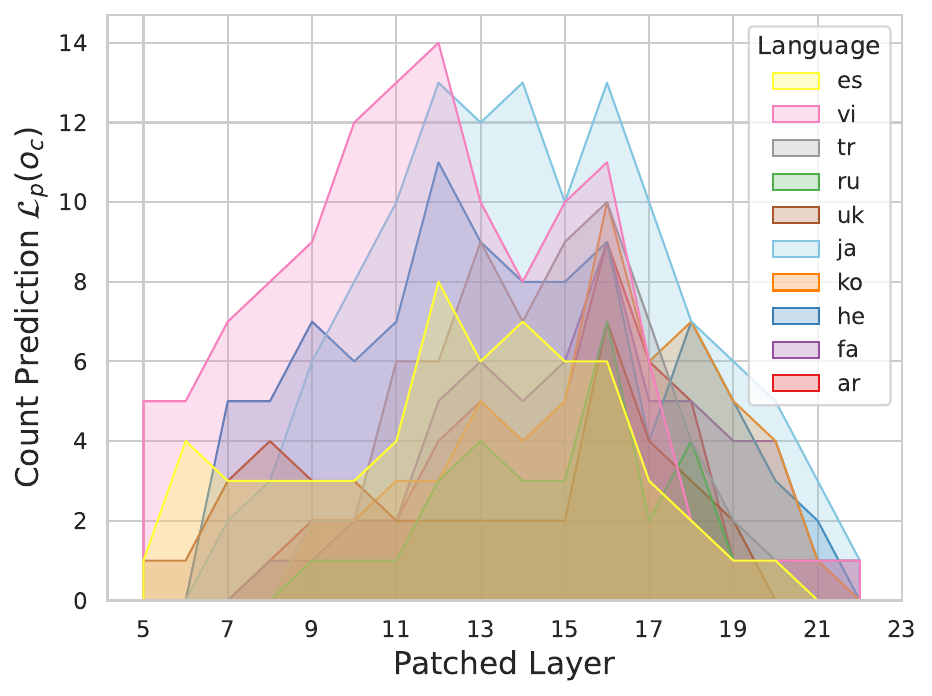}
\end{minipage}
\hfill
\begin{minipage}[t]{0.48\linewidth}
    \centering
    \includegraphics[width=\linewidth]{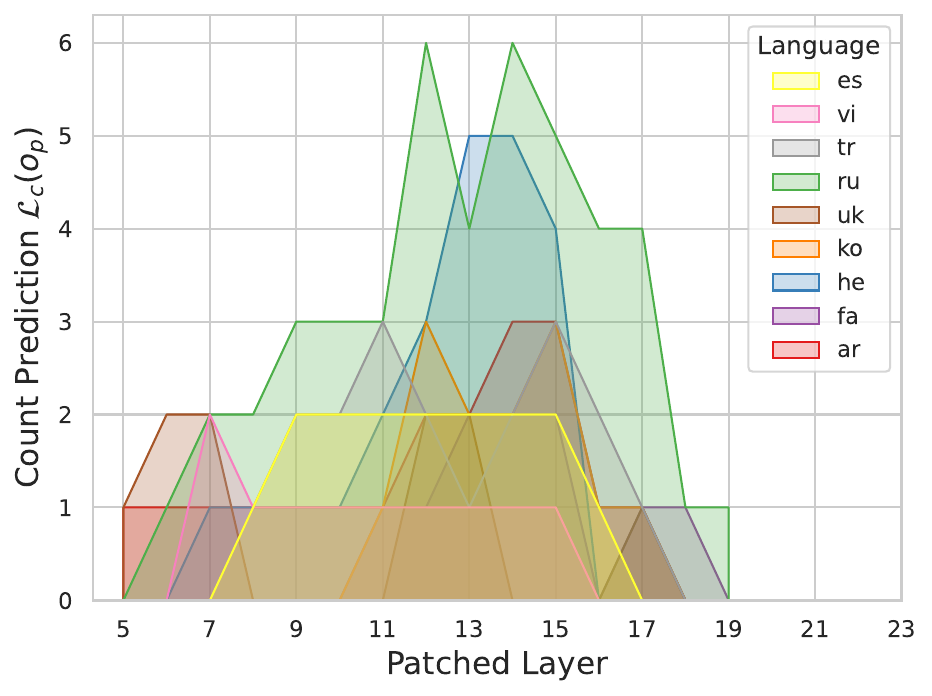}
\end{minipage}
\vspace{-2mm}
\caption{Number of times the object is predicted in the opposite language in \mtfive{} in the \(\{\ne r, =s, \ne \mathcal{L}\}\) experiment.}
\label{fig:mt5_diff_r_same_s_layers}
\vspace{-2mm}
\end{figure*}

\begin{figure*}[t]
\centering
\begin{minipage}[t]{0.48\linewidth}
    \centering
    \includegraphics[width=\linewidth]{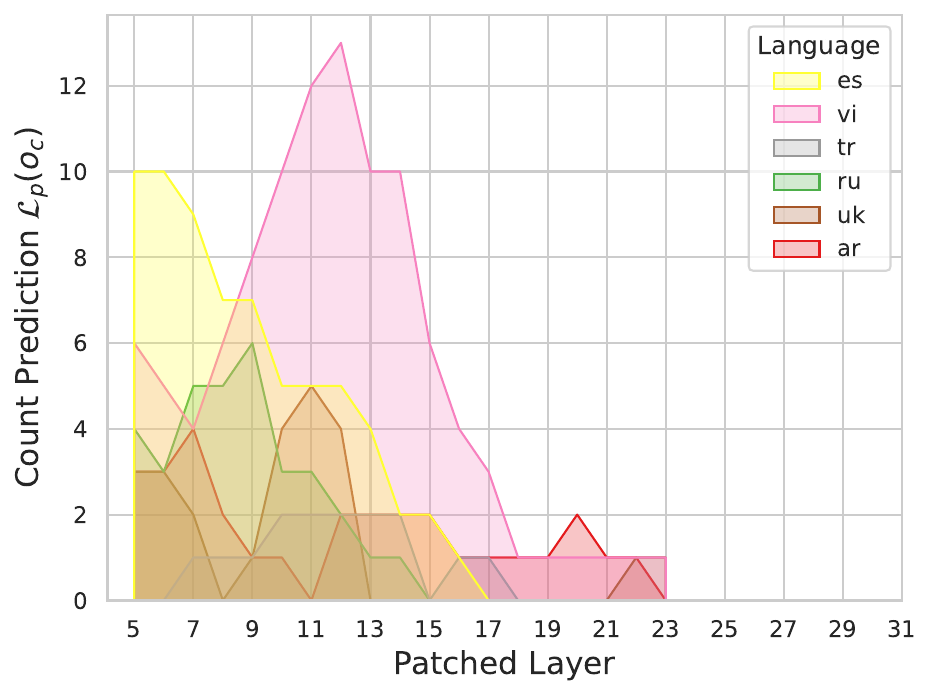}
\end{minipage}
\hfill
\begin{minipage}[t]{0.48\linewidth}
    \centering
    \includegraphics[width=\linewidth]{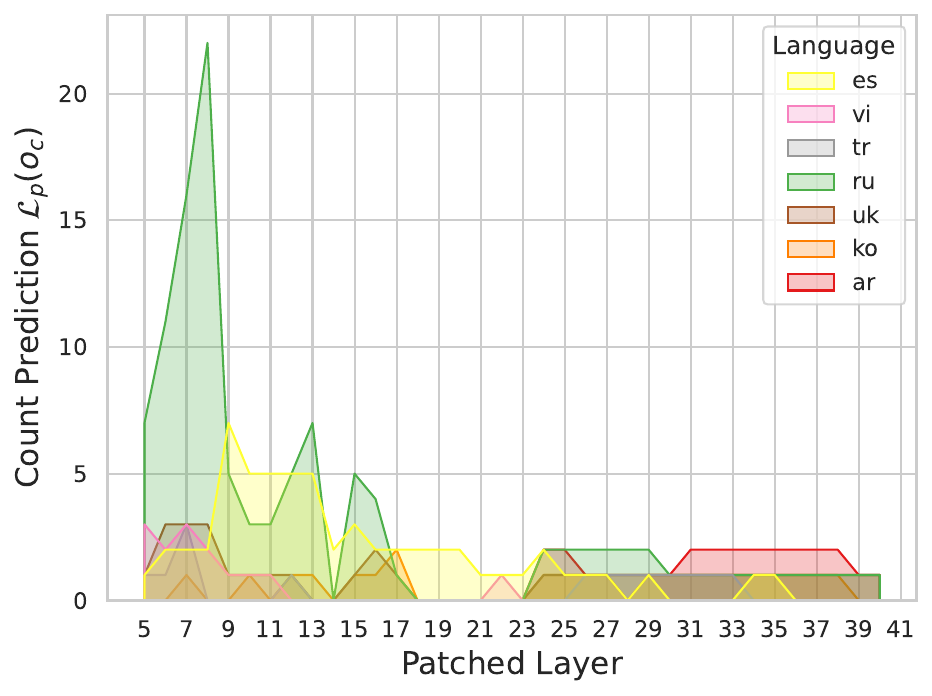}
\end{minipage}
\vspace{-2mm}
\caption{Number of times the context object is predicted in the patch language in the \(\{\ne r, =s, \ne \mathcal{L}\}\) experiment. Left \xglm{}, right \eurollm{}.}
\label{fig:diff_r_same_s_layers_appendix}
\vspace{-2mm}
\end{figure*}

\end{document}